\documentclass[fleqn,10pt]{wlscirep}
\usepackage[utf8]{inputenc}
\usepackage[T1]{fontenc}

\usepackage[
sorting=none,
maxbibnames=99,
backend=biber
]{biblatex}
\definecolor{citecolor}{HTML}{0071bc}
\definecolor{citered}{HTML}{8b0000}
\usepackage{hyperref}
\usepackage{url}

\usepackage{booktabs}

\usepackage{amsmath,amsfonts,bm}

\def\eqref#1{equation~\ref{#1}}
\def\1{\bm{1}}

\def\vx{{\bm{x}}}

\DeclareMathAlphabet{\mathsfit}{\encodingdefault}{\sfdefault}{m}{sl}
\SetMathAlphabet{\mathsfit}{bold}{\encodingdefault}{\sfdefault}{bx}{n}

\DeclareMathOperator*{\argmin}{arg\,min}

\usepackage{graphicx}
\usepackage{float}
\usepackage{amsfonts,amsmath,amssymb,amsthm}
\usepackage{cleveref}
\usepackage{multicol, multirow}
\usepackage{makecell}
\usepackage{caption}
\usepackage{subcaption}
\usepackage{adjustbox}
\usepackage{enumitem}
\usepackage{wrapfig}
\usepackage{algorithm}
\usepackage{algorithmic}
\usepackage{afterpage}

\newcommand{\ie}{\em{i.e.}}
\newcommand{\eg}{\em{e.g.}}
\usepackage{cleveref}

\usepackage{lineno}

\usepackage{xcolor}

\newcommand{\one}{one}
\newcommand{\two}{two}
\newcommand{\three}{three}
\newcommand{\four}{four}
\newcommand{\five}{five}
\newcommand{\six}{six}
\newcommand{\eight}{eight}
\newcommand{\nine}{nine}
\newcommand{\TotalEditingTaskNum}{20}
\newcommand{\model}[1]{MoleculeSTM}
\newcommand{\DatasetName}[1]{PubChemSTM}

\title{
Multi-modal Molecule Structure-text Model for Text-based Retrieval and Editing
}
\author[1,2]{Shengchao Liu}
\author[3]{Weili Nie}
\author[4]{Chengpeng Wang}
\author[1,2]{Jiarui Lu}
\author[5]{Zhuoran Qiao}
\author[6]{Ling Liu}
\author[1,7]{Jian Tang$^*$}
\author[3,8]{Chaowei Xiao$^*$}
\author[3,5]{Animashree Anandkumar$^*$}
\affil[1]{Mila-Québec Artificial Intelligence Institute, Montréal, QC H2S 3H1, Canada}
\affil[2]{Université de Montréal, Montréal, QC H3T 1J4, Canada}
\affil[3]{NVIDIA Research, Santa Clara, CA 95051, United States}
\affil[4]{University of Illinois Urbana-Champaign, Champaign, IL 61801, United States}
\affil[5]{California Institute of Technology, Pasadena, CA 91125, United States}
\affil[6]{Princeton University, Princeton, NJ 08544, United States}
\affil[7]{HEC Montréal, Montréal, QC H3T 2A7, Canada}
\affil[8]{Arizona State University, Tempe, AZ 85281, United States}

\begin{abstract}
There is increasing adoption of artificial intelligence in drug discovery. However, existing studies use machine learning to mainly utilize the chemical structures of molecules but ignore the vast textual knowledge available in chemistry. Incorporating textual knowledge enables us to realize new drug design objectives, adapt to text-based instructions and predict complex biological activities. Here we present a multi-modal molecule structure-text model, MoleculeSTM, by jointly learning molecules' chemical structures and textual descriptions via a contrastive learning strategy. To train MoleculeSTM, we construct a large multi-modal dataset, namely, PubChemSTM, with over 280,000 chemical structure-text pairs. To demonstrate the effectiveness and utility of MoleculeSTM, we design two challenging zero-shot tasks based on text instructions, including structure-text retrieval and molecule editing. MoleculeSTM has two main properties: open vocabulary and compositionality via natural language. In experiments, MoleculeSTM obtains the state-of-the-art generalization ability to novel biochemical concepts across various benchmarks.\looseness=-1
\end{abstract}

\begin{document}
\flushbottom
\maketitle
\thispagestyle{empty}

\begin{figure}[ht!]
\centering
\includegraphics[width=\linewidth]{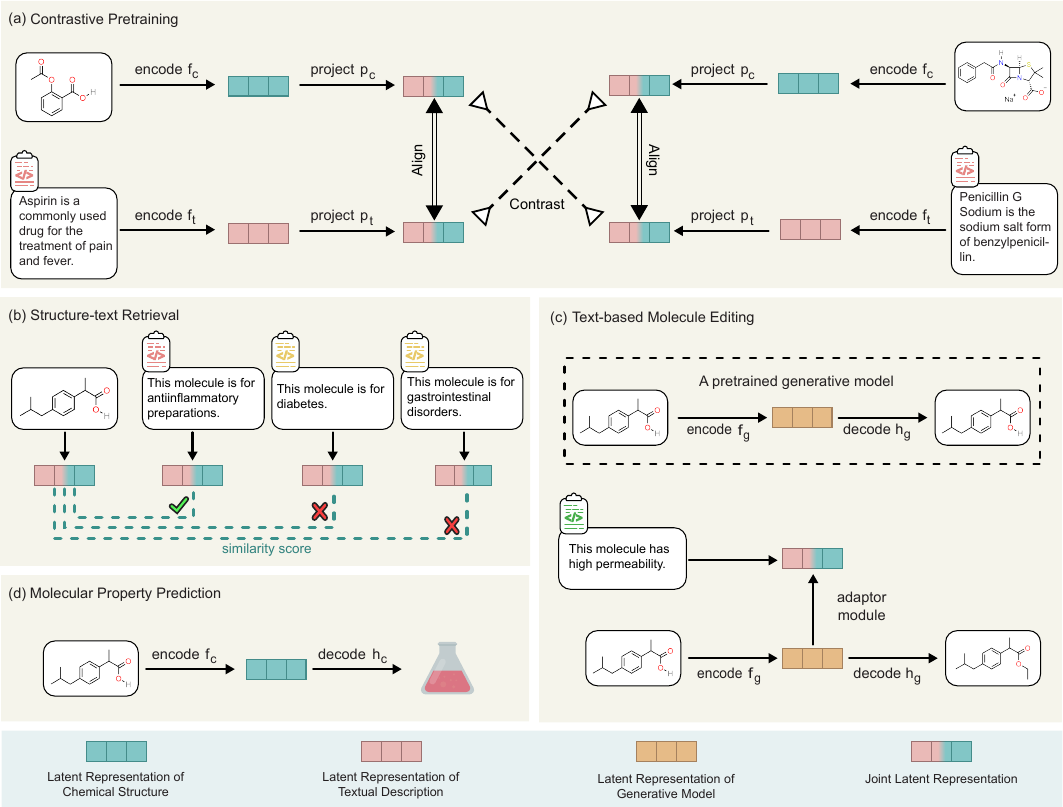}
\vspace{-3ex}
\caption{
Pipeline of pretraining and downstream tasks.
(a) \model{} pretraining with two branches, the chemical structure (green) and textual description (pink).
(b) Structure-text retrieval downstream task.
(c) Text-based molecule editing downstream task.
(d) Molecular property prediction downstream task.
}
\label{fig:pipeline}
\vspace{-3ex}
\end{figure}

Recent progress in artificial intelligence (AI) promises to be transformative for drug discovery~\cite{sullivan2019tough}. AI methods have been used to augment and accelerate current computational pipelines~\cite{patronov2022has,jayatunga2022ai,jumper2021highly}, including but not limited to virtual screening~\cite{doi:10.1021/ci8002649,liu2018practical}, metabolic property prediction~\cite{duvenaud2015convolutional,liu2018n,wu2018moleculenet}, and targeted chemical structure generation and editing~\cite{jin2020hierarchical,irwin2022chemformer,wang2022retrieval,liu2022graphcg}.\looseness=-1

Existing machine learning (ML) methods mainly focus on modeling the chemical structure of molecules through one-dimensional descriptions~\cite{krenn2020self}, two-dimensional molecular graphs~\cite{duvenaud2015convolutional,xu2018powerful,liu2018n}, or three-dimensional geometric structures~\cite{schutt2018schnet,satorras2021n,atz2021geometric}. They also use supervised signals, {\eg}, toxicity labels, quantum-mechanical properties, and binding affinity measurements. However, such a supervised setting requires expensive annotations on pre-determined label categories, impeding the application to unseen categories and tasks~\cite{ji2022drugood}. To overcome this issue, unsupervised pretraining on large-scale databases~\cite{irwin2012zinc} has been proposed, with the main advantage being the ability to learn chemical structures without supervised annotation by reconstructing the masked topological~\cite{hu2019strategies} or geometric~\cite{liu2022molecular} substructures. Compared to the supervised setting, although such pretrained models~\cite{hu2019strategies,liu2022molecular} have proven to be more effective in generalizing to various downstream tasks by fine-tuning on a few labeled examples, it is still an open challenge to generalize unseen categories and tasks without such labeled examples or fine-tuning ({\ie}, the so-called \textit{zero-shot} setting~\cite{larochelle2008zero} in ML). Additionally, existing molecule pretraining methods mostly incorporate only chemical structures, leaving the multi-modal representation less explored.\looseness=-1

We have a vast amount of textual data that is human-understandable and easily accessible. This is now being harnessed in large-scale multi-modal models for images and videos~\cite{radford2021learning,nichol2021glide,ramesh2022hierarchical,patashnik2021styleclip}. A natural language interface is an intuitive way to enable open vocabulary and description of tasks. Pretrained multi-modal models can generalize well to new categories and tasks, even in the zero-shot setting~\cite{radford2021learning,nichol2021glide,ramesh2022hierarchical,patashnik2021styleclip}. They also enable agents to interactively learn to solve new tasks and explore new environments~\cite{li2022pre,fan2022minedojo}. We believe similar capabilities can also be obtained in molecular models by incorporating the vast textual knowledge available in the literature.\looseness=-1

Previous work~\cite{zeng2022deep} has attempted to leverage the textual knowledge to learn the molecule representation. However, it only supports modeling with the 1D description (the simplified molecular-input line-entry system or SMILES) and learns the chemical structures and textual descriptions on a small-scale dataset (10K structure-text pairs). Furthermore, it unifies \two{} modalities into a single language modeling framework and requires aligned data, {\ie}, chemical structure and text for each sample, for training. As a result, it cannot adopt existing powerful pretrained models, and the availability of aligned data is extremely limited.\looseness=-1

{\bf Our approach:} We design a multi-modal foundation model for molecular understanding that incorporates both molecular structural information and textual knowledge. We demonstrate zero-shot generalization to new drug design objectives using text-based instructions and to the prediction of new complex biological activities without the need for labeled examples or fine-tuning.\looseness=-1

We propose \model{}, consisting of \two{} branches: the chemical structure branch and the textual description branch, to handle the molecules' internal structures and external domain knowledge, respectively. Such a disentangled design enables \model{} to be integrated with the powerful existing models trained on each modality separately, {\ie}, molecular structural models~\cite{irwin2022chemformer,liu2021pre} and scientific language models~\cite{Beltagy2019SciBERT}. Given these pretrained models, \model{} bridges the \two{} branches via a contrastive learning paradigm~\cite{liu2021pre,oord2018representation}.\looseness=-1

To align such two branches with \model{}, we construct a structure-text dataset called \DatasetName{} from PubChem~\cite{kim2021pubchem}, which is the largest multi-modal dataset to date in the community (28$\times$ larger than the existing dataset~\cite{zeng2022deep}). In \DatasetName{}, each chemical structure is paired with a textual description, illustrating the chemical and physical properties or high-level bioactivities accordingly. Since \model{} is trained on a large-scale structure-text pair dataset and such textual data contains open-ended chemical information, it can be generalized to diverse downstream tasks in a zero-shot manner.\looseness=-1

To demonstrate the advantages of introducing the language modality, we design \two{} challenging downstream tasks: the structure-text retrieval task and text-based molecule editing task, and we apply the pretrained \model{} on them in a zero-shot manner. By studying these tasks, we summarize \two{} main attributes of \model{}: the open vocabulary and compositionality. (1) Open vocabulary means our proposed \model{} is not limited to a fixed set of pre-defined molecule-related textual descriptions and can support exploring a wide range of biochemical concepts with the unbound vocabulary depicted by the natural language. In the drug discovery pipeline, such an attribute can be used for the text-based molecule editing in the lead optimization task and the novel disease-drug relation extraction in the drug re-purposing task. (2) Compositionality implies that we can express a complex concept by decomposing it into several simple concepts. This can be applied for the text-based multi-objective lead optimization task~\cite{hughes2011principles} where the goal is to generate molecules satisfying multiple properties simultaneously.\looseness=-1

Empirically, \model{} reaches the best performance on \six{} zero-shot retrieval tasks (up to 50\% higher accuracy) and \TotalEditingTaskNum{} zero-shot text-based editing tasks (up to 40\% higher hit ratio) compared to the state-of-the-art methods. Furthermore, for molecular editing tasks, visual inspections reveal that \model{} can successfully detect critical structures implied in text descriptions. Additionally, we also explore whether \model{} can improve the performance on the standard molecular property prediction benchmark~\cite{wu2018moleculenet} via fine-tuning. Our results show that \model{} can achieve the best overall performance among \nine{} baselines on \eight{} property prediction tasks.\looseness=-1

%%%%%%%%%%%%%%%%%%%%%%%%%%%%%%%%%%%%%%%%%%%%%%%%%%
\section*{Results}
\subsection*{Overview and Preliminaries}
In this section, we first provide an overview of \model{}. Then, we introduce how to pretrain \model{} and apply the pretrained \model{} to three types of downstream tasks (\Cref{fig:pipeline}).\looseness=-1

\textbf{Overview.} \model{} consists of two branches: the chemical structure branch and the textual description branch ($\vx_c$ and $\vx_t$). The chemical structure branch illustrates the arrangement of atoms in a molecule. We consider \two{} types of encoders $f_c$: Transformer~\cite{vaswani2017attention} on the SMILES string and GNNs~\cite{duvenaud2015convolutional,liu2018n,xu2018powerful} on the 2D molecular graph. The textual description branch provides a high-level description of the molecule's functionality, and we use the language model from a recent work~\cite{devlin2018bert} as the encoder $f_t$.\looseness=-1

\textbf{Pretraining.} Within this design, \model{} aims to map the representations extracted from \two{} branches to a joint space using \two{} projectors ($p_c$ and $p_t$) via contrastive learning~\cite{liu2021pre,oord2018representation}. The essential idea of contrastive learning is to reduce the representation distance between the chemical structure and textual description pairs of the same molecule and increase the representation distance between the pairs from different molecules. Specifically, we initialize these two branch encoders with the pretrained single-modal checkpoints~\cite{irwin2022chemformer,liu2021pre,Beltagy2019SciBERT} and then perform an end-to-end contrastive pretraining on collected dataset \DatasetName{}.
Specifically for \DatasetName{}, it is constructed from PubChem~\cite{kim2021pubchem}. We extract molecules with the textual description fields, leading to 281K chemical structure and text pairs.
More details can be found in Supplementary A.1.
\looseness=-1

%%%%%%%%%%%%%%%%%%%%%%%%%%%%%%%%%%%%%%%%%%%%%%%%%%
\subsection*{Two Principles for Downstream Task Design}
We want to emphasize that for these downstream tasks, the language model in the pretrained \model{} reveals certain appealing attributes for molecule modeling and drug discovery. We summarize the \two{} key points below.\looseness=-1

\textbf{Open vocabulary.} Language is by nature open vocabulary and free form~\cite{gu2021open}. The large language model has proven its generalization ability in various art-related applications~\cite{radford2021learning,nichol2021glide,ramesh2022hierarchical}, and we find that it can also provide promising and insightful observations for drug discovery tasks. In this vein, our method is not limited to a fixed set of pre-defined molecule-related annotations but can support the exploration of novel biochemical concepts with unbound vocabulary. One example is the drug re-purposing. Suppose we have a textual description for a new disease or protein target functionality. In that case, we can obtain its similarity with all the existing drugs using \model{} and retrieve the drugs with the highest rankings, which can be adopted for the later stages, such as clinical trials. Another example is text-based lead optimization. We use natural language to depict an entirely new property, which can be reflected in the generated molecules after the optimization.\looseness=-1

\textbf{Compositionality.} Another attribute is compositionality. In natural language, a complex concept can be expressed by decomposing it into simple concepts. This is crucial for certain domain-specific tasks, {\eg}, multi-objective lead optimization~\cite{hughes2011principles} where we need to generate molecules with multiple desired properties simultaneously. Existing solutions are either (1) learning one classifier for each desired property and doing filtering on a large candidate pool~\cite{jin2020hierarchical} or (2) optimizing a retrieval database to modify molecules to achieve the multi-objective goal~\cite{wang2022retrieval}. The main limitation is that the success ratio highly depends on the availability of the labeled data for training the classifier or the retrieval database. While with the language model in \model{}, we provide an alternative solution. We first craft a natural text, called the text prompt, as the task description. The text prompt can be multi-objective and consists of the description for each property ({\eg}, ``molecule is soluble in water and has high permeability''). With the pretrained joint space between chemical structures and textual descriptions, \model{} can transform the molecule property compositionality problem into the language compositionality problem, which is more tractable using the language model.\looseness=-1

%%%%%%%%%%%%%%%%%%%%%%%%%%%%%%%%%%%%%%%%%%%%%%%%%%
\subsection*{Downstream: Zero-shot Structure-text Retrieval}
\textbf{Experiments.} For the zero-shot retrieval, we construct \three{} datasets from DrugBank~\cite{wishart2018drugbank}. DrugBank is by far the most comprehensive database for drug-like molecules. Here we extract \three{} fields in DrugBank: the description field, the pharmacodynamics field, and the anatomical therapeutic chemical (ATC) field. These fields illustrate the chemical properties and drug effects on the target organism. Then the retrieval task can be viewed as a $T$-choose-one multiple-choice problem, where $T$ is the number of choices. Specifically, we have \two{} settings: (1) given chemical structure to retrieve the textual description and (2) given the textual description to retrieve the chemical structure. The retrieval accuracy is used as the evaluation metric.\looseness=-1

\textbf{Baselines.} We first consider \two{} baselines with the pretrained single-modal encoders~\cite{irwin2022chemformer,liu2021pre,Beltagy2019SciBERT}. (1) \textit{Frozen} is that we take the pretrained encoders for the \two{} branches and \two{} randomly initialized projectors. (2) \textit{Similarity} is that we take the similarity from a single branch only. For example, in the first setting, when given chemical structure, we retrieve the most similar chemical structure from \DatasetName{}, then we take the corresponding paired text representation in \DatasetName{} as the proxy representation. Based on this, we can calculate the similarity score between the proxy representation and $T$ requested text representations. (3) We further consider the third baseline, a pretrained language model for knowledgeable and versatile machine reading (KV-PLM)~\cite{zeng2022deep} on SMILES-text pairs.\looseness=-1

\begin{figure}[t!]
\centering
\includegraphics[width=\linewidth]{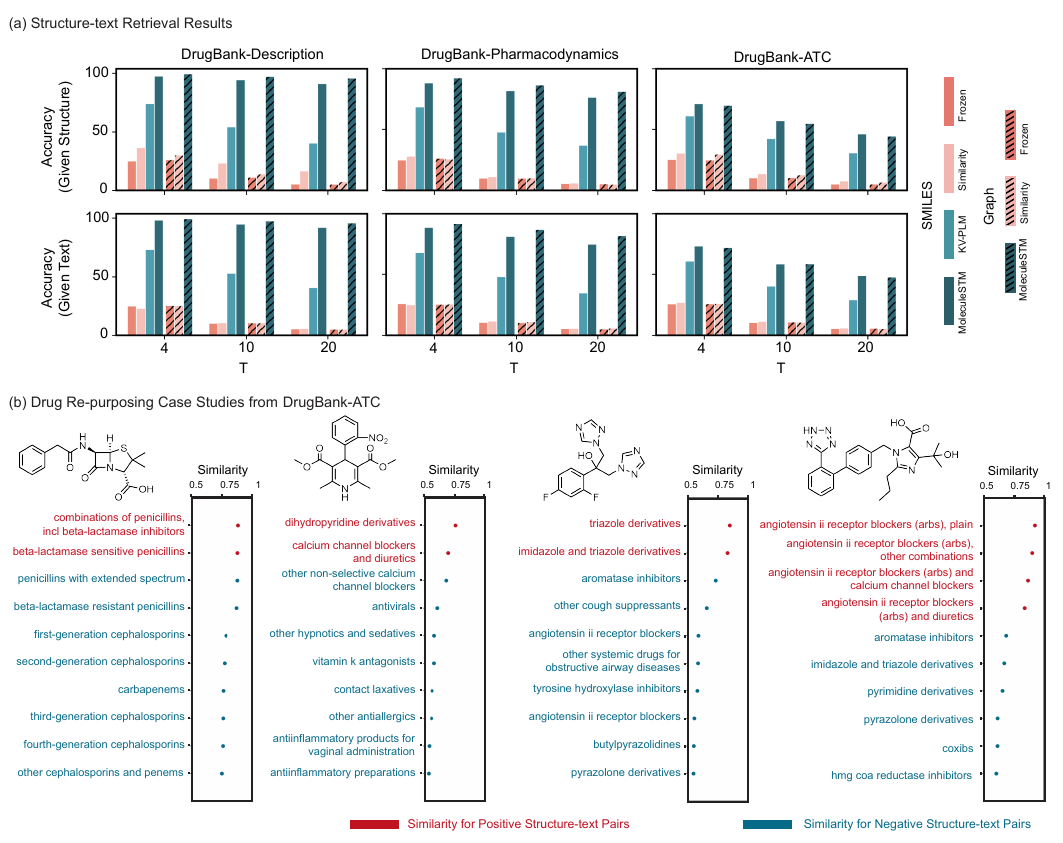}
\vspace{-3ex}
\caption{
Results for zero-shot structure-text retrieval. (a) Accuracy for zero-shot structure-text retrieval on three DrugBank datasets. (b) Four case studies on DrugBank-ATC retrieval. HMG-CoA is $\beta$-Hydroxy $\beta$-methylglutaryl-CoA.
}
\label{fig:retrieval_general_results}
\end{figure}

\textbf{Results.} The zero-shot retrieval results are shown in~\Cref{fig:retrieval_general_results} (a). First, we observe that all the algorithms' accuracies are quite similar between the \two{} settings. Then, as expected, we observe that the baseline Frozen performs no better than the random guess because of the randomly-initialized projectors. The Similarity baseline is better than the chance performance by a modest margin, verifying that the pretrained single-modality does learn semantic information but cannot generalize well between modalities. KV-PLM, on the other hand, learns semantically meaningful information from SMILES-text pairs, and thus, it achieves much higher accuracies on \three{} datasets. For \model{}, the graph representation from GNNs has higher accuracy on Description and Pharmacodynamics than the SMILES representation from the transformer model; yet, both of them outperform all the other methods on \three{} datasets and \two{} settings by a large margin. For example, the accuracy improvements are around 50\%, 40\%, and 15\% compared to the best baseline with $T=20$. Such large improvement gaps verify that \model{} can play a better role in understanding and bridging the \two{} modalities of molecules.\looseness=-1

\textbf{Case study on drug re-purposing analysis.} In~\Cref{fig:retrieval_general_results} (b), we further show \four{} case studies on the retrieval quality of ATC. Specifically, given the molecule's chemical structure, we take 10 (out of 600) most similar ATC labels. It is observed that \model{} can retrieve the ground-truth ATC labels with high rankings.\looseness=-1

%%%%%%%%%%%%%%%%%%%%%%%%%%%%%%%%%%%%%%%%%%%%%%%%%%
\subsection*{Downstream: Zero-shot Text-based Molecule Editing}

\begin{figure}[t!]
\centering
\includegraphics[width=\linewidth]{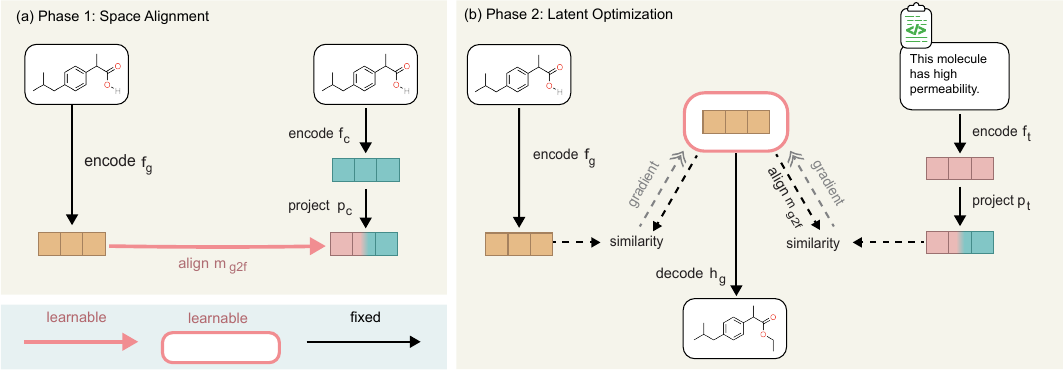}
\vspace{-3ex}
\caption{
Pipelines for the zero-shot text-based molecule editing.
(a) The space alignment step aligns the representation space of a pretrained molecule generation model and the representation space of \model{}.
(b) The latent optimization step learns a latent representation that can be similar to both input molecules and textual descriptions.
}
\label{fig:editing_pipeline}
\end{figure}

\textbf{Experiments.} For molecule editing, we randomly sample 200 molecules from ZINC~\cite{irwin2012zinc} and a text prompt as the inputs. Four categories of text prompts have been covered: (1) \textit{Single-objective editing} is the text prompt using the single drug-related property for editing, such as ``molecule with high solubility'' and ``molecule more like a drug''. (2) \textit{Multi-objective (compositionality) editing} is the text prompt applying multiple properties simultaneously, such as ``molecule with high solubility and high permeability''. (3) \textit{Binding-affinity-based editing} is the text prompt for assay description, where each assay corresponds to one binding affinity task. A concrete example is ChEMBL 1613777~\cite{mendez2018chembl} with prompt as ``This molecule is tested positive in an assay that are inhibitors and substrates of an enzyme protein. It uses molecular oxygen inserting one oxygen atom into a substrate, and reducing the second into a water molecule.''. The output molecules should possess higher binding affinity scores. (4) \textit{Drug relevance editing} is the text prompt to make molecules structurally similar to certain common drugs, {\eg}, ``this molecule looks like Penicillin''. We expect the output molecules to be more similar to the target drug than the input drug. For more detailed descriptions of the text prompts, please check Supplementary D. The evaluation is the satisfactory hit ratio, and it is a hit if the metric difference between output and input is over threshold $\Delta$. The $\Delta$ value is task-specific, and we consider two typical cases: $\Delta=0$ indicates a loose condition, and $\Delta>0$ is a strict condition with a larger positive influence. We provide the algorithm pipeline in~\Cref{fig:editing_pipeline}, and more details can be found in the Methods Section.
\looseness=-1

\textbf{Baselines.} We consider \four{} baselines. The first \three{} baselines~\cite{liu2022graphcg} modify the representation of input molecules, followed by the decoding to the molecule space. \textit{Random} is that we take a random noise as the perturbation to the representation of input molecules. \textit{PCA} is that we take the eigenvectors as latent directions, where the eigenvectors are obtained after decomposing the latent representation of input molecules using principle component analysis (PCA). \textit{High Variance} is that we take the latent representation dimension with the highest variance and apply the one-hot encoding on it as a semantic direction for editing. In addition, we also consider a baseline directly modifying the molecule space, the \textit{genetic search (GS)}. It is a variant of graph genetic algorithm~\cite{jensen2019graph}, while the difference is that GS does a random search instead of a guided search by a reward function since no retrieval database is available in the zero-shot setting.\looseness=-1

\begin{figure}[t!]
\centering
\includegraphics[width=\linewidth]{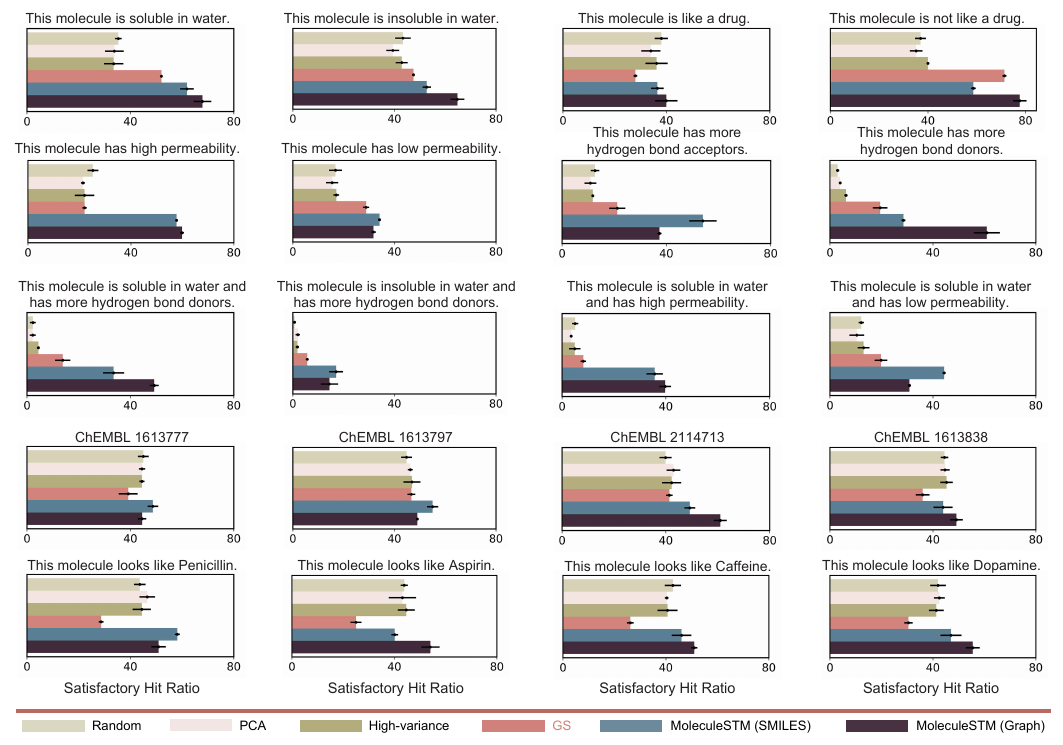}
\vspace{-3ex}
\caption{
Visualization results for the zero-shot text-based molecule editing. Satisfactory hit ratios (\%) of \four{} types text-based editing tasks: \eight{} single-objective, \four{} multi-objective, \four{} ChEMBL binding-affinity-based editing tasks (pretrained random forest as an evaluator, and detailed text prompts are in Supplementary D), and \four{} drug relevance editing tasks. The satisfactory threshold ($\Delta$) is 0 for all visualized results. Each task runs for \three{} random seeds, and the length of each error bar represents the standard deviation.
}
\label{fig:editing_general_results}
\end{figure}

\begin{figure}[t!]
\centering
\includegraphics[width=\linewidth]{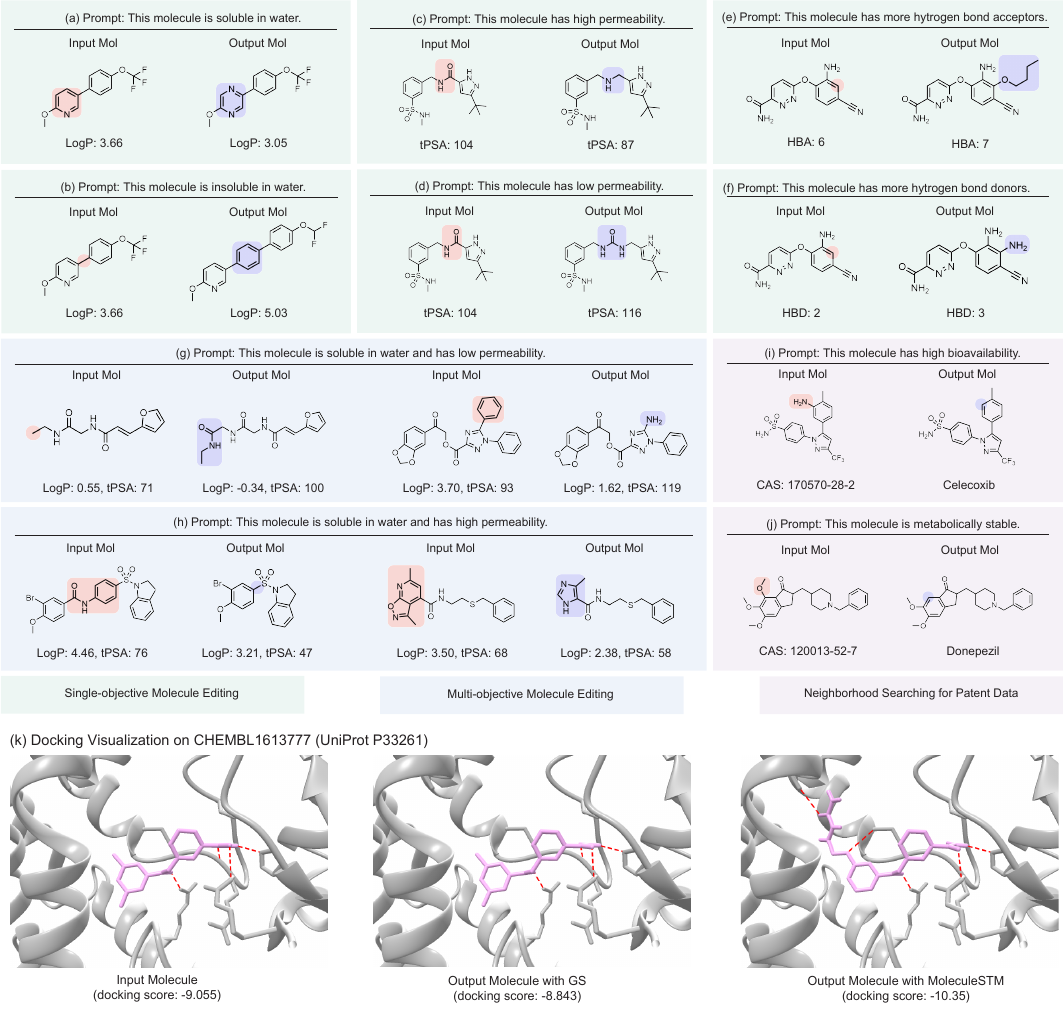}
\vspace{-3ex}
\caption{\small
Visual analysis on text-based molecule editing. 
Case studies for solubility editing (a,b), permeability editing (c,d), acceptor and donor editing (e,f), solubility and permeability editing (g,h), and neighborhood searching for patent data (i,j).
The pink and blue regions mark the functional groups before and after the editing, and we list the chemical abstracts service (CAS) registry number. (k) visualizes binding-affinity-based editing, and the dashed red lines mark the potential bindings.
}
\label{fig:editing_visualization_analysis}
\end{figure}

\textbf{Results.} First, we provide the quantitative results for \TotalEditingTaskNum{} editing tasks across \four{} editing task types in~\Cref{fig:editing_general_results}. The empirical results illustrate that the satisfactory hit ratios of \model{} are the best among all \TotalEditingTaskNum{} tasks. It verifies that, for both SMILES and molecular graph encoders, \model{} enables a better semantic understanding of the natural language to explore output molecules with the desired properties. Next, we scrutinize the quality of output molecules in~\Cref{fig:editing_visualization_analysis} with detailed analysis as follows.\looseness=-1

\textbf{Visual analysis on single-objective molecule editing.} We visually analyze the difference between input and output molecules using the single-objective property. Typical modifications are the addition, removal, and replacement of functional groups or cores of the molecules. For example, \Cref{fig:editing_visualization_analysis} (a) and (b) show \two{} different edits on the same molecule leading to opposite directions in solubility change depending on the text prompt. Replacement of pyridine to a pyrazine core improves the solubility, while insertion of a benzene linkage yields an insoluble molecule. In \Cref{fig:editing_visualization_analysis} (c) and (d), changing an amide linkage to an alkyl amine and an urea results in higher and lower permeability of the edited molecules, respectively. Finally, \Cref{fig:editing_visualization_analysis} (e) and (f) add a butyl ether and a primary amine to the exact position of the molecule, bringing more hydrogen bond acceptors and donors, respectively.\looseness=-1

\textbf{Visual analysis on multi-objective molecule editing.} We further analyze the multi-objective (compositional) property editing. Water solubility improvement and permeability reduction are consistent when introducing polar groups to the molecule and removing lipophilic hydrocarbons, such as an amide or primary amine replacing a methyl or phenyl in \Cref{fig:editing_visualization_analysis} (g). However, higher solubility and permeability are achievable if polar functionalities are removed or reduced in number together with hydrophobic components. For example, in \Cref{fig:editing_visualization_analysis} (h), an amide and a benzene linkage are both removed in the left case, and a \textit{[1,2]oxazolo[5,4-b]pyridine} substituent is replaced by a water-soluble imidazole with a smaller polar surface in the right case.\looseness=-1

\textbf{Case studies on neighborhood searching for patent drug molecules.} In drug discovery, improvement of drug-like properties of lead molecules is crucial for finding drug candidates~\cite{hughes2011principles}. Herein we demonstrate \two{} examples of generating approved drugs from their patented analogs by addressing their property deficiencies based on text prompts. \Cref{fig:editing_visualization_analysis} (i) generates Celecoxib from its amino-substituted derivative~\cite{talley1998substituted}, where the removal of the amino group yields a greater intestinal permeability of the molecule leading to higher bioavailability~\cite{pharmaceutics11080411}. In~\Cref{fig:editing_visualization_analysis} (j), the trimethoxy benzene moiety, an electron-rich arene known to undergo oxidative phase I metabolisms~\cite{guroff1967hydroxylation}, is replaced by a dimethoxy arene in Donepezil by calling for a metabolically stable molecule.\looseness=-1

In summary, we conduct rich experiments on \four{} types and \TotalEditingTaskNum{} text-based molecule editing tasks, where the satisfactory hit ratios of \model{} are superior to baseline methods. Moreover, our editing results can match the expected outcomes based on chemistry domain knowledge. Both quantitative and qualitative results illustrate that \model{} can learn semantically meaningful information useful for domain applications, which encourages us to explore more challenging tasks with \model{} in the future.\looseness=-1

%%%%%%%%%%%%%%%%%%%%%%%%%%%%%%%%%%%%%%%%%%%%%%%%%%
\subsection*{Downstream: Molecular Property Prediction}

\begin{table}[t!]
\centering
\begin{adjustbox}{max width=\textwidth}
\begin{tabular}{l l c c c c c c c c c c c c}
\toprule
& method & BBBP $\uparrow$ & Tox21 $\uparrow$ & ToxCast $\uparrow$ & Sider $\uparrow$ & ClinTox $\uparrow$ & MUV $\uparrow$ & HIV $\uparrow$ & Bace $\uparrow$ & Avg $\uparrow$\\
\midrule
\multirow{4}{*}{SMILES}
& -- (random initialized) & 66.54$\pm$0.95  & 71.18$\pm$0.67  & 61.16$\pm$1.15  & 58.31$\pm$0.78  & 88.11$\pm$0.70  & 62.74$\pm$1.57  & 70.32$\pm$1.51  & 80.02$\pm$1.66  & 69.80\\
& \makecell[l]{MegaMolBART} & 68.89$\pm$0.17  & 73.89$\pm$0.67  & 63.32$\pm$0.79  & 59.52$\pm$1.79  & 78.12$\pm$4.62  & 61.51$\pm$2.75  & 71.04$\pm$1.70  & \textbf{82.46$\pm$0.84}  & 69.84\\
& KV-PLM & 70.50$\pm$0.54  & 72.12$\pm$1.02  & 55.03$\pm$1.65  & 59.83$\pm$0.56  & \textbf{89.17$\pm$2.73}  & 54.63$\pm$4.81  & 65.40$\pm$1.69  & 78.50$\pm$2.73  & 68.15\\
& \model\, & \textbf{70.75$\pm$1.90}  & \textbf{75.71$\pm$0.89}  & \textbf{65.17$\pm$0.37}  & \textbf{63.70$\pm$0.81}  & 86.60$\pm$2.28  & \textbf{65.69$\pm$1.46}  & \textbf{77.02$\pm$0.44}  & 81.99$\pm$0.41  & \textbf{73.33}\\
\midrule
\multirow{7}{*}{Graph}
& -- (random initialized) & 63.90$\pm$2.25  & 75.06$\pm$0.24  & 64.64$\pm$0.76  & 56.63$\pm$2.26  & 79.86$\pm$7.23  & 70.43$\pm$1.83  & 76.23$\pm$0.80  & 73.14$\pm$5.28  & 69.99\\
& AttrMask & 67.79$\pm$2.60 & 75.00$\pm$0.20 & 63.57$\pm$0.81 & 58.05$\pm$1.17 & 75.44$\pm$8.75 & 73.76$\pm$1.22 & 75.44$\pm$0.45 & 80.28$\pm$0.04 & 71.17 \\
& ContextPred & 63.13$\pm$3.48 & 74.29$\pm$0.23 & 61.58$\pm$0.50 & 60.26$\pm$0.77 & 80.34$\pm$3.79 & 71.36$\pm$1.44 & 70.67$\pm$3.56 & 78.75$\pm$0.35 & 70.05 \\
& InfoGraph & 64.84$\pm$0.55 & 76.24$\pm$0.37 & 62.68$\pm$0.65 & 59.15$\pm$0.63 & 76.51$\pm$7.83 & 72.97$\pm$3.61 & 70.20$\pm$2.41 & 77.64$\pm$2.04 & 70.03 \\
& MolCLR & 67.79$\pm$0.52 & 75.55$\pm$0.43 & 64.58$\pm$0.07 & 58.66$\pm$0.12 & 84.22$\pm$1.47 & 72.76$\pm$0.73 & 75.88$\pm$0.24 & 71.14$\pm$1.21 & 71.32 \\
& GraphMVP & 68.11$\pm$1.36  & \textbf{77.06$\pm$0.35}  & \textbf{65.11$\pm$0.27}  & 60.64$\pm$0.13  & 84.46$\pm$3.10  & \textbf{74.38$\pm$2.00}  & \textbf{77.74$\pm$2.51}  & 80.48$\pm$2.68  & 73.50\\
& \model{} & \textbf{69.98$\pm$0.52}  & 76.91$\pm$0.51  & 65.05$\pm$0.39  & \textbf{60.96$\pm$1.05}  & \textbf{92.53$\pm$1.07}  & 73.40$\pm$2.90  & 76.93$\pm$1.84  & \textbf{80.77$\pm$1.34}  & \textbf{74.57}\\
\bottomrule
\end{tabular}
\end{adjustbox}
\vspace{-2ex}
\caption{
Results on \eight{} MoleculeNet binary classification tasks. Mean and standard deviation of test ROC-AUC on three random seeds are reported.
}
\label{tab:molecular_property_prediction}
\end{table}

\textbf{Experiments.} One advantage for \model{} is that the pretrained chemical structure representation shares information with the external domain knowledge, and such implicit bias can be beneficial for the property prediction tasks. Similar to previous works on molecule pretraining~\cite{hu2019strategies,liu2021pre}, we adopt the MoleculeNet benchmark~\cite{wu2018moleculenet}. It contains \eight{} single-modal binary classification datasets to evaluate the expressiveness of the pretrained molecule representation methods. The evaluation metric is the area under the receiver operating characteristic curve (ROC-AUC)~\cite{bradley1997use}.\looseness=-1

\textbf{Baselines.} We consider \two{} types of chemical structures, the SMILES string and the molecular graph. For the SMILES string, we take three baselines: the \textit{randomly initialized} models and \two{} pretrained language models (\textit{MegaMolBART}~\cite{irwin2022chemformer} and \textit{KV-PLM}~\cite{zeng2022deep}). For the molecular graph, in addition to the \textit{random initialization}, we consider \five{} pretraining-based methods as baselines: \textit{AttrMasking}~\cite{hu2019strategies}, \textit{ContextPred}~\cite{hu2019strategies}, \textit{InfoGraph}~\cite{sun2019infograph}, \textit{MolCLR}~\cite{wang2021molclr}, and \textit{GraphMVP}~\cite{liu2018n}.\looseness=-1

\textbf{Results.} As shown in~\Cref{tab:molecular_property_prediction}, we first observe that pretraining-based methods improve the overall classification accuracy compared to the randomly-initialized ones. \model{} on the SMILES string has consistent improvements on \six{} out of \eight{} tasks compared to the \three{} baselines. \model{} on the molecular graph performs the best on \four{} out of \eight{} tasks, while it performs comparably to the best baselines in other \four{} tasks. In both cases, the overall performances ({\ie}, taking an average across all \eight{} tasks) of \model{} are the best among all the methods.\looseness=-1

%%%%%%%%%%%%%%%%%%%%%%%%%%%%%%%%%%%%%%%%%%%%%%%%%%
\section*{Discussion}
In this work, we have presented a multi-modal model, \model{}, to illustrate the effectiveness of incorporating textual descriptions for molecule representation learning. On \two{} newly proposed zero-shot tasks and \one{} standard property prediction benchmark, we confirmed consistently improved performance of \model{} compared to the existing methods. Additionally, we observed that \model{} can retrieve novel drug-target relations and successfully modify molecule substructures to gain the desired properties. These functionalities may accelerate various downstream drug discovery practices, such as re-purposing and multi-objective lead optimization. Furthermore, the outcomes of such downstream tasks have been found to be consistent with the feedback from chemistry experts, reflecting the domain knowledge exploration ability of \model{}.\looseness=-1

One limitation of this work is data insufficiency. Although \DatasetName{} is $28\times$ larger than the dataset used in existing works, it can be further improved and may require support from the entire community in the future. The second bottleneck of this work is the expressiveness of chemical structure models, including the SMILES encoder, the GNN encoder, and the SMILES-based molecule generative model. The development of more expressive architectures is perpendicular to this work and can be feasibly adapted to our multi-modal pretraining framework.\looseness=-1

For future directions, we would like to extend \model{} from cheminformatics to bioinformatics tasks with richer textual information. This enables us to consider structure-based drug design problems such as protein-ligand binding and fragment design. Besides, the 3D geometric information has become more important for small molecules and polymers and can thus be merged into our foundation model.  Last but not least, the tokenization of the textual description may require extra effort. Certain tasks possess rich terminologies ({\eg}, the ATC codes in DrugBank-ATC), and the overall performance is affected accordingly. Such fundamental problems should be handled carefully.

%%%%%%%%%%%%%%%%%%%%%%%%%%%%%%%%%%%%%%%%%%%%%%%%%%
\section*{Methods}
This section briefly describes certain modules in both pretraining and downstream tasks. Detailed specifications, such as dataset construction, model architectures, and hyperparameters, can be found in Supplementary A.\looseness=-1

\subsection*{\model{} Pretraining}
\textbf{Dataset construction.} For the structure-text pretraining, we consider the PubChem database~\cite{kim2021pubchem} as the data source. PubChem includes 112M molecules, which is one of the largest public databases for molecules. The PubChem database has many fields, and previous work~\cite{zeng2022deep} uses the synonym field to match with an academic paper corpus~\cite{lo2019s2orc}, resulting in a dataset with 10K structure-text pairs. Meanwhile, the PubChem database has another field called ``string'' with more comprehensive and versatile molecule annotations. We utilize this field to construct a large-scale dataset called \DatasetName{}, consisting of 250K molecules and 281K structure-text pairs.\looseness=-1

In addition, even though \DatasetName{} is the largest dataset with textual descriptions, its dataset size is comparatively small compared to the peers from other domains ({\eg}, 400M in the vision-language domain~\cite{radford2021learning}). To mitigate such a data insufficiency issue, we adopt the pretrained models from existing checkpoints and then conduct the end-to-end pretraining, as will be discussed next.\looseness=-1

\textbf{Chemical structure branch $f_c$.} This work considers two types of chemical structures: the SMILES string views the molecule as a sequence, and the 2D molecular graph takes the atoms and bonds as the nodes and edges, respectively. Then, based on the chemical structures, we apply a deep learning encoder $f_c$ to get a latent vector as molecule representation. Specifically, for the SMILES string, we take the encoder from MegaMolBART~\cite{irwin2022chemformer}, which is pretrained on 500M molecules from ZINC database~\cite{sterling2015zinc}. For the molecular graph, we take a pretrained graph isomorphism network (GIN)~\cite{xu2018powerful} using GraphMVP pretraining~\cite{liu2021pre}. GraphMVP is doing a multi-view pretraining between the 2D topologies and 3D geometries on 250K conformations from GEOM dataset~\cite{axelrod2022geom}. Thus, though we are not explicitly utilizing the 3D geometries, the state-of-the-art pretrained GIN models can implicitly encode such information.\looseness=-1

\textbf{Textual description branch $f_t$.} The textual description branch provides a high-level description of the molecule's functionality. We can view this branch as domain knowledge to strengthen the molecule representation. Such domain knowledge is in the form of natural language, and we use the BERT model~\cite{devlin2018bert} as the text encoder $f_t$. We further adapt the pretrained SciBERT~\cite{Beltagy2019SciBERT}, which was pretrained on the textual data from the chemical and biological domain.\looseness=-1

\textbf{Contrastive pretraining.} For the \model{} pretraining, we adopt the contrastive learning strategy, {\eg}, EBM-NCE~\cite{liu2021pre} and InfoNCE~\cite{oord2018representation}. EBM-NCE and InfoNCE align the structure-text pairs for the same molecule and contrast the pairs for different molecules simultaneously. We consider the selection of contrastive pretraining methods as one important hyperparameter. The objectives for EBM-NCE and InfoNCE are
\begin{equation}
\label{eq:MolCLIP_contrastive_pretraining}
\fontsize{8}{3}\selectfont
\begin{aligned}
\mathcal{L}_{\text{EBM-NCE}} & = -\frac{1}{2} \Big(
\mathbb{E}_{\vx_c, \vx_t} \big[\log \sigma (E(\vx_c, \vx_t) \big] + \mathbb{E}_{\vx_c, \vx_t'} \big[ \log (1 - \sigma (E(\vx_c, \vx_t'))\big] \Big) + \mathbb{E}_{\vx_c, \vx_t} \big[\log \sigma (E(\vx_c, \vx_t) \big] + \mathbb{E}_{\vx_c',\vx_t} \big[ \log (1 - \sigma (E(\vx_c', \vx_t))\big]
\Big),\\
\mathcal{L}_{\text{InfoNCE}} & = -\frac{1}{2} \mathbb{E}_{\vx_c, \vx_t} \Big[
\log \frac{\exp ( E(\vx_c, \vx_t)) }{\exp ( E(\vx_c, \vx_t)) + \sum\limits_{\vx_{t'}} \exp ( E(\vx_c, \vx_{t'}))} + 
\log \frac{\exp ( E(\vx_c, \vx_t)) }{\exp ( E(\vx_c, \vx_t)) + \sum\limits_{\vx_{c'}} \exp ( E(\vx_{c'}, \vx_{t}))} \Big],
\end{aligned}
\end{equation}
where $\sigma$ is the sigmoid activation function, $\vx_c$ and $\vx_t$ form the structure-text pair for each molecule, and $\vx_{c'}$ and $\vx_{t'}$ are the negative samples randomly sampled from the noise distribution, which we use the empirical data distribution. $E(\cdot)$ is the energy function with a flexible formulation, and we use the dot product on the jointly learned space, {\ie}, $E(\vx_c, \vx_t) = \langle p_c \circ f_c (\vx_c), p_t \circ f_t (\vx_t) \rangle$, where $\circ$ is the function composition.

%%%%%%%%%%%%%%%%%%%%%%%%%%%%%%%%%%%%%%%%%%%%%%%%%%
\subsection*{Downstream: Zero-shot Structure-text Retrieval}
Given a chemical structure and $T$ textual descriptions, the retrieval task is to select the textual description with the highest similarity to the chemical structure (or vice versa) based on a score calculated on the joint representation space. This is appealing for specific drug discovery tasks, such as drug re-purposing or indication expansion~\cite{zeng2022deep,aggarwal2010targeted}. We highlight that pretrained models are used for retrieval in the zero-shot setting, {\ie}, without model optimization for this retrieval task. Existing works~\cite{guney2017reproducible} have witnessed the potential issue that utilizing the chemical structure alone is not sufficient, while \model{} enables a novel perspective by adopting the textual description with the utilization of the high-level functionality of molecules.\looseness=-1

In such a zero-shot task setting, all the encoders ($f_c, f_t$) and projectors ($p_c, p_t$) are pretrained from \model{}, and stay frozen in this downstream task. An example of the retrieval task of setting (1) is
\begin{equation}
\small
\text{Retrieval}(\vx_c) = \arg \max_{\tilde\vx_t} \Big\{ \big\langle p_c \circ f_c (\vx_c), p_t \circ f_t (\tilde\vx_t) \big\rangle \Big| \tilde\vx_t \in \text{T textual descriptions} \Big\}.
\end{equation}

%%%%%%%%%%%%%%%%%%%%%%%%%%%%%%%%%%%%%%%%%%%%%%%%%%
\subsection*{Downstream: Zero-shot Text-based Molecule Editing}
The objective of the molecule editing task is to modify the chemical structure of molecules such as functional group change~\cite{ertl2020most} and scaffold hopping~\cite{bohm2004scaffold,hu2017recent}. Traditional methods for molecule editing highly rely on domain experts and could be subjective or biased~\cite{drews2000drug,gomez2018decision}. ML methods have provided an alternative strategy to solve this issue. Given a fixed pretrained molecule generative model (encoder $f_g$ and decoder $h_g$), the ML editing methods learn a semantically meaningful direction on the latent representation (or latent code) space. The decoder $h_g$ then generates output molecules with the desired properties by moving along the direction. In \model{}, with the pretrained joint representation space, we can accomplish this task by injecting the textual description in a zero-shot manner. As shown in~\Cref{fig:editing_pipeline} (a, b), we need \two{} phases. The first phase is space alignment, where we train an adaptor module to align the representation space of the generative model to the joint representation space of \model{}. The second phase is latent optimization, where we directly learn the latent code using \two{} similarity scores as the objective function. Finally, decoding the optimized latent code can lead to the output molecules. Notice that during this editing process, both the \model{} ($f_c,p_c,f_t,p_t$) and a pretrained molecule generative model ($f_g, h_g$) are frozen. 

\textbf{Phase 1: space alignment.} In this phase, the goal is to learn an adaptor module to align the representation space of the generative model to the joint representation space of \model{}. Following the Gaussian distribution, the objective function is\looseness=-1
\begin{equation}
\label{eq:space_alignment}
\small
\begin{aligned}
\mathcal{L} = \| m_{g2f} \circ f_g (\vx_c) - p_c \circ f_c (\vx_c) \|^2,
\end{aligned}
\end{equation}
where $\circ$ is the function composition function, and $m_{g2f}$ is the adaptor module optimized to align the two latent spaces.\looseness=-1

\textbf{Phase 2: latent optimization.} In this phase, given an input molecule $\vx_{c,\text{in}}$ and a text prompt $\vx_t$, the goal is to optimize a latent code $w$ directly. The optimal $w$ should be close to the representations of $\vx_{c,\text{in}}$ and $\vx_t$ simultaneously, as:
\begin{equation} \label{eq:latent_optimization}
\small
\begin{aligned}
w = \argmin_{w \in \mathcal{W}} \Big( - \mathcal{L}_{\text{cosine-sim}}\big(m_{g2f}(w), p_t \circ f_t(\vx_t) \big) + \lambda \cdot \mathcal{L}_{l_2} \big( w, f_g(\vx_{c,\text{in}})\big) \Big),
\end{aligned}
\end{equation}
where $\mathcal{W}$ is the latent code space, $\mathcal{L}_{\text{cosine-sim}}$ is the cosine-similarity, and $\mathcal{L}_{l_2}$ is the $l_2$ distance, and $\lambda$ is a coefficient to balance these two similarity terms. Finally, after we optimize the latent code $w$, we will do decoding using the decoder from the pretrained generative model to obtain the output molecule: $\vx_{c,\text{out}} = h_g(w)$.\looseness=-1

\textbf{Evaluation.} The evaluation metric is the satisfactory hit ratio. Suppose we have an input molecule $\vx_{c,\text{in}}$ and a text prompt $\vx_t$, the editing algorithm will generate an output molecule $\vx_{c,\text{out}}$. Then we use the hit ratio to measure if the output molecule can satisfy the conditions as indicated in the text prompt.\looseness=-1
\begin{equation} \label{eq:language_guided_editing_eval_metric_hit_ratio}
\small{
\begin{aligned}
\text{hit}(\vx_{c,\text{in}}, \vx_t) & = \begin{cases}
1, & \exists \lambda \, \text{, s.t. $\vx_{c,\text{out}}$ = $h_g$ (w; $\lambda$) $\wedge$ satisfy } (\vx_{c,\text{in}}, \vx_{c,\text{out}}, \vx_t) \\
0, & \text{otherwise}
\end{cases},
\quad\quad\quad
\text{hit}(t) = \frac{\sum_{i=1}^N \text{hit}(\vx_{c,\text{in}}^i, \vx_t)}{N},
\end{aligned}
}
\end{equation}
where $N$ is the total number of editing outputs, and $\text{satisfy}(\cdot)$ is the satisfaction condition. It is task-specific, and we list the five key points below. (1) For single-objective property-based editing, we use the logarithm of partition coefficient (LogP), quantitative estimate of drug-likeness (QED), and topological polar surface area (tPSA) as the proxies to measure the molecule solubility~\cite{doi:10.1021/cr60274a001}, drug likeness~\cite{bickerton2012quantifying}, and permeability~\cite{doi:10.1021/jm000942e}, respectively. The count of hydrogen bond acceptors (HBA) and hydrogen bond donors (HBD) are calculated explicitly. It will be a successful hit once the measurement difference between the input molecule and output molecule is above a certain threshold $\Delta$. (2) For multiple-objective property-based editing, we feed in a text prompt describing multiple properties' composition. The $\Delta$ is composed of the threshold on each individual property, and a successful hit needs to satisfy all the properties simultaneously. (3) For binding-affinity-based editing, we take the ground-truth data from ChEMBL to train a binary classifier and test if the output molecules have higher confidence than the input molecules, and $\Delta$ is fixed to 0. (4) For drug relevance editing, we use Tanimoto similarity to quantify the structural similarity~\cite{butina_1999}. It will be a hit if the similarity score between the output molecule and target drug is higher than the similarity between the input molecule and target drug by a threshold $\Delta$. (5) Besides, the choice of satisfactory threshold $\Delta$ is also task-specific, and the higher the values are, the stricter the satisfaction condition is. The details of the threshold values can be found in Supplementary D.\looseness=-1

%%%%%%%%%%%%%%%%%%%%%%%%%%%%%%%%%%%%%%%%%%%%%%%%%%
\subsection*{Downstream: Molecular Property Prediction}
For modeling, we take the pretrained encoder $f_c$ and add a prediction head $h_c$ to predict a categorical-valued or scalar-valued molecular property such as binding affinity or toxicity. Both $f_c$ and $h_c$ are optimized to fit the target property, {\ie}, in a fine-tuning manner~\cite{hu2019strategies,liu2021pre}.

%%%%%%%%%%%%%%%%%%%%%%%%%%%%%%%%%%%%%%%%%%%%%%%%%%
\section*{Data Availability}
All the datasets are provided at \href{https://huggingface.co/datasets/chao1224/MoleculeSTM/tree/main}{this Hugging Face link}. Specifically for the release of \DatasetName{}, we encountered a big challenge regarding the textual data license. As confirmed with the PubChem group, performing research on these data does not violate their license; however, PubChem does not possess the license for the textual data, which necessitates an extensive evaluation of the license for each of the 280 structure-text pairs in PubChemSTM. This has hindered the release of PubChemSTM. Nevertheless, we have (1) described the detailed preprocessing steps in Supplementary A.1, (2) provided the \href{https://huggingface.co/datasets/chao1224/MoleculeSTM/blob/main/PubChemSTM_data/raw/CID2SMILES.csv}{molecules with CID file} in \DatasetName{} and (3) have also provided the detailed \href{https://github.com/chao1224/MoleculeSTM/tree/main/preprocessing/PubChemSTM}{preprocessing scripts}. By utilizing these scripts, users can easily reconstruct the \DatasetName{} dataset.

%%%%%%%%%%%%%%%%%%%%%%%%%%%%%%%%%%%%%%%%%%%%%%%%%%
\section*{Code Availability}
The source code can be found at this \href{https://github.com/chao1224/MoleculeSTM/tree/main}{GitHub repository} and Zenodo~\cite{shengchaoliu_code}. The scripts for pretraining and three downstream tasks are provided \href{https://github.com/chao1224/MoleculeSTM/tree/main/scripts}{here}. The checkpoints of the pretrained models are provided at this \href{https://huggingface.co/chao1224/MoleculeSTM/tree/main}{Hugging Face link}. Beyond the methods described so far, to help users try our MoleculeSTM model, this release includes \href{https://github.com/chao1224/MoleculeSTM}{demos in notebooks}. Furthermore, users can customize their own datasets by checking the \href{https://github.com/chao1224/MoleculeSTM/tree/main/MoleculeSTM/datasets}{datasets folder}.

%%%%%%%%%%%%%%%%%%%%%%%%%%%%%%%%%%%%%%%%%%%%%%%%%%
\section*{Acknowledgements}
This work was done during Shengchao Liu's internship at NVIDIA Research. The authors would like to thank the insightful comments from Michelle Lynn Gill, Abe Stern, and other team members from AIAlgo and Clara team at NVIDIA. The authors would also like to thank the kind help from Teresa Dierks, Evan Bolton, Paul Thiessen, et al from PubChem for confirming the PubChem license.

%%%%%%%%%%%%%%%%%%%%%%%%%%%%%%%%%%%%%%%%%%%%%%%%%%
\section*{Author Contributions Statement}
S.L., W.N., C.W., Z.Q., C.X., and A.A. conceived and designed the experiments. S.L. performed the experiments. S.L. and C.W. analyzed the data. S.L., C.W., and J.L. contributed analysis tools. S.L., W.N., C.W., J.L., Z.Q., L.L., J. T., C.X., and A.A. wrote the paper. J. T., C.X., and A.A. contributed equally to advising this project.

%%%%%%%%%%%%%%%%%%%%%%%%%%%%%%%%%%%%%%%%%%%%%%%%%%
\section*{Competing Interests Statement}
The authors declare no competing interests.

{
\renewcommand*{\bibfont}{\small}
\printbibliography[keyword={main},title={References}]
}

\clearpage
\newpage
\appendix

{\parindent0pt
\Large \textsf{\bf Supplementary Information}
}

\begin{refsection}
%%%%%%%%%%%%%%%%%%%%%%%%%%%%%%%%%%%%%%%%%%%%%%%%%%
\section{Pretraining} \label{app:sec:pretraining}

\subsection{\DatasetName{} Construction}
We construct a chemical structure-text pair dataset called \DatasetName{}, which is extracted from the PubChem database~\cite{kim2021pubchem}. Below we explain the key steps of the dataset construction.

\begin{enumerate}[noitemsep,topsep=0pt]
    \item We use the \href{https://pubchemdocs.ncbi.nlm.nih.gov/pug-view}{PUG View} (a REST-style web service) to download the textual descriptions of molecules. It has in total of 290 pages, and each page is downloaded in XML format. For reference, an example page (the first page) can be found \href{https://pubchem.ncbi.nlm.nih.gov/rest/pug_view/annotations/heading/json?heading_type=Compound&heading=Record+Description&page=1}{here}. There is a ``string'' field in the XML data, and we treat it as the textual descriptions for molecules. After construction, we have 250K molecules (with unique PubChem ID) and 281K chemical structure-text pairs. Notice that each molecule can have multiple annotations from different resources.
    \begin{itemize}[noitemsep,topsep=0pt]
        \item Most of the molecule annotations start with the common name or the International Union of Pure and Applied Chemistry (IUPAC) name. We can either use the raw description (with a common name or IUPAC name) or replace it with the text template ({\eg}, ``This molecule is ...''). \item Thus, we construct \two{} versions of \DatasetName{} datasets, \DatasetName{}-raw and \DatasetName{}-extracted, corresponding to using the raw annotation or replacing the molecule name with the text prompt, respectively. These \two{} versions of \DatasetName{} share the molecules, except for the molecule names.
    \end{itemize}
    \item We download the 326 SDF files from the PubChem \href{https://ftp.ncbi.nlm.nih.gov/pubchem/Compound/CURRENT-Full/SDF/}{FTP service}. Each SDF file contains the structural information ({\eg}, the SMILES string and molecular graph) for a batch of molecules.
    \item We match the annotation and chemical structure for each molecule from the previous \two{} steps using the PubChem ID, and most of the molecules from the first step contain the corresponding chemical structures from the SDF files. In specific, only 12 molecules failed to find the valid SMILES from SDF files, and we ignore these molecules.
    \item Ultimately, following the above \three{} steps will lead to a structure-text pair dataset with 281K pairs and 250K unique molecules. Note that the PubChem database~\cite{kim2021pubchem} is updated online frequently, and the above numbers are collected in March 2022.
\end{enumerate}

\paragraph{Pre-processing Details}
There is one field in the PubChem database called ``name'', which includes either the common name or the IUPAC name for each molecule. Notice that the tokenization on IUPAC is nontrivial. Thus we carry out \two{} versions to test its effect, {\ie}, the \DatasetName{}-raw and \DatasetName{}-extracted. We find that there exist several patterns of textual descriptions in \DatasetName{}-raw, which are further utilized to extract the cleaner version of molecule description as in PubChem-extract. A detailed illustration is given below: 
\begin{itemize}[noitemsep,topsep=0pt]
    \item The most common pattern is that the molecule annotation starts with ``XXX (name) is / are / was / were / appears / occurs / stands for / belongs to / exits ...''. We manually extract this to obtain most of the molecule names and replace them with "This molecule ..." or "These molecules ...".
    \item \textbf{Extra word "Pure".} Some molecule annotations start with ``Pure xxx ...'' and we remove the word ``Pure''.
    \item \textbf{Typos.} For example, the "Mercurycombines ..." should be "Mercury combines ...".
\end{itemize}

\paragraph{Dataset Examples}
We provide \four{} examples of the \DatasetName{}-raw and \DatasetName{}-extracted in~\Cref{tab:sec:dataset_examples}.

\begin{table}[htb]
\caption{\small Examples on \DatasetName{}. Here for the chemical structure, we only list the SMILES string, since the 2D topology graph can be obtained using the RDKit package.}
\label{tab:sec:dataset_examples}
\vspace{-2ex}
\centering
    \begin{adjustbox}{max width=\textwidth}
    \begin{tabular}{p{0.5\textwidth} p{0.5\textwidth}}
    \toprule
    \multicolumn{1}{c}{\DatasetName{}-raw} & \multicolumn{1}{c}{\DatasetName{}-extracted}\\
    \midrule
    \multicolumn{2}{c}{SMILES: c1ccccc1}\\ % PubChemID: 241, 
    Benzene is a colorless liquid with a sweet odor. It evaporates into the air very quickly and dissolves slightly in water.
    & \textit{This molecule is} a colorless liquid with a sweet odor. It evaporates into the air very quickly and dissolves slightly in water.\\
    \toprule
    \multicolumn{2}{c}{SMILES: Oc1ccccc1}\\ % PubChemID: 996, 
    Phenol is both a manufactured chemical and a natural substance. It is a colorless-to-white solid when pure.
    & \textit{This molecule is} both a manufactured chemical and a natural substance. It is a colorless-to-white solid when pure.\\
    \toprule
    \multicolumn{2}{c}{SMILES: CC(=O)Oc1ccccc1C(=O)O}\\ % PubChemID: 2244, 
    Acetylsalicylic acid appears as odorless white crystals or crystalline powder with a slightly bitter taste.
    & \textit{This molecule appears} as odorless white crystals or crystalline powder with a slightly bitter taste.\\
    \toprule
    \multicolumn{2}{c}{SMILES: CC1(C)SC2C(NC(=O)Cc3ccccc3)C(=O)N2C1C(=O)O}\\ % PubChemID: 5904, 
    Benzylpenicillin is a penicillin in which the substituent at position 6 of the penam ring is a phenylacetamido group. It has a role as an antibacterial drug, an epitope and a drug allergen.
    & \textit{This molecule is} a penicillin in which the substituent at position 6 of the penam ring is a phenylacetamido group. It has a role as an antibacterial drug, an epitope, and a drug allergen.\\
    \bottomrule
    \end{tabular}
    \end{adjustbox}
\end{table}

\paragraph{Reproducibility}
Because the PubChem database~\cite{kim2021pubchem} has been updated online frequently, so we provide all the pre-processed datasets used in this work for reproducibility. In addition, the source codes for the above steps are also provided for future usage.\looseness=-1

\paragraph{Comparison}
As mentioned, we adopt a pretrained SciBERT model~\cite{Beltagy2019SciBERT} and continue training on \DatasetName{}. SciBERT is a BERT model specifically trained for scientific discovery. It randomly samples 1.14M papers from Semantic Scholar~\cite{ammar2018construction}, where around 18\% papers are from the computer science domain and 82\% papers are from the broad biomedical domain. Its corpus has 3.17B tokens and the vocabulary size is 31K. Besides, SciBERT was trained on the full paper, not just the abstract. One potential issue is the vocabulary shift from the Semantic Scholar to \DatasetName{}. Although we adapt the pretrained checkpoints from SciBERT (together with its vocabulary) in this work, we still want to carefully examine the vocabulary for the textual data.\looseness=-1

\begin{table}[htb!]
\caption{
\small
The vocabulary comparison.
}
\label{tab:app:vocabulary_size_comparison}
\centering
\vspace{-2ex}
\begin{adjustbox}{max width=\textwidth}
\begin{tabular}{l l r r}
    \toprule
    Data Source & Tokenization Method & size of vocabulary & overlap with SciBERT \\
    \midrule
    Semantic Scholar (used in SciBERT) & SciBERT tokenizer & 31,090 & - \\
    \midrule
    \multirow{3}{*}{\DatasetName{}-raw}
     & white space & 315,704 & 7,635\\
     & spaCy & 114,976 & 719\\
     & SciBERT tokenizer & 18,320 & 18,320\\
    \midrule
    \multirow{3}{*}{\DatasetName{}-extract}
     & white space & 100,877 & 7,562\\
     & spaCy & 27,519 & 691\\
     & SciBERT tokenizer & 17,442 & 17,442\\
    \bottomrule
\end{tabular}
\end{adjustbox}
\end{table}

In~\Cref{tab:app:vocabulary_size_comparison}, we list the vocabulary size of \DatasetName{}-raw and \DatasetName{}-extract with \three{} tokenization methods: using white space, spaCy~\cite{honnibal2020boyd}, and the SciBERT tokenizer. We can observe that the difference between \DatasetName{}-raw and \DatasetName{}-extract using the SciBERT tokenizer is quite small, compared to the ones using white space and spaCy. Thus, we want to claim that vocabulary is also an important factor, and the SciBERT tokenizer has shown quite a stable tokenization effect. In the future, more comprehensive tokenization and vocabulary are required to push forwards this research line, {\ie}, to enable the large language model for drug discovery. But it is beyond the scope of this paper and requires efforts from the entire community.\looseness=-1

\subsection{Architecture Details}
We have two branches, the chemical structure branch $f_c$ and the textual description branch $f_t$.

\paragraph{Chemical structure branch $f_c$}
This work considers two types of chemical structures: the SMILES string views the molecule as a sequence and the 2D molecular graph takes the atoms and bonds as the nodes and edges, respectively. Then based on the chemical structures, we apply a deep learning encoder $f_c$ to get a latent vector as molecule representation. Specifically, for the SMILES string, we take the encoder from MegaMolBART~\cite{irwin2022chemformer}, which is pretrained on 500M molecules from ZINC database~\cite{sterling2015zinc}. For the molecular graph, we take a pretrained graph isomorphism network (GIN)~\cite{xu2018powerful} using GraphMVP pretraining~\cite{liu2021pre}. GraphMVP is doing a multi-view pretraining between the 2D topologies and 3D geometries on 250K conformations from GEOM dataset~\cite{axelrod2022geom}. Thus, though we are not explicitly utilizing the 3D geometries, the state-of-the-art pretrained GIN models can implicitly encode such information.

\paragraph{Textual description branch $f_t$}
The textual description branch provides a high-level description of the molecule's functionality. We can view this branch as domain knowledge to strengthen the molecule representation. Such domain knowledge is in the form of natural language, and we use the BERT model~\cite{devlin2018bert} as the text encoder $f_t$. We further adapt the pretrained SciBERT~\cite{Beltagy2019SciBERT}, which was pretrained on the textual data from the chemical and biological domain.

\begin{table}[htb]
\centering
\setlength{\tabcolsep}{5pt}
\fontsize{9}{9}\selectfont
\caption{
\small
Model specifications. \# parameters in each model.
}
\vspace{-2ex}
\begin{adjustbox}{max width=\textwidth}
\begin{tabular}{c l r}
\toprule
Branch & Model & \# parameters\\
\midrule
\multirow{2}{*}{Chemical structure}
& MegaMolBART & 10,010,635\\
& GIN & 1,885,206\\
\midrule
\multirow{1}{*}{Textual description}
& SciBERT & 109,918,464\\
\bottomrule
\end{tabular}
\end{adjustbox}
\end{table}

\subsection{Pretraining Details}
\paragraph{Pretraining Objective}
For the \model{} pretraining, we apply contrastive learning. More concretely, we choose one of the EBM-NCE~\cite{liu2021pre} and InfoNCE~\cite{oord2018representation}. Both are essentially doing the same thing, yet EBM-NCE has been found to be more effective for graph-data~\cite{liu2021pre,hassani2020contrastive}. The objective for EBM-NCE is:
\begin{equation}
\fontsize{7.6}{0}\selectfont
\small
\begin{aligned}
\mathcal{L} & = -\frac{1}{2} \Big(
\mathbb{E}_{\vx_c, \vx_t} \big[\log \sigma (E(\vx_c, \vx_t) \big] + \mathbb{E}_{\vx_c, \vx_t'} \big[ \log (1 - \sigma (E(\vx_c, \vx_t'))\big] \Big) -\frac{1}{2} \Big( \mathbb{E}_{\vx_c, \vx_t} \big[\log \sigma (E(\vx_c, \vx_t) \big] + \mathbb{E}_{\vx_c',\vx_t} \big[ \log (1 - \sigma (E(\vx_c', \vx_t))\big]
\Big),
\end{aligned}
\end{equation}
where $\vx_c$ and $\vx_t$ form the structure-text pair for each molecule, and $\vx_{c'}$ and $\vx_{t'}$ are the negative samples randomly sampled from the noise distribution, which we use the empirical data distribution. $E(\cdot)$ is the energy function with a flexible formulation, and we use the dot product on the jointly learned space, {\ie}, $E(\vx_c, \vx_t) = \langle p_c \circ f_c (\vx_c), p_t \circ f_t (\vx_t) \rangle$. Similarly, we have the objective for InfoNCE as:
\begin{equation}
\fontsize{7.6}{0}\selectfont
\small
\begin{aligned}
\mathcal{L} & = -\frac{1}{2} \mathbb{E} \big[
\log \frac{\exp ( E(\vx_c, \vx_t)) }{\exp ( E(\vx_c, \vx_t)) + \sum\limits_{\vx_{t'}} \exp ( E(\vx_c, \vx_{t'}))} + 
\log \frac{\exp ( E(\vx_c, \vx_t)) }{\exp ( E(\vx_c, \vx_t)) + \sum\limits_{\vx_{c'}} \exp ( E(\vx_{c'}, \vx_{t}))} \big].
\end{aligned}
\end{equation}

\paragraph{Hyperparameters}
We list the key hyperparameters used for \model{} pretraining with the SMILES string and 2D molecular graph as inputs, respectively.
\begin{table}[htb]
\centering
\setlength{\tabcolsep}{5pt}
\fontsize{9}{9}\selectfont
\caption{
\small
Hyperparameter specifications for \model{} pretraining.
}
\vspace{-2ex}
\begin{adjustbox}{max width=\textwidth}
\begin{tabular}{l l l}
\toprule
Input & Hyperparameter & Value\\
\midrule
\multirow{4}{*}{SMILES string}
& epochs & \{32\} \\
& learning rate for text branch & \{1e-4\} \\
& learning rate for chemical structure branch & \{1e-5, 3e-5\} \\
& objective function & \{ EBM-NCE, InfoNCE\}\\
\midrule
\multirow{4}{*}{2D molecular graph}
& epochs & \{32\} \\
& learning rate for text branch & \{1e-4\} \\
& learning rate for chemical structure branch & \{1e-5, 3e-5\} \\
& objective function & \{ EBM-NCE, InfoNCE\}\\
\bottomrule
\end{tabular}
\end{adjustbox}
\end{table}

\paragraph{Running time}
We list the running time of \model{} with the SMILES string and 2D molecular graph as inputs, respectively.
\begin{table}[htb]
\centering
\setlength{\tabcolsep}{5pt}
\fontsize{9}{9}\selectfont
\caption{
\small
Running time for \model{} pretraining.
}
\vspace{-2ex}
\begin{adjustbox}{max width=\textwidth}
\begin{tabular}{l l}
\toprule
Input & Running Time\\
\midrule
SMILES string & ~44min / epoch\\
2D molecular graph & ~42min / epoch\\
\bottomrule
\end{tabular}
\end{adjustbox}
\end{table}

\clearpage
\section{Design Principles for Downstream Tasks} \label{sec:app:downstream}
In this section, we discuss the key principles when designing downstream tasks.

\paragraph{Applicable Evaluation}
One of the biggest differences between the foundation model in the vision-language domain and our \model{} can be reflected in the evaluation. Most of the vision and language tasks can be viewed as art problems, {\ie}, there does not exist a standard and exact solution that is applicable for evaluation. For instance, we can detect if the image is "a horse riding an astronaut" or "a panda making latte art"~\cite{saharia2022photorealistic}, but only visually not computationally, which prevents large-scale evaluation. This is not the case for drug discovery, because it is a scientific task, where the results ({\eg}, properties of the output molecules in the editing task) can be evaluated exactly, either in vitro or in silico. Following this, the physical experiments are usually expensive and long-lasting, so in this work, we want to focus on tasks that are computationally feasible for evaluation.

\paragraph{Fuzzy Matching}
Specifically for the molecule editing task, the text prompts should follow the ``fuzzy matching'' criterion because there could exist multiple output molecules. This is in contradiction with "exact matching", where the output molecules are deterministic. For example, for the functional group change, we can feed in the prompts like "change the third nitrogen in the ring to oxygen". This prompt is very explicit with an exact solution, and there exist rule-based chemistry tools in handling this problem perfectly. Thus, text-based editing cannot show its benefits in this track. Instead, text-based editing can provide more benefits in the fuzzy matching setting by wandering around the semantically meaningful directions in the latent space. This also reflects the \textit{open vocabulary} attribute of the language model that we have been focusing on.

\clearpage
\section{Downstream: Zero-shot Structure-text Retrieval} \label{sec:app:retrieval}

%%%%%%%%%%%%%%%%%%%%%%%%%%%%%%%%%%%%%%%%%%%%%%%%%%
\subsection{Dataset Construction}
\newcommand{\DrugBankData}[1]{\text{text}_{\text{DrugBank}}}
\newcommand{\PubChemCLIPData}[1]{\text{text}_{\text{\DatasetName{}}}}

The DrugBank database~\cite{wishart2018drugbank} has many fields that can be interesting to explore drug discovery tasks. Here we extract \three{} fields of each small molecule drug for the zero-shot retrieval task: the Description field, the Pharmacodynamics field, and the anatomical therapeutic chemical (ATC) field, as detailed below:
% https://docs.drugbank.com/csv/?_ga=2.99255582.1667802599.1665608295-1206864075.1648665017#drugs7
\begin{itemize}[noitemsep,topsep=0pt]
    \item \textbf{DrugBank-Description.} The Description field gives a high-level review of the drug's chemical properties, history, and regulatory status.
    \item \textbf{DrugBank-Pharmacodynamics.} This illustrates how the drug modifies or affects the organism it is being used in. This field may include effects in the body that are desired and undesired (also known as the side effects).
    \item \textbf{DrugBank-ATC.} Anatomical therapeutic chemical (ATC) is a classification system that categorizes the molecule into different groups according to the organ or system on which they act and their therapeutic, pharmacological, and chemical properties.
\end{itemize}

We list the key steps in dataset construction as follows:
\begin{enumerate}[noitemsep,topsep=0pt]
    \item We download the full DrugBank database (in XML format) and small chemical structure files (in SDF format) from the \href{https://go.drugbank.com/releases/latest}{website}.
    \item We parse the XML file, and extract the data with \three{} fields: Description, Pharmacodynamics, and ATC.
    \item We do the mapping from the extracted files to chemical structures in SDF files. For DrugBank-Description and DrugBank-Pharmacodynamics datasets, we exclude the molecules that have shown up in \DatasetName{}, filtered with the canonical SMILES. Meanwhile, for DrugBank-ATC, we exclude the molecules satisfying the following two criteria simultaneously:
        \begin{itemize}[noitemsep,topsep=0pt]
            \item \textbf{Chemical structure filtering} If the molecule with the same canonical SMILES has shown up in the \DatasetName{};
            \item \textbf{Textual data filtering} We first need to define a similarity between two textual data as in~\Cref{eq:app:text_similarity}, where $\DrugBankData{}$ and $\PubChemCLIPData{}$ are the textual data for the same molecule from DrugBank and \DatasetName{}, respectively, len() is the length of textual data, and Levenshtein() is the Levenshtein distance between two textual data.
            Thus, the second condition is: if the similarity between the DrugBank text and the \DatasetName{} text is above a certain threshold ({\eg}, 0.6). 
        \end{itemize}
    Another detail is that, for DrugBank-ATC, there exist multiple ATC fields ($\DrugBankData{}$) for each small molecule. In \DatasetName{}, there also exist multiple textual descriptions ($\PubChemCLIPData{}$) for each molecule. 
    Thus during the textual data filtering step, for each shared molecule between DrugBank and \DatasetName{}, we calculate the similarity for all the $\DrugBankData{}$-$\PubChemCLIPData{}$ pairs, and exclude the molecule if there exists one pair with similarity above the threshold 0.6.
    \item Some basic dataset statistics can be found in~\Cref{tab:app:drugbank_dataset_statistics}. Notice that ATC has many levels, and we are using level 5 for retrieval in this work.
\end{enumerate}

\begin{equation}
\label{eq:app:text_similarity}
    \small
    \text{sim} \big( \DrugBankData{}, \PubChemCLIPData{}\big) 
    = 1 -
    \frac{\text{Levenshtein} \big( \DrugBankData{}, \PubChemCLIPData{} \big)}
    {\text{len}\big(\DrugBankData{} \big)}.
\end{equation}

\begin{table}[H]
\setlength{\tabcolsep}{5pt}
\fontsize{9}{9}\selectfont
\caption{
\small
Statistics on three fields in DrugBank. The filtering steps have been illustrated above.
}
\label{tab:app:drugbank_dataset_statistics}
\centering
\vspace{-2ex}
\begin{adjustbox}{max width=\textwidth}
\begin{tabular}{l r r r}
    \toprule
    Field &
    \makecell[r]{\# structure-text pairs\\molecule not in \DatasetName{}} &
    \makecell[r]{\# structure-text pairs\\molecule shared in \DatasetName{}\\but text similarity below 0.6} &
    total \\
    \midrule
    DrugBank-Description & 1,154 & -- & 1,154\\
    DrugBank-Pharmacodynamics & 1,005 & -- & 1,005\\
    DrugBank-ATC & 1,507 & 1,500 & 3,007\\
    \bottomrule
\end{tabular}
\end{adjustbox}
\end{table}

%%%%%%%%%%
\clearpage
\subsection{Experiments}
For experiments, we introduce \three{} baselines in the main body. As a proof-of-concept, we carry out another baseline called Random. For Random, both encoders ($f_c$ and $f_t$) are randomly initialized. The zero-shot retrieval results on \three{} datasets are shown in~\Cref{tab:main_molecule_description,tab:main_molecule_pharmacodynamics,tab:main_molecule_ATC}.

\begin{table}[htb]
\setlength{\tabcolsep}{10pt}
\fontsize{9}{9}\selectfont
\caption{
\small
Accuracy (\%) of DrugBank-Description $T$-choose-one retrieval.
}
\label{tab:main_molecule_description}
\centering
\vspace{-2ex}
\begin{adjustbox}{max width=\textwidth}
\begin{tabular}{ll rrrrrr}
    \toprule
    & & \multicolumn{3}{c}{Given Chemical Structure} & \multicolumn{3}{c}{Given Text}\\
    \cmidrule(lr){3-5} \cmidrule(lr){6-8}
    & T & 4 & 10 & 20 & 4 & 10 & 20 \\
    \midrule
    \multirow{4}{*}{SMILES}
    % mode: pretrain_baselines/load_mode_0/SciBERT_MegaMolBART/downstream_retrieval_DrugBank_zero_shot
    & Random & 24.59 $\pm$ 1.14 & 10.12 $\pm$ 1.38 & 4.97 $\pm$ 0.42 & 24.54 $\pm$ 0.97 & 9.97 $\pm$ 0.81 & 5.09 $\pm$ 0.37\\
    % mode: pretrain_baselines/load_mode_1/SciBERT_MegaMolBART/downstream_retrieval_DrugBank_zero_shot
    & Frozen & 25.07 $\pm$ 1.24 & 10.22 $\pm$ 1.19 & 5.12 $\pm$ 0.65 & 24.69 $\pm$ 1.87 & 10.20 $\pm$ 1.38 & 5.37 $\pm$ 1.15\\
    % mode: pretrain_baselines/Retrieval/MegaMolBART_load_mode_1/downstream_retrieval_DrugBank_zero_shot
    & Similarity & 36.35 $\pm$ 0.59 & 23.22 $\pm$ 0.58 & 16.40 $\pm$ 0.59 & 22.74 $\pm$ 0.24 & 10.31 $\pm$ 0.24 & 5.34 $\pm$ 0.24\\
    & KV-PLM & 73.80 $\pm$ 0.00 & 53.96 $\pm$ 0.29 & 40.07 $\pm$ 0.38 & 72.86 $\pm$ 0.00 & 52.55 $\pm$ 0.29 & 40.33 $\pm$ 0.00\\
    % mode: pretrain_PubChem/SciBERT-MegaMolBART-3e-5-1-1e-4-1-InfoNCE-0.1-32-32/downstream_retrieval_DrugBank_zero_shot
    & \model\, & 97.50 $\pm$ 0.46 & 94.18 $\pm$ 0.46 & 91.12 $\pm$ 0.46 & 98.21 $\pm$ 0.00 & 94.54 $\pm$ 0.37 & 91.97 $\pm$ 0.46\\
    \midrule
    \multirow{3}{*}{Graph}
    % mode: pretrain_baselines/load_mode_0/SciBERT_Graph/downstream_retrieval_DrugBank_zero_shot
    & Random & 25.78 $\pm$ 1.43 & 10.71 $\pm$ 0.97 & 4.83 $\pm$ 1.00 & 24.98 $\pm$ 0.32 & 10.20 $\pm$ 0.40 & 4.80 $\pm$ 0.21\\
    % mode: pretrain_baselines/load_mode_1/SciBERT_Graph/downstream_retrieval_DrugBank_zero_shot
    & Frozen & 24.01 $\pm$ 1.34 & 9.39 $\pm$ 0.92 & 4.85 $\pm$ 0.52 & 24.00 $\pm$ 1.66 & 9.91 $\pm$ 0.71 & 5.07 $\pm$ 0.75\\
    % mode: pretrain_baselines/Retrieval/Graph_load_mode_1/downstream_retrieval_DrugBank_zero_shot
    & Similarity & 30.03 $\pm$ 0.38 & 13.63 $\pm$ 0.27 & 7.07 $\pm$ 0.10 & 24.81 $\pm$ 0.27 & 10.22 $\pm$ 0.24 & 4.74 $\pm$ 0.24\\
    % mode: pretrain_PubChem/SciBERT-Graph-3e-5-1-1e-4-1-InfoNCE-0.1-32-32/downstream_retrieval_DrugBank_zero_shot
    & \model\, & 99.15 $\pm$ 0.00 & 97.19 $\pm$ 0.00 & 95.66 $\pm$ 0.00 & 99.05 $\pm$ 0.37 & 97.50 $\pm$ 0.46 & 95.71 $\pm$ 0.46\\
    \bottomrule
\end{tabular}
\end{adjustbox}
\end{table}

\begin{table}[htb]
\setlength{\tabcolsep}{10pt}
\fontsize{9}{9}\selectfont
\caption{
\small
Accuracy (\%) of DrugBank-Pharmacodynamics $T$-choose-one retrieval.
}
\label{tab:main_molecule_pharmacodynamics}
\centering
\vspace{-2ex}
\begin{adjustbox}{max width=\textwidth}
\begin{tabular}{ll rrrrrr}
    \toprule
    & & \multicolumn{3}{c}{Given Chemical Structure} & \multicolumn{3}{c}{Given Text}\\
    \cmidrule(lr){3-5} \cmidrule(lr){6-8}
    & T & 4 & 10 & 20 & 4 & 10 & 20 \\
    \midrule
    \multirow{4}{*}{SMILES}
    % mode: pretrain_baselines/load_mode_0/SciBERT_MegaMolBART/downstream_retrieval_DrugBank_zero_shot
    & Random & 24.49 $\pm$ 0.68 & 9.73 $\pm$ 0.34 & 5.14 $\pm$ 0.57 & 25.61 $\pm$ 0.62 & 10.10 $\pm$ 0.91 & 5.07 $\pm$ 0.69\\
    % mode: pretrain_baselines/load_mode_1/SciBERT_MegaMolBART/downstream_retrieval_DrugBank_zero_shot
    & Frozen & 25.47 $\pm$ 1.12 & 10.55 $\pm$ 0.75 & 5.48 $\pm$ 0.70 & 25.34 $\pm$ 0.41 & 9.86 $\pm$ 0.44 & 4.84 $\pm$ 0.26\\
    % mode: pretrain_baselines/Retrieval/MegaMolBART_load_mode_1/downstream_retrieval_DrugBank_zero_shot
    & Similarity & 27.85 $\pm$ 0.03 & 10.75 $\pm$ 0.02 & 5.67 $\pm$ 0.01 & 24.58 $\pm$ 0.03 & 11.25 $\pm$ 0.03 & 5.29 $\pm$ 0.02\\
    % mode: pretrain_baselines/KV-PLM/downstream_retrieval_DrugBank_zero_shot
    & KV-PLM & 68.38 $\pm$ 0.03 & 47.59 $\pm$ 0.03 & 36.54 $\pm$ 0.03 & 67.68 $\pm$ 0.03 & 48.00 $\pm$ 0.02 & 34.66 $\pm$ 0.02\\
    % mode: pretrain_PubChem_Raw/SciBERT-MegaMolBART-1e-5-1-1e-4-1-InfoNCE-0.1-32-32/downstream_retrieval_DrugBank_zero_shot
    & \model\, & 88.07 $\pm$ 0.01 & 81.70 $\pm$ 0.02 & 75.94 $\pm$ 0.02 & 88.46 $\pm$ 0.01 & 81.01 $\pm$ 0.02 & 74.64 $\pm$ 0.03\\
    \midrule
    \multirow{3}{*}{Graph}
    % mode: pretrain_baselines/load_mode_0/SciBERT_Graph/downstream_retrieval_DrugBank_zero_shot
    & Random & 26.00 $\pm$ 0.37 & 9.65 $\pm$ 0.88 & 4.95 $\pm$ 0.36 & 25.11 $\pm$ 0.63 & 9.99 $\pm$ 0.62 & 4.82 $\pm$ 0.54\\
    % mode: pretrain_baselines/load_mode_1/SciBERT_Graph/downstream_retrieval_DrugBank_zero_shot
    & Frozen & 25.49 $\pm$ 1.82 & 10.19 $\pm$ 1.47 & 4.74 $\pm$ 0.56 & 25.55 $\pm$ 0.45 & 10.15 $\pm$ 0.77 & 4.88 $\pm$ 0.55\\
    % mode: pretrain_baselines/Retrieval/Graph_load_mode_1/downstream_retrieval_DrugBank_zero_shot
    & Similarity & 25.33 $\pm$ 0.27 & 9.89 $\pm$ 0.52 & 4.61 $\pm$ 0.08 & 25.28 $\pm$ 0.03 & 10.64 $\pm$ 0.02 & 5.47 $\pm$ 0.02\\
    % mode: pretrain_PubChem_Raw/SciBERT-Graph-1e-5-1-1e-4-1-InfoNCE-0.1-32-32/downstream_retrieval_DrugBank_zero_shot
    & \model\, & 92.14 $\pm$ 0.02 & 86.27 $\pm$ 0.02 & 81.08 $\pm$ 0.05 & 91.44 $\pm$ 0.02 & 86.76 $\pm$ 0.03 & 81.68 $\pm$ 0.03\\
    \bottomrule
\end{tabular}
\end{adjustbox}
\end{table}

\begin{table}[htb]
\setlength{\tabcolsep}{10pt}
\fontsize{9}{9}\selectfont
\caption{
\small
Accuracy (\%) of molecule-ATC $T$-choose-one retrieval.
}
\label{tab:main_molecule_ATC}
\centering
\vspace{-2ex}
\begin{adjustbox}{max width=\textwidth}
\begin{tabular}{l l c c c c c c}
    \toprule
    & & \multicolumn{3}{c}{Given Chemical Structure} & \multicolumn{3}{c}{Given Text}\\
    \cmidrule(lr){3-5} \cmidrule(lr){6-8}
    & T & 4 & 10 & 20 & 4 & 10 & 20 \\
    \midrule
    \multirow{4}{*}{SMILES}
    % mode: pretrain_baselines/load_mode_0/SciBERT_MegaMolBART/downstream_retrieval_DrugBank_zero_shot
    & Random & 25.03 $\pm$ 0.33 & 9.83 $\pm$ 0.19 & 4.80 $\pm$ 0.22 & 25.44 $\pm$ 1.21 & 10.03 $\pm$ 0.94 & 5.11 $\pm$ 0.79\\
    % mode: pretrain_baselines/load_mode_1/SciBERT_MegaMolBART/downstream_retrieval_DrugBank_zero_shot
    & Frozen & 25.05 $\pm$ 0.94 & 10.17 $\pm$ 0.63 & 4.99 $\pm$ 0.54 & 25.35 $\pm$ 0.78 & 10.32 $\pm$ 0.44 & 5.22 $\pm$ 0.34\\
    % mode: pretrain_baselines/Retrieval/MegaMolBART_load_mode_1/downstream_retrieval_DrugBank_zero_shot
    & Similarity & 30.03 $\pm$ 0.00 & 13.35 $\pm$ 0.02 & 7.53 $\pm$ 0.02 & 26.74 $\pm$ 0.03 & 11.01 $\pm$ 0.00 & 5.62 $\pm$ 0.00\\
    % mode: pretrain_baselines/KV-PLM/downstream_retrieval_DrugBank_zero_shot
    & KV-PLM & 60.94 $\pm$ 0.00 & 42.35 $\pm$ 0.00 & 30.32 $\pm$ 0.00 & 60.67 $\pm$ 0.00 & 40.19 $\pm$ 0.00 & 29.02 $\pm$ 0.00\\
    % mode: pretrain_PubChem/SciBERT-MegaMolBART-1e-5-1-1e-4-1-InfoNCE-0.1-32-32/downstream_retrieval_DrugBank_zero_shot
    & \model\,& 70.84 $\pm$ 0.07 & 56.75 $\pm$ 0.05 & 46.12 $\pm$ 0.07 & 73.07 $\pm$ 0.03 & 58.19 $\pm$ 0.03 & 48.97 $\pm$ 0.06\\
    \midrule
    \multirow{3}{*}{Graph}
    % mode: pretrain_baselines/load_mode_0/SciBERT_Graph/downstream_retrieval_DrugBank_zero_shot
    & Random & 24.48 $\pm$ 0.66 & 9.97 $\pm$ 0.25 & 4.81 $\pm$ 0.34 & 25.48 $\pm$ 0.59 & 10.40 $\pm$ 0.37 & 5.38 $\pm$ 0.30\\
    % mode: pretrain_baselines/load_mode_1/SciBERT_Graph/downstream_retrieval_DrugBank_zero_shot
    & Frozen & 24.19 $\pm$ 0.77 & 10.24 $\pm$ 0.71 & 4.87 $\pm$ 0.47 & 24.95 $\pm$ 1.52 & 10.07 $\pm$ 0.80 & 5.06 $\pm$ 0.36\\
    % mode: pretrain_baselines/Retrieval/Graph_load_mode_1/downstream_retrieval_DrugBank_zero_shot
    & Similarity & 29.46 $\pm$ 0.00 & 12.34 $\pm$ 0.00 & 6.52 $\pm$ 0.00 & 25.78 $\pm$ 1.53 & 10.23 $\pm$ 0.70 & 5.06 $\pm$ 0.67\\
    % mode: pretrain_PubChem/SciBERT-Graph-1e-5-1-1e-4-1-InfoNCE-0.1-32-32/downstream_retrieval_DrugBank_zero_shot
    & \model\, & 69.33 $\pm$ 0.03 & 54.83 $\pm$ 0.04 & 44.13 $\pm$ 0.05 & 71.81 $\pm$ 0.05 & 58.34 $\pm$ 0.07 & 47.58 $\pm$ 0.05\\
    \bottomrule
\end{tabular}
\end{adjustbox}
% \vspace{-5ex}
\end{table}

\clearpage
%%%%%%%%%%%%%%%%%%%%%%%%%%%%%%%%%%%%%%%%%%%%%%%%%%
\subsection{Ablation Study: Fixed Pretrained Encoders}
In the main body, we conduct pretraining by adopting pretrained single-modality checkpoints, {\ie}, the GraphMVP and MegaMolBART for $f_c$, and SciBERT for$f_t$. Then for \model{} pretraining, we use contrastive learning and update all the model parameters. Here we take an ablation study by only optimizing the projection layers to the joint space of the two branches ($p_c, p_t$) while keeping the two encoders ($f_c, f_t$) fixed. The results on the three datasets are shown in~\Cref{tab:main_molecule_description_frozen,tab:main_molecule_pharmacodynamics_frozen,tab:main_molecule_ATC_frozen}.

\begin{table}[h]
\setlength{\tabcolsep}{10pt}
\fontsize{9}{9}\selectfont
\caption{
\small
Accuracy (\%) of DrugBank-Description $T$-choose-one retrieval.
}
\label{tab:main_molecule_description_frozen}
\centering
\vspace{-2ex}
\begin{adjustbox}{max width=\textwidth}
\begin{tabular}{ll rrrrrr}
    \toprule
    & & \multicolumn{3}{c}{Given Chemical Structure} & \multicolumn{3}{c}{Given Text}\\
    \cmidrule(lr){3-5} \cmidrule(lr){6-8}
    & T & 4 & 10 & 20 & 4 & 10 & 20 \\
    \midrule
    \multirow{4}{*}{SMILES}
    % mode: pretrain_baselines/load_mode_0/SciBERT_MegaMolBART/downstream_retrieval_DrugBank_zero_shot
    & Random & 24.59 $\pm$ 1.14 & 10.12 $\pm$ 1.38 & 4.97 $\pm$ 0.42 & 24.54 $\pm$ 0.97 & 9.97 $\pm$ 0.81 & 5.09 $\pm$ 0.37\\
    % mode: pretrain_baselines/load_mode_1/SciBERT_MegaMolBART/downstream_retrieval_DrugBank_zero_shot
    & Frozen & 25.07 $\pm$ 1.24 & 10.22 $\pm$ 1.19 & 5.12 $\pm$ 0.65 & 24.69 $\pm$ 1.87 & 10.20 $\pm$ 1.38 & 5.37 $\pm$ 1.15\\
    % mode: pretrain_baselines/Retrieval/MegaMolBART_load_mode_1/downstream_retrieval_DrugBank_zero_shot
    & Similarity & 36.35 $\pm$ 0.59 & 23.22 $\pm$ 0.58 & 16.40 $\pm$ 0.59 & 22.74 $\pm$ 0.24 & 10.31 $\pm$ 0.24 & 5.34 $\pm$ 0.24\\
    % mode: pretrain_PubChem_Raw_frozen/SciBERT-MegaMolBART-3e-5-1-1e-4-1-InfoNCE-0.1-32-32/downstream_retrieval_DrugBank_zero_shot
    & \model\, & 47.64 $\pm$ 0.40 & 29.21 $\pm$ 0.47 & 19.69 $\pm$ 0.47 & 52.60 $\pm$ 0.46 & 32.24 $\pm$ 0.37 & 21.45 $\pm$ 0.37\\
    \midrule
    \multirow{3}{*}{Graph}
    % mode: pretrain_baselines/load_mode_0/SciBERT_Graph/downstream_retrieval_DrugBank_zero_shot
    & Random & 25.78 $\pm$ 1.43 & 10.71 $\pm$ 0.97 & 4.83 $\pm$ 1.00 & 24.98 $\pm$ 0.32 & 10.20 $\pm$ 0.40 & 4.80 $\pm$ 0.21\\
    % mode: pretrain_baselines/load_mode_1/SciBERT_Graph/downstream_retrieval_DrugBank_zero_shot
    & Frozen & 24.01 $\pm$ 1.34 & 9.39 $\pm$ 0.92 & 4.85 $\pm$ 0.52 & 24.00 $\pm$ 1.66 & 9.91 $\pm$ 0.71 & 5.07 $\pm$ 0.75\\
    % mode: pretrain_baselines/Retrieval/Graph_load_mode_1/downstream_retrieval_DrugBank_zero_shot
    & Similarity & 30.03 $\pm$ 0.38 & 13.63 $\pm$ 0.27 & 7.07 $\pm$ 0.10 & 24.81 $\pm$ 0.27 & 10.22 $\pm$ 0.24 & 4.74 $\pm$ 0.24\\
    % mode: pretrain_PubChem_Raw_frozen/SciBERT-Graph-3e-5-1-1e-4-1-InfoNCE-0.1-32-32/downstream_retrieval_DrugBank_zero_shot
    & \model\, & 51.28 $\pm$ 0.00 & 31.99 $\pm$ 0.41 & 20.71 $\pm$ 0.47 & 55.27 $\pm$ 0.00 & 33.08 $\pm$ 0.00 & 21.77 $\pm$ 0.00\\
    \bottomrule
\end{tabular}
\end{adjustbox}
\end{table}

\begin{table}[h]
\setlength{\tabcolsep}{10pt}
\fontsize{9}{9}\selectfont
\caption{
\small
Accuracy (\%) of DrugBank-Pharmacodynamics $T$-choose-one retrieval.
}
\label{tab:main_molecule_pharmacodynamics_frozen}
\centering
\vspace{-2ex}
\begin{adjustbox}{max width=\textwidth}
\begin{tabular}{ll rrrrrr}
    \toprule
    & & \multicolumn{3}{c}{Given Chemical Structure} & \multicolumn{3}{c}{Given Text}\\
    \cmidrule(lr){3-5} \cmidrule(lr){6-8}
    & T & 4 & 10 & 20 & 4 & 10 & 20 \\
    \midrule
    \multirow{4}{*}{SMILES}
    % mode: pretrain_baselines/load_mode_0/SciBERT_MegaMolBART/downstream_retrieval_DrugBank_zero_shot
    & Random & 24.49 $\pm$ 0.68 & 9.73 $\pm$ 0.34 & 5.14 $\pm$ 0.57 & 25.61 $\pm$ 0.62 & 10.10 $\pm$ 0.91 & 5.07 $\pm$ 0.69\\
    % mode: pretrain_baselines/load_mode_1/SciBERT_MegaMolBART/downstream_retrieval_DrugBank_zero_shot
    & Frozen & 25.47 $\pm$ 1.12 & 10.55 $\pm$ 0.75 & 5.48 $\pm$ 0.70 & 25.34 $\pm$ 0.41 & 9.86 $\pm$ 0.44 & 4.84 $\pm$ 0.26\\
    % mode: pretrain_baselines/Retrieval/MegaMolBART_load_mode_1/downstream_retrieval_DrugBank_zero_shot
    & Similarity & 27.85 $\pm$ 0.03 & 10.75 $\pm$ 0.02 & 5.67 $\pm$ 0.01 & 24.58 $\pm$ 0.03 & 11.25 $\pm$ 0.03 & 5.29 $\pm$ 0.02\\
    % mode: pretrain_PubChem_Raw_frozen/SciBERT-MegaMolBART-1e-5-1-1e-4-1-InfoNCE-0.1-32-32/downstream_retrieval_DrugBank_zero_shot
    & \model\, & 46.43 $\pm$ 0.00 & 27.42 $\pm$ 0.47 & 18.24 $\pm$ 0.47 & 52.53 $\pm$ 0.41 & 30.53 $\pm$ 0.00 & 19.98 $\pm$ 0.00\\
    \midrule
    \multirow{3}{*}{Graph}
    % mode: pretrain_baselines/load_mode_0/SciBERT_Graph/downstream_retrieval_DrugBank_zero_shot
    & Random & 26.00 $\pm$ 0.37 & 9.65 $\pm$ 0.88 & 4.95 $\pm$ 0.36 & 25.11 $\pm$ 0.63 & 9.99 $\pm$ 0.62 & 4.82 $\pm$ 0.54\\
    % mode: pretrain_baselines/load_mode_1/SciBERT_Graph/downstream_retrieval_DrugBank_zero_shot
    & Frozen & 25.49 $\pm$ 1.82 & 10.19 $\pm$ 1.47 & 4.74 $\pm$ 0.56 & 25.55 $\pm$ 0.45 & 10.15 $\pm$ 0.77 & 4.88 $\pm$ 0.55\\
    % mode: pretrain_baselines/Retrieval/Graph_load_mode_1/downstream_retrieval_DrugBank_zero_shot
    & Similarity & 25.33 $\pm$ 0.27 & 9.89 $\pm$ 0.52 & 4.61 $\pm$ 0.08 & 25.28 $\pm$ 0.03 & 10.64 $\pm$ 0.02 & 5.47 $\pm$ 0.02\\
    % mode: pretrain_PubChem_Raw_frozen/SciBERT-Graph-1e-5-1-1e-4-1-InfoNCE-0.1-32-32/downstream_retrieval_DrugBank_zero_shot
    & \model\, & 46.29 $\pm$ 0.03 & 27.18 $\pm$ 0.02 & 17.73 $\pm$ 0.02 & 50.95 $\pm$ 0.04 & 31.65 $\pm$ 0.03 & 23.00 $\pm$ 0.03\\
    \bottomrule
\end{tabular}
\end{adjustbox}
\end{table}

\begin{table}[h]
\setlength{\tabcolsep}{10pt}
\fontsize{9}{9}\selectfont
\caption{
\small
Accuracy (\%) of DrugBank-ATC $T$-choose-one retrieval.
}
\label{tab:main_molecule_ATC_frozen}
\centering
\vspace{-2ex}
\begin{adjustbox}{max width=\textwidth}
\begin{tabular}{l l c c c c c c}
    \toprule
    & & \multicolumn{3}{c}{Given Chemical Structure} & \multicolumn{3}{c}{Given Text}\\
    \cmidrule(lr){3-5} \cmidrule(lr){6-8}
    & T & 4 & 10 & 20 & 4 & 10 & 20 \\
    \midrule
    \multirow{4}{*}{SMILES}
    % mode: pretrain_baselines/load_mode_0/SciBERT_MegaMolBART/downstream_retrieval_DrugBank_zero_shot
    & Random & 25.03 $\pm$ 0.33 & 9.83 $\pm$ 0.19 & 4.80 $\pm$ 0.22 & 25.44 $\pm$ 1.21 & 10.03 $\pm$ 0.94 & 5.11 $\pm$ 0.79\\
    % mode: pretrain_baselines/load_mode_1/SciBERT_MegaMolBART/downstream_retrieval_DrugBank_zero_shot
    & Frozen & 25.05 $\pm$ 0.94 & 10.17 $\pm$ 0.63 & 4.99 $\pm$ 0.54 & 25.35 $\pm$ 0.78 & 10.32 $\pm$ 0.44 & 5.22 $\pm$ 0.34\\
    % mode: pretrain_baselines/Retrieval/MegaMolBART_load_mode_1/downstream_retrieval_DrugBank_zero_shot
    & Similarity & 30.03 $\pm$ 0.00 & 13.35 $\pm$ 0.02 & 7.53 $\pm$ 0.02 & 26.74 $\pm$ 0.03 & 11.01 $\pm$ 0.00 & 5.62 $\pm$ 0.00\\
    % mode: pretrain_PubChem_Raw_frozen/SciBERT-MegaMolBART-1e-5-1-1e-4-1-InfoNCE-0.1-32-32/downstream_retrieval_DrugBank_zero_shot
    & \model\, & 43.41 $\pm$ 0.12 & 25.66 $\pm$ 0.06 & 15.69 $\pm$ 0.06 & 48.75 $\pm$ 0.11 & 29.44 $\pm$ 0.06 & 19.75 $\pm$ 0.03\\
    \midrule
    \multirow{3}{*}{Graph}
    % mode: pretrain_baselines/load_mode_0/SciBERT_Graph/downstream_retrieval_DrugBank_zero_shot
    & Random & 24.48 $\pm$ 0.66 & 9.97 $\pm$ 0.25 & 4.81 $\pm$ 0.34 & 25.48 $\pm$ 0.59 & 10.40 $\pm$ 0.37 & 5.38 $\pm$ 0.30\\
    % mode: pretrain_baselines/load_mode_1/SciBERT_Graph/downstream_retrieval_DrugBank_zero_shot
    & Frozen & 24.19 $\pm$ 0.77 & 10.24 $\pm$ 0.71 & 4.87 $\pm$ 0.47 & 24.95 $\pm$ 1.52 & 10.07 $\pm$ 0.80 & 5.06 $\pm$ 0.36\\
    % mode: pretrain_baselines/Retrieval/Graph_load_mode_1/downstream_retrieval_DrugBank_zero_shot
    & Similarity & 29.46 $\pm$ 0.00 & 12.34 $\pm$ 0.00 & 6.52 $\pm$ 0.00 & 25.78 $\pm$ 1.53 & 10.23 $\pm$ 0.70 & 5.06 $\pm$ 0.67\\
    % mode: pretrain_PubChem_Raw_frozen/SciBERT-Graph-1e-5-1-1e-4-1-InfoNCE-0.1-32-32/downstream_retrieval_DrugBank_zero_shot
    & \model\, & 42.53 $\pm$ 0.07 & 24.34 $\pm$ 0.00 & 14.78 $\pm$ 0.03 & 48.91 $\pm$ 0.03 & 28.77 $\pm$ 0.07 & 19.28 $\pm$ 0.07\\
    \bottomrule
\end{tabular}
\end{adjustbox}
\end{table}

\clearpage
\section{Downstream: Zero-shot Text-based Molecule Editing} \label{app:sec:text_based_editing}
Molecule editing or controllable molecule generation refers to changing the structures of the molecules based on a given and pretrained molecule generative model. In this work, with the help of a large language model in \model{}, we are able to do the zero-shot text-based molecule editing. First, we would like to list \two{} key challenges, comparing the editing task between the vision domain and molecule domain, as follows:
\begin{itemize}[noitemsep,topsep=0pt]
    \item \textbf{Backbone generative model.}
    For domains in vision, the image controllable generation can be quite feasible based on StyleGAN~\cite{karras2019style}, a well-disentangled backbone model. However, it is nontrivial for deep molecule generative models. A recent work GraphCG~\cite{liu2022graphcg} has explored the disentanglement property of the graph-based controllable molecule generation methods, and the conclusion is that, even though the backbone generative models are not perfectly disentangled, there still exist methods for controllable generation on highly structured data like molecular graphs or point clouds. Meanwhile, developing a novel disentangled molecule generative model is out of the scope of this work, since the editing solution by \model{} is model-agnostic, and can be easily generalized to future models.
    \item \textbf{Evaluation.} Image controllable generation is an art problem, {\ie}, it is subjective and can have multiple (or even infinitely many) answers. On the contrary, controllable molecule generation is a science problem, {\ie}, it is objective and has only a few answers. This has been discussed in~\Cref{sec:app:downstream}.
\end{itemize}

%%%%%%%%%%%%%%%%%%%%%%%%%%%%%%%%%%%%%%%%%%%%%%%%%%
\subsection{Experiment Set-up}
\paragraph{Implementation Details}
Because most of the modules are fixed, we only need to learn the adaptor module and the optimized latent code $w$. The two key hyperparameters are the learning rate \{1e-2, 1e-3\} and $\lambda \in \{1e1, 1e0, 1e-1, 1e-2, 1e-3\}$. As a fair comparison, for baselines, we take the form of $w = w_{\text{in}} + \alpha \cdot D$, where $D$ is obtained using random, PCA and variance and $\lambda \in \{ 1.0, 1.5, 2.0, 2.5, 3.0 \}$. For GS, we repeat the random sampling five times of each input molecule.

Next, we will conduct the zero-shot text-based molecule editing on \four{} types of editing tasks, as well as \three{} case study, as discussed below:
\begin{itemize}[noitemsep,topsep=0pt]
    \item Single-objective molecule editing in~\Cref{sec:app:single_objective_property_editing} (\eight{} tasks).
    \item Multi-objective molecule editing in~\Cref{sec:app:multi_objective_property_editing} (\six{} tasks).
    \item Binding-affinity-based molecule editing in~\Cref{sec:app:ChEMBL_assay_editing} (\six{} tasks).
    \item Drug relevance editing in~\Cref{sec:app:common_drug_editing} (\four{} tasks).
    \item Neighborhood searching for patent drug molecules in~\Cref{sec:app:patent_drug_searching} (\three{} case studies).
\end{itemize}
Due to the page limit, we only show \four{} multi-objective and \four{} binding-affinity-based editing tasks in the main body. Here we show more comprehensive results.

\begin{figure}[h]
\centering
\begin{subfigure}[h]{0.24\textwidth}
    \centering
    \includegraphics[width=\textwidth]{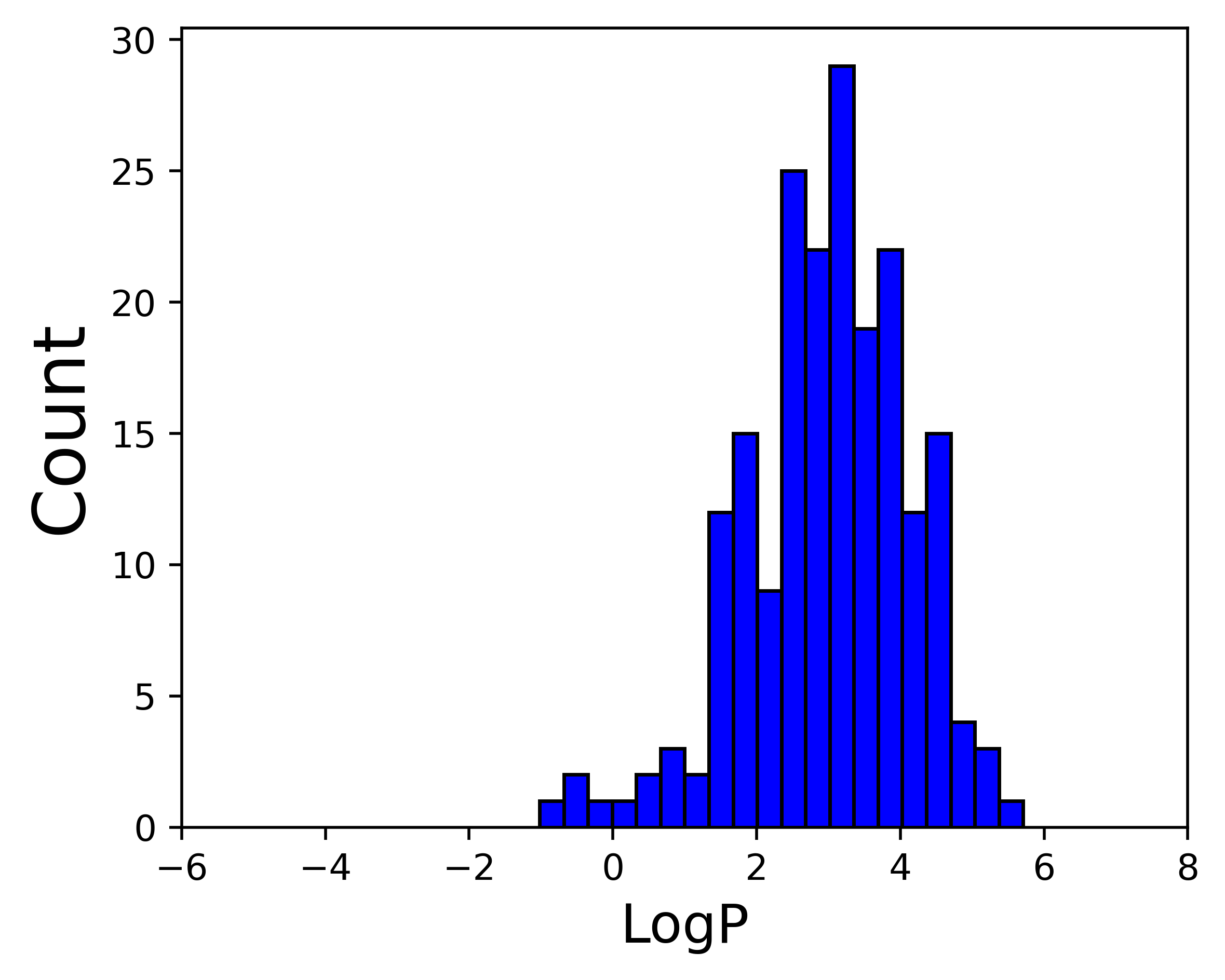}
    \vspace{-3ex}
\end{subfigure}
\begin{subfigure}[h]{0.24\textwidth}
    \centering
    \includegraphics[width=\textwidth]{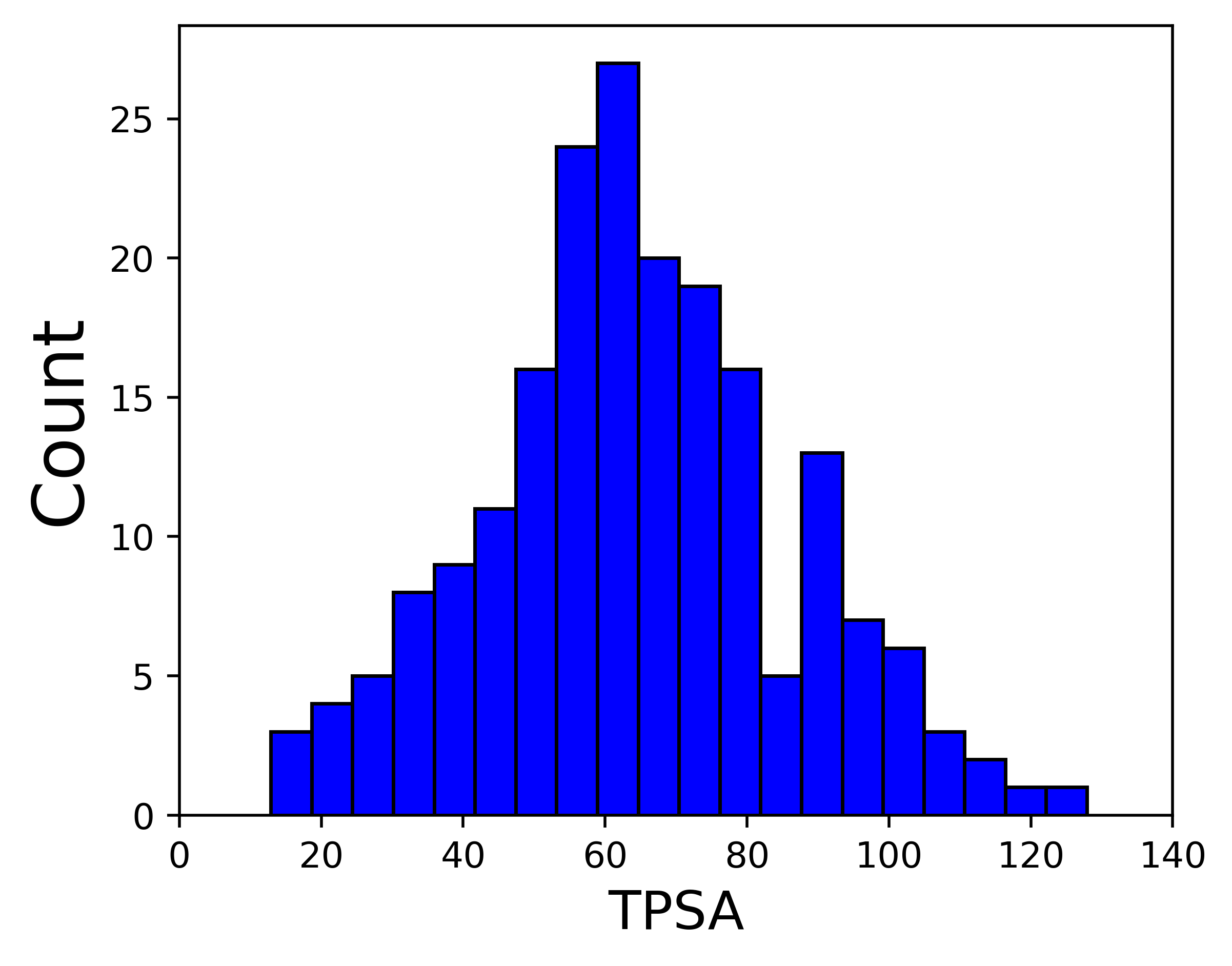}
    \vspace{-3ex}
\end{subfigure}
\begin{subfigure}[h]{0.24\textwidth}
    \centering
    \includegraphics[width=\textwidth]{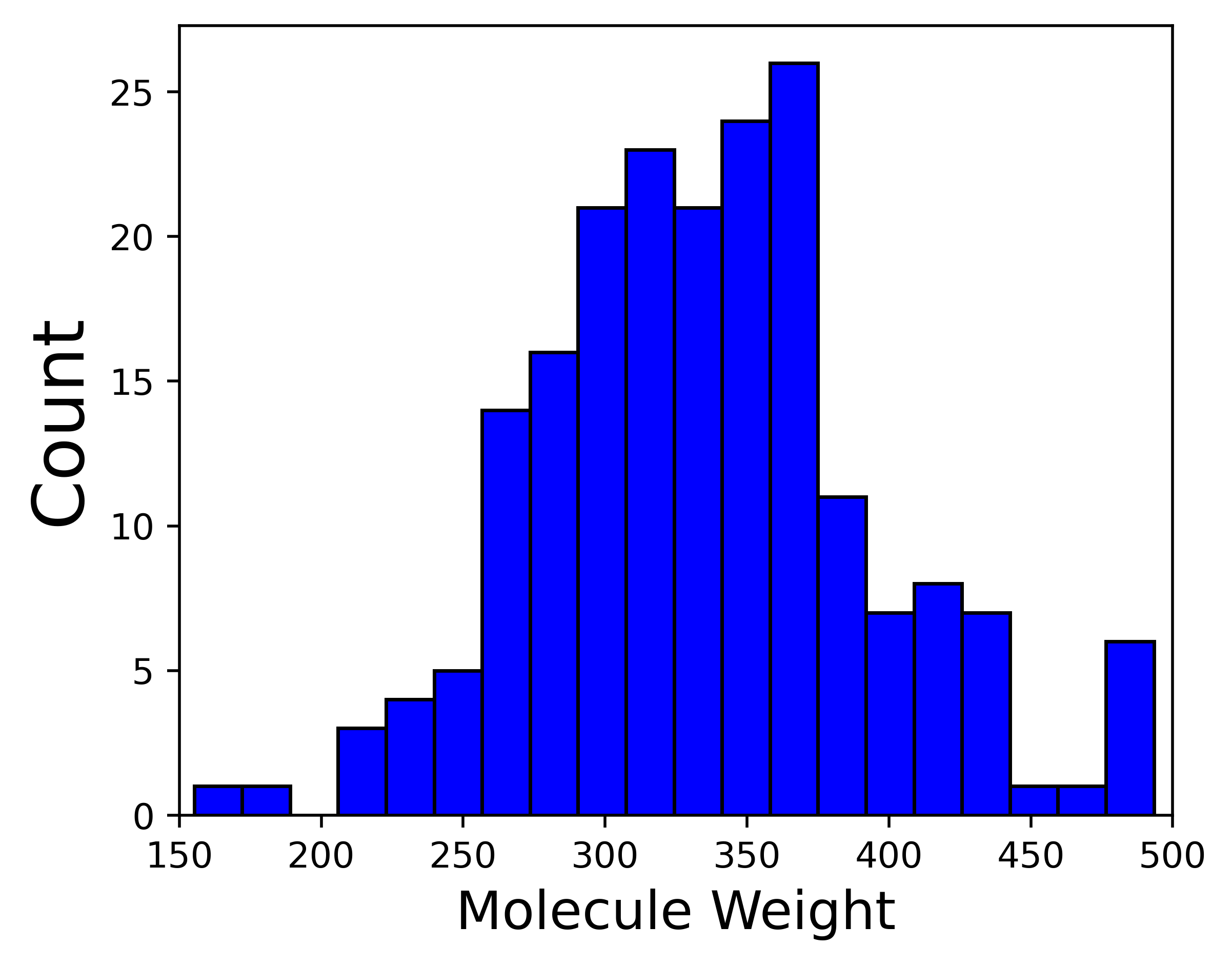}
    \vspace{-3ex}
\end{subfigure}
\vspace{-2ex}
\caption{\small Three property distributions on 200 randomly sampled molecules for editing.}
\label{fig:editing_molecule_distribution}
\vspace{+2ex}
\begin{subfigure}[h]{0.24\textwidth}
    \centering
    \includegraphics[width=\textwidth]{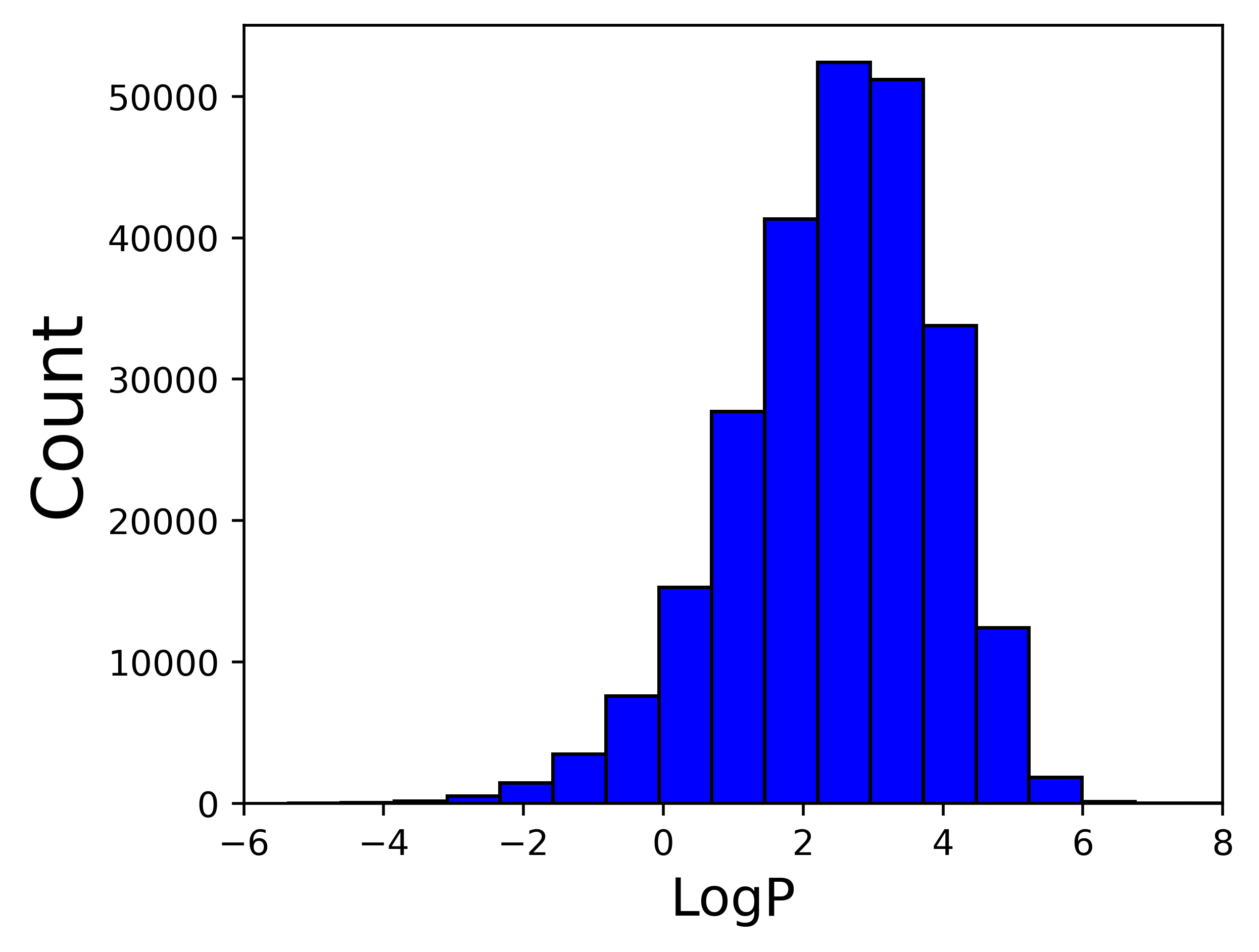}
    \vspace{-3ex}
\end{subfigure}
\begin{subfigure}[h]{0.24\textwidth}
    \centering
    \includegraphics[width=\textwidth]{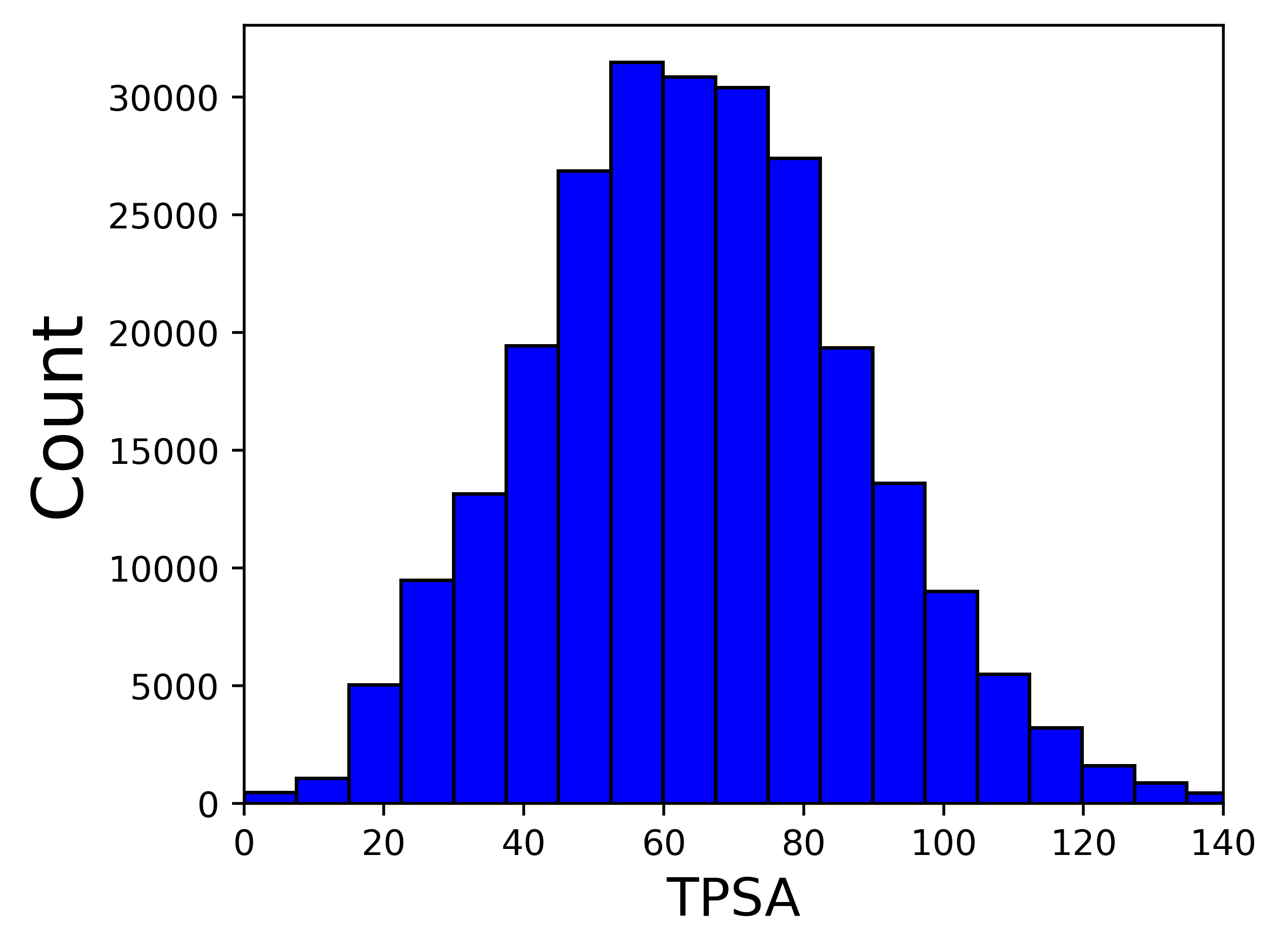}
    \vspace{-3ex}
\end{subfigure}
\begin{subfigure}[h]{0.24\textwidth}
    \centering
    \includegraphics[width=\textwidth]{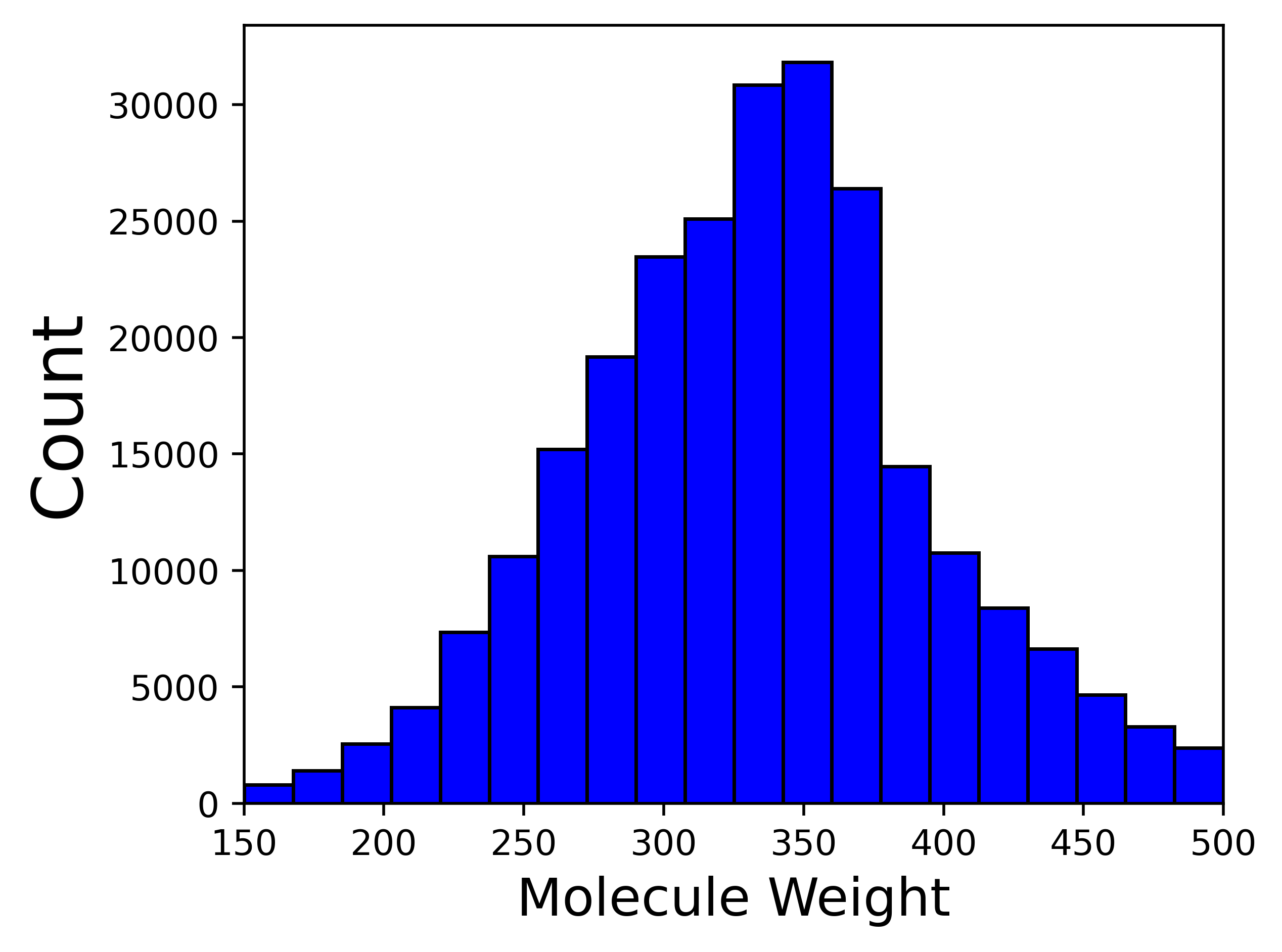}
    \vspace{-3ex}
\end{subfigure}
\vspace{-2ex}
\caption{\small Three property distributions on 250K molecules from ZINC250K.}
\label{fig:ZINC250K_molecule_distribution}
\end{figure}

We want to mention that for single- and multi-objective editing, we randomly select 200 molecules from ZINC as the input molecules. None of these 200 input molecules appears in PubChemSTM. Furthermore, the random selection process ensures that the property distributions of these 200 molecules remain consistent with the entire dataset. Illustrated below (\Cref{fig:editing_molecule_distribution,fig:ZINC250K_molecule_distribution}) are three examples of molecular properties: LogP (measuring water solubility), tPSA (measuring permeability) and molecular weight.

\newpage
%%%%%%%%%%%%%%%%%%%%%%%%%%%%%%%%%%%%%%%%%%%%%%%%%%
\subsection{Single-objective Molecule Editing} \label{sec:app:single_objective_property_editing}
We first consider \eight{} single-objective properties for molecule editing. As shown in the Methods section, the definitions of the satisfaction function and threshold $\Delta$ are based on each task specifically, as:
\begin{itemize}[noitemsep,topsep=0pt]
    \item We use LogP to evaluate the solubility and insolubility. We take 0 and 0.5 as the different thresholds.
    \item We use QED to evaluate the drug-likeness. We take 0 and 0.1 as the different thresholds.
    \item We use tPSA to evaluate the high and low permeability. We take 0 and 10 as the different thresholds.
    \item For the hydrogen bond acceptor (HBA) and hydrogen bond donor (HBD), we can directly count them in the molecules, and we use 0 and 1 as the different thresholds.
\end{itemize}
For $\Delta$, it is the threshold that only difference above it can be viewed as a hit. So the larger $\Delta$ means a stricter editing criterion. Below we show both the quantitative and qualitative results on \eight{} single-objective property molecule editing results.

\begin{table}[htb!]
\caption{
\small
Results on \eight{} single-objective molecule editing. The inputs are 200 molecules randomly sampled from ZINC, and the evaluation is the hit ratio of the property change. The latent optimization is text-based molecule editing with \model{}, with the SMILES string and the molecular graph, respectively.
}
\centering
\vspace{-2ex}
\begin{adjustbox}{max width=\textwidth}
\begin{tabular}{l l c c c c c c c}
\toprule
 & & \multicolumn{4}{c}{baseline} & \multicolumn{2}{c}{latent optimization}\\
\cmidrule(lr){3-6} \cmidrule(lr){7-8}
 & $\Delta$ & Random & PCA & High Variance & GS-Mutate & SMILES & Graph\\
\midrule 
\multirow{2}{*}{This molecule is \textit{soluble in water}.} & 0 & 35.33 $\pm$ 1.31 & 33.80 $\pm$ 3.63 & 33.52 $\pm$ 3.75 & 52.00 $\pm$ 0.41 & 61.87 $\pm$ 2.67 & 67.86 $\pm$ 3.46\\
 & 0.5 & 11.04 $\pm$ 2.40 & 10.66 $\pm$ 3.24 & 10.86 $\pm$ 2.56 & 14.67 $\pm$ 0.62 & 49.02 $\pm$ 1.84 & 54.44 $\pm$ 3.99\\
\midrule
\multirow{2}{*}{This molecule is \textit{insoluble in water}.} & 0 & 43.36 $\pm$ 3.06 & 39.36 $\pm$ 2.55 & 42.89 $\pm$ 2.36 & 47.50 $\pm$ 0.41 & 52.71 $\pm$ 1.67 & 64.79 $\pm$ 2.76\\
 & 0.5 & 19.75 $\pm$ 1.56 & 15.12 $\pm$ 2.93 & 18.22 $\pm$ 0.33 & 12.50 $\pm$ 0.82 & 30.47 $\pm$ 3.26 & 47.09 $\pm$ 3.42\\
\midrule
\multirow{2}{*}{This molecule is \textit{like a drug}.} & 0 & 38.06 $\pm$ 2.57 & 33.99 $\pm$ 3.72 & 36.20 $\pm$ 4.34 & 28.00 $\pm$ 0.71 & 36.52 $\pm$ 2.46 & 39.97 $\pm$ 4.32\\
 & 0.1 & 5.27 $\pm$ 0.24 & 3.97 $\pm$ 0.10 & 4.44 $\pm$ 0.58 & 6.33 $\pm$ 2.09 & 8.81 $\pm$ 0.82 & 14.06 $\pm$ 3.18\\
\midrule
\multirow{2}{*}{This molecule is \textit{not like a drug}.} & 0 & 36.96 $\pm$ 2.25 & 35.17 $\pm$ 2.61 & 39.99 $\pm$ 0.57 & 71.33 $\pm$ 0.85 & 58.59 $\pm$ 1.01 & 77.62 $\pm$ 2.80\\
 & 0.1 & 6.16 $\pm$ 1.87 & 5.26 $\pm$ 0.95 & 7.56 $\pm$ 0.29 & 27.67 $\pm$ 3.79 & 37.56 $\pm$ 1.76 & 54.22 $\pm$ 3.12\\
\midrule
\multirow{2}{*}{This molecule has \textit{high permeability}.} & 0 & 25.23 $\pm$ 2.13 & 21.36 $\pm$ 0.79 & 21.98 $\pm$ 3.77 & 22.00 $\pm$ 0.82 & 57.74 $\pm$ 0.60 & 59.84 $\pm$ 0.78\\
 & 10 & 17.41 $\pm$ 1.43 & 14.52 $\pm$ 0.80 & 14.66 $\pm$ 2.13 & 6.17 $\pm$ 0.62 & 47.51 $\pm$ 1.88 & 50.42 $\pm$ 2.73\\
\midrule
\multirow{2}{*}{This molecule has \textit{low permeability}.} & 0 & 16.79 $\pm$ 2.54 & 15.48 $\pm$ 2.40 & 17.10 $\pm$ 1.14 & 28.83 $\pm$ 1.25 & 34.13 $\pm$ 0.59 & 31.76 $\pm$ 0.97\\
 & 10 & 11.02 $\pm$ 0.71 & 10.62 $\pm$ 1.86 & 12.01 $\pm$ 1.01 & 15.17 $\pm$ 1.03 & 26.48 $\pm$ 0.97 & 19.76 $\pm$ 1.31\\
\midrule
\multirow{2}{*}{This molecule has \textit{more hydrogen bond acceptors}.} & 0 & 12.64 $\pm$ 1.64 & 10.85 $\pm$ 2.29 & 11.78 $\pm$ 0.15 & 21.17 $\pm$ 3.09 & 54.01 $\pm$ 5.26 & 37.35 $\pm$ 0.79\\
 & 1 & 0.69 $\pm$ 0.01 & 0.90 $\pm$ 0.84 & 0.67 $\pm$ 0.01 & 1.83 $\pm$ 0.47 & 27.33 $\pm$ 2.62 & 16.13 $\pm$ 2.87\\
\midrule
\multirow{2}{*}{This molecule has \textit{more hydrogen bond donors}.} & 0 & 2.97 $\pm$ 0.61 & 3.97 $\pm$ 0.55 & 6.23 $\pm$ 0.66 & 19.50 $\pm$ 2.86 & 28.55 $\pm$ 0.76 & 60.97 $\pm$ 5.09\\
 & 1 & 0.00 $\pm$ 0.00 & 0.00 $\pm$ 0.00 & 0.00 $\pm$ 0.00 & 1.33 $\pm$ 0.24 & 7.69 $\pm$ 0.56 & 32.35 $\pm$ 2.57\\
\bottomrule
\end{tabular}
\end{adjustbox}
\end{table}

\begin{table}[htb]
\setlength{\tabcolsep}{10pt}
\fontsize{9}{9}\selectfont
\centering
\caption{\small
Visualization of text-based editing on solubility, measured by the logarithm of the octanol-water partition coefficient (LogP) of the molecules. Generally, molecules with smaller LogP are more soluble in water. For generating molecules soluble in water, we can add polar components ({\eg}, oxygens and nitrogens), remove hydrophobic moieties ({\eg}, benzene and cyclohexane), or replace hydrophobic groups with polar functionalities in the input molecule. For generating molecules insoluble in water, we can make opposite modifications to the input molecule. The pink and blue regions highlight the modified structure in the input and output molecules, respectively.
}
\vspace{-2ex}
    \begin{adjustbox}{max width=\textwidth}
    \small
    \begin{tabular}{cccccc}
    \toprule
    \multicolumn{6}{c}{Text Prompt: This molecule is \textit{soluble in water}.}\\
    \cmidrule(lr){1-6}
    {Input Molecule} & {Output Molecule} & {Input Molecule} & {Output Molecule} & {Input Molecule} & {Output Molecule}\\
    \cmidrule(lr){1-1}\cmidrule(lr){2-2}\cmidrule(lr){3-3}\cmidrule(lr){4-4}\cmidrule(lr){5-5}\cmidrule(lr){6-6}
    \adjustbox{valign=c}{\includegraphics[width=0.16\linewidth]{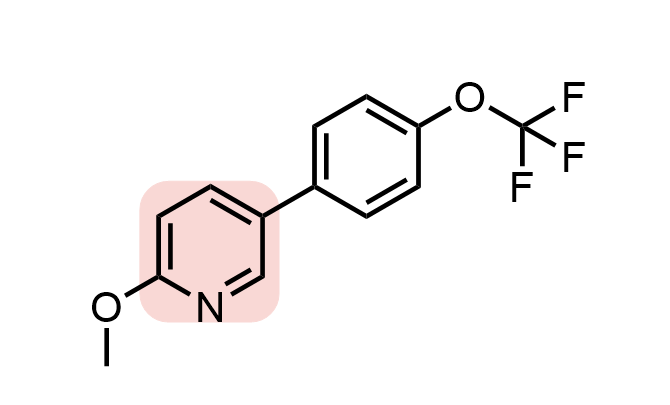}} &
    \adjustbox{valign=c}{\includegraphics[width=0.16\linewidth]{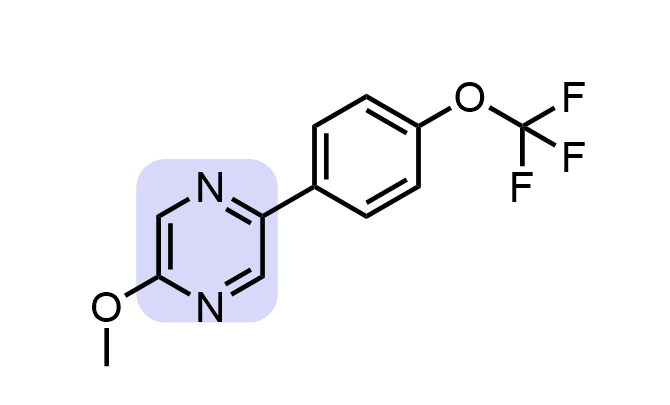}} &
    \adjustbox{valign=c}{\includegraphics[width=0.16\linewidth]{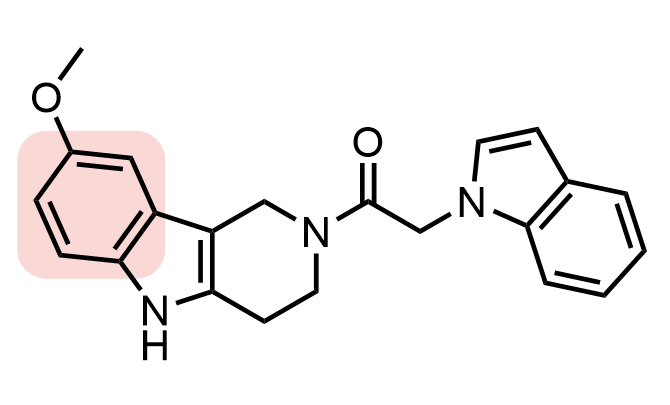}} &
    \adjustbox{valign=c}{\includegraphics[width=0.16\linewidth]{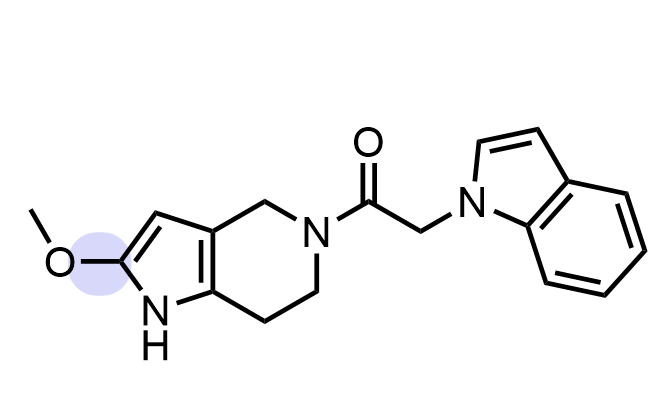}} &
    \adjustbox{valign=c}{\includegraphics[width=0.16\linewidth]{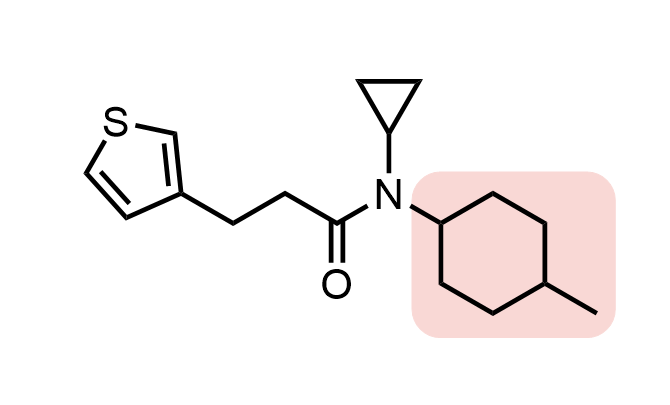}} &
    \adjustbox{valign=c}{\includegraphics[width=0.16\linewidth]{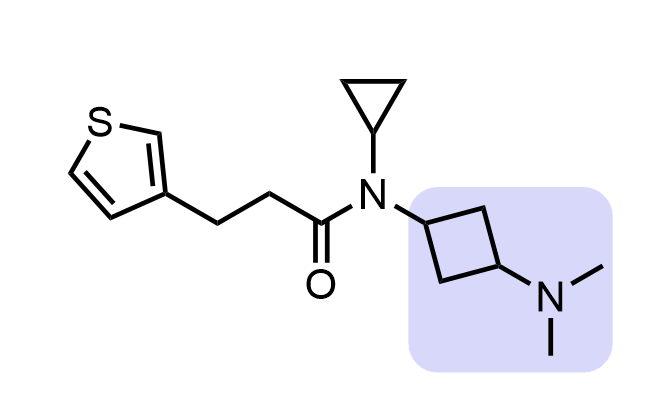}}
    \\
    {\small LogP: 3.66} & {\small LogP: 3.05} & {\small LogP: 3.72} & {\small LogP: 2.56} & {\small LogP: 4.25} & {\small LogP: 2.76}\\[5pt]
    \toprule
    \multicolumn{6}{c}{Text Prompt: This molecule is \textit{insoluble in water}.}\\
    \cmidrule(lr){1-6}
    {Input Molecule} & {Output Molecule} & {Input Molecule} & {Output Molecule} & {Input Molecule} & {Output Molecule}\\
    \cmidrule(lr){1-1}\cmidrule(lr){2-2}\cmidrule(lr){3-3}\cmidrule(lr){4-4}\cmidrule(lr){5-5}\cmidrule(lr){6-6}
    \adjustbox{valign=c}{\includegraphics[width=0.16\linewidth]{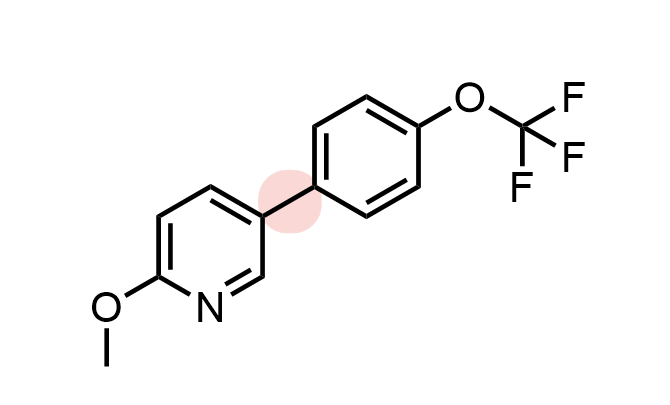}} &
    \adjustbox{valign=c}{\includegraphics[width=0.16\linewidth]{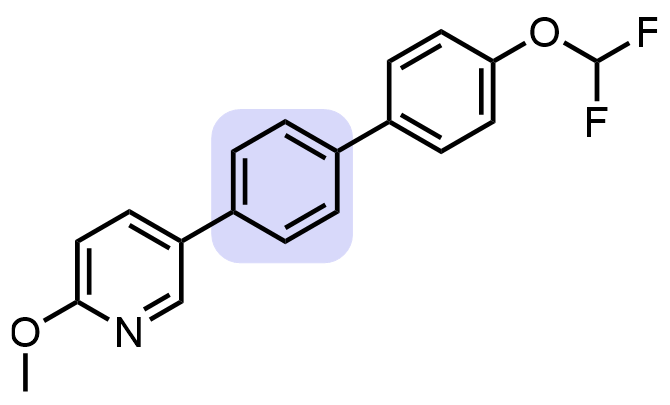}} &
    \adjustbox{valign=c}{\includegraphics[width=0.16\linewidth]{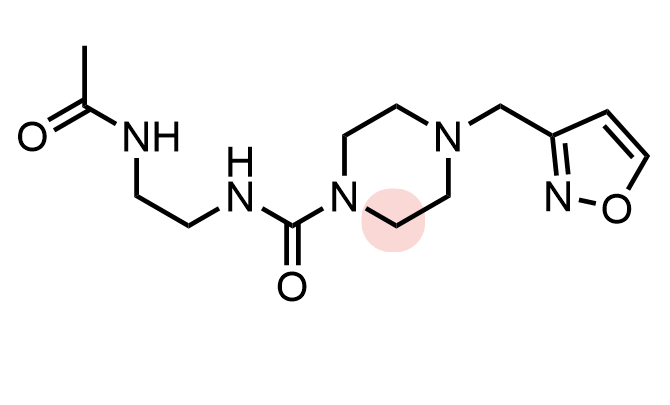}} &
    \adjustbox{valign=c}{\includegraphics[width=0.16\linewidth]{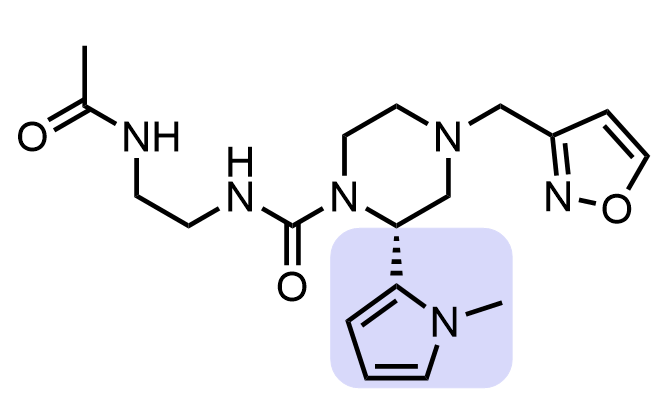}} &
    \adjustbox{valign=c}{\includegraphics[width=0.16\linewidth]{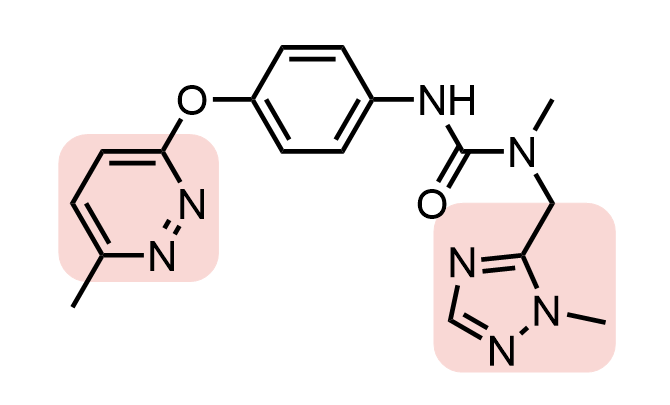}} &
    \adjustbox{valign=c}{\includegraphics[width=0.16\linewidth]{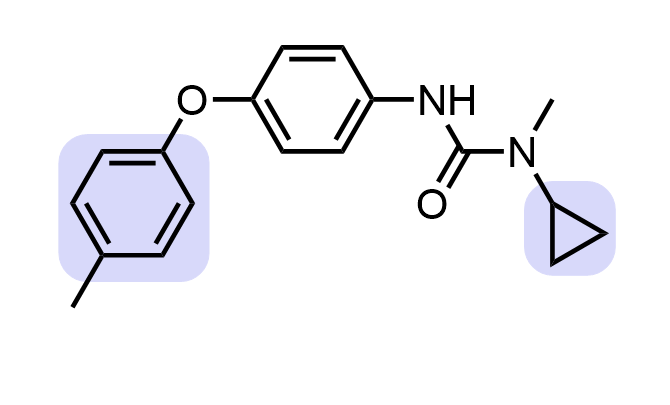}}
    \\
    {\small LogP: 3.66} & {\small LogP: 5.03} & {\small LogP: -0.36} & {\small LogP: 0.72} & {\small LogP: 2.37} & {\small LogP: 4.41}\\[5pt]
    \bottomrule
    \end{tabular}
    \end{adjustbox}
\end{table}

\newpage
\clearpage
\null
\begin{table}[htb!]
\setlength{\tabcolsep}{10pt}
\fontsize{9}{9}\selectfont
\centering
\caption{\small
Visualization of text-based editing on permeability, measured by the topological polar surface area (tPSA) of the molecules. Generally, molecules with smaller tPSA are more permeable. For generating molecules with high permeability, we can remove functional groups or heterocycles with high polarity from the input molecule, such as amides, sulfonamides, ureas, nitro groups, and nitrogen-containing arenes. For generating molecules with low permeability, we can make opposite modifications to the input molecule. The pink and blue regions highlight the modified structure in the input and output molecules, respectively.
}
\vspace{-2ex}
    \begin{adjustbox}{max width=\textwidth}
    \small
    \begin{tabular}{cccccc}
    \toprule
    \multicolumn{6}{c}{Text Prompt: This molecule has \textit{high permeability}.}\\
    \cmidrule(lr){1-6}
    {Input Molecule} & {Output Molecule} & {Input Molecule} & {Output Molecule} & {Input Molecule} & {Output Molecule}\\
    \cmidrule(lr){1-1}\cmidrule(lr){2-2}\cmidrule(lr){3-3}\cmidrule(lr){4-4}\cmidrule(lr){5-5}\cmidrule(lr){6-6}
    \adjustbox{valign=c}{\includegraphics[width=0.16\linewidth]{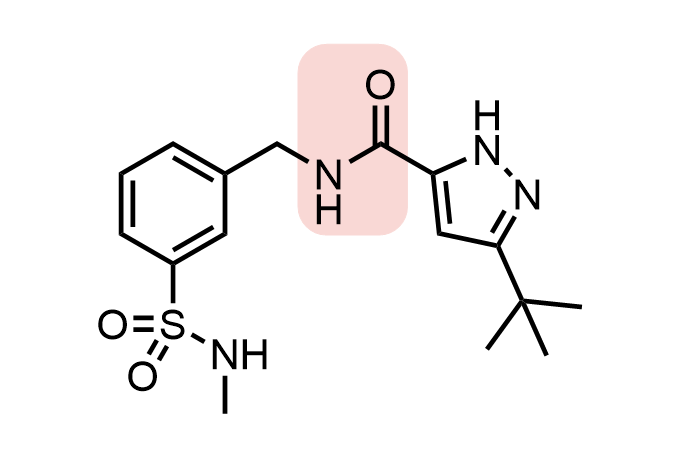}} &
    \adjustbox{valign=c}{\includegraphics[width=0.16\linewidth]{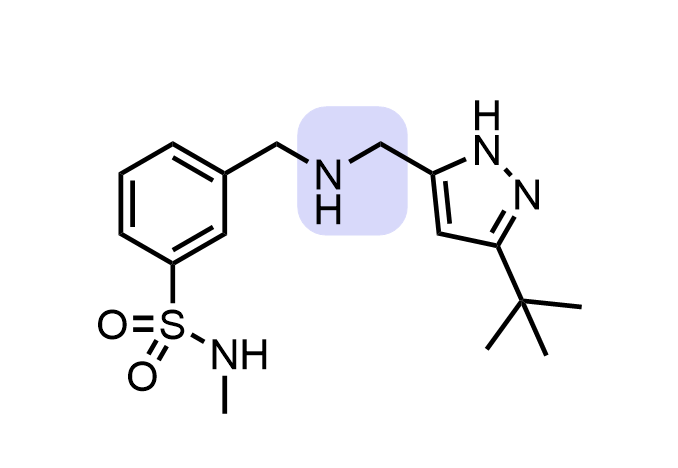}} &
    \adjustbox{valign=c}{\includegraphics[width=0.16\linewidth]{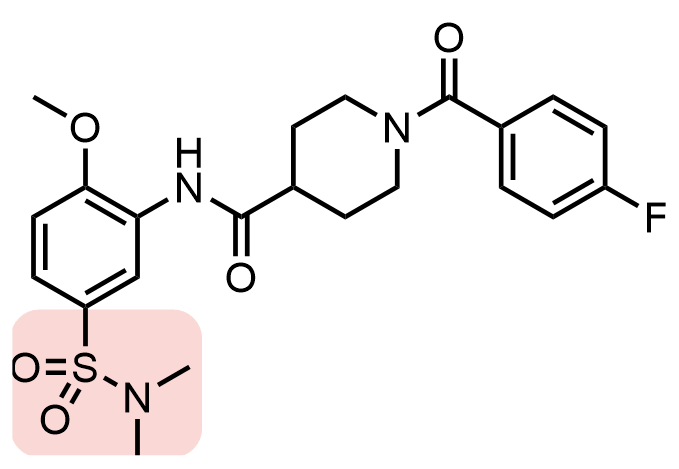}} &
    \adjustbox{valign=c}{\includegraphics[width=0.16\linewidth]{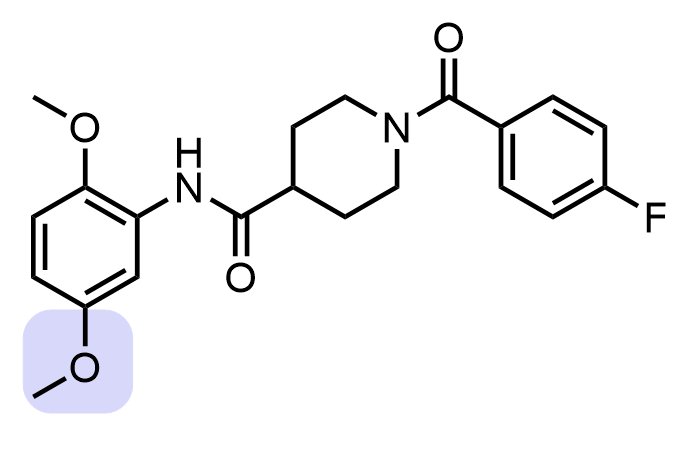}} &
    \adjustbox{valign=c}{\includegraphics[width=0.16\linewidth]{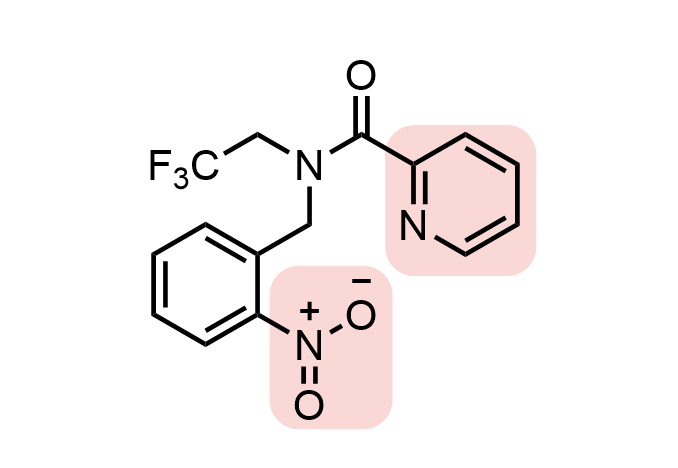}} &
    \adjustbox{valign=c}{\includegraphics[width=0.16\linewidth]{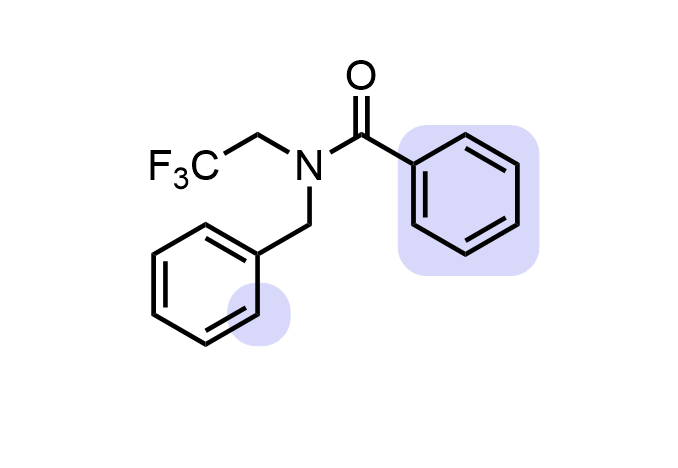}}
    \\
    {\small tPSA: 104} & {\small tPSA: 87} & {\small tPSA: 96} & {\small tPSA: 68} & {\small tPSA: 76} & {\small tPSA: 20}\\[5pt]
    \toprule
    \multicolumn{6}{c}{Text Prompt: This molecule has \textit{low permeability}.}\\
    \cmidrule(lr){1-6}
    {Input Molecule} & {Output Molecule} & {Input Molecule} & {Output Molecule} & {Input Molecule} & {Output Molecule}\\
    \cmidrule(lr){1-1}\cmidrule(lr){2-2}\cmidrule(lr){3-3}\cmidrule(lr){4-4}\cmidrule(lr){5-5}\cmidrule(lr){6-6}
    \adjustbox{valign=c}{\includegraphics[width=0.16\linewidth]{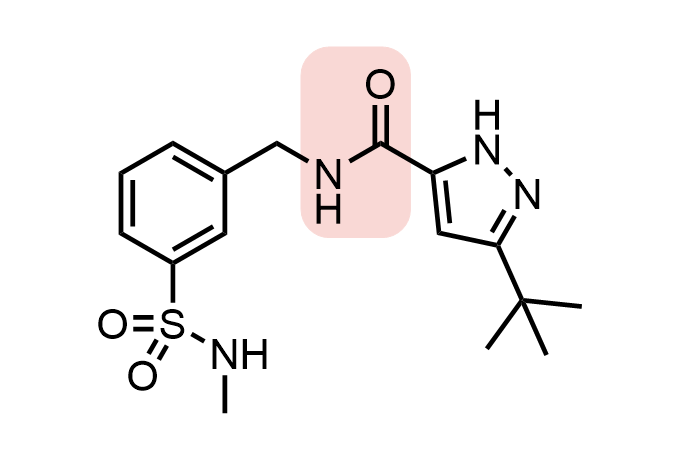}} &
    \adjustbox{valign=c}{\includegraphics[width=0.16\linewidth]{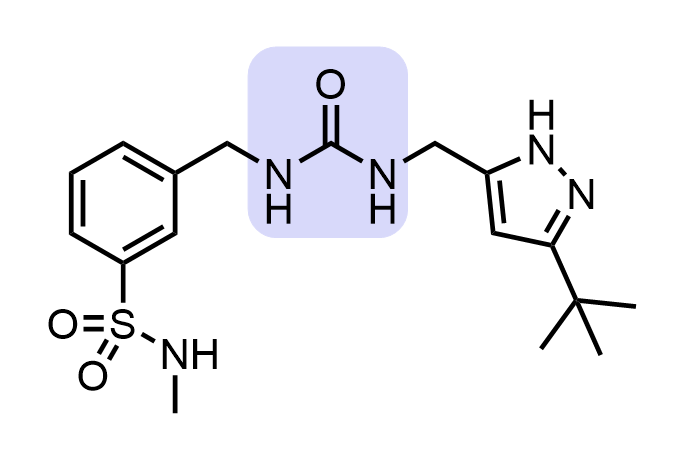}} &
    \adjustbox{valign=c}{\includegraphics[width=0.16\linewidth]{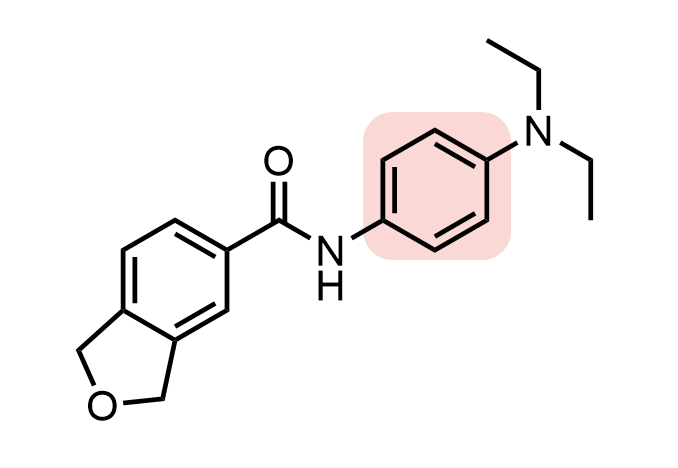}} &
    \adjustbox{valign=c}{\includegraphics[width=0.16\linewidth]{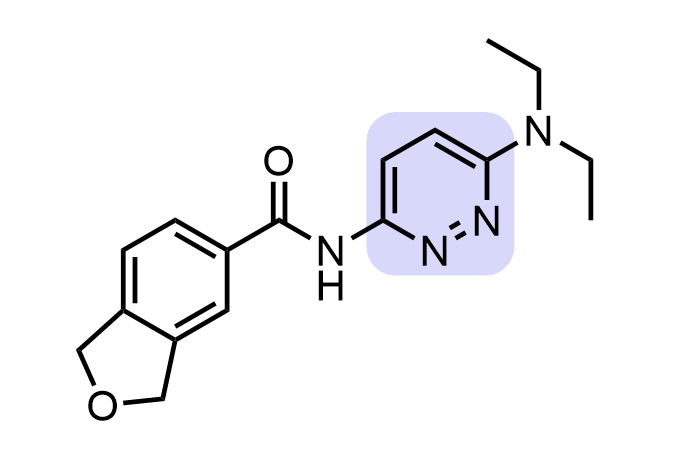}} &
    \adjustbox{valign=c}{\includegraphics[width=0.16\linewidth]{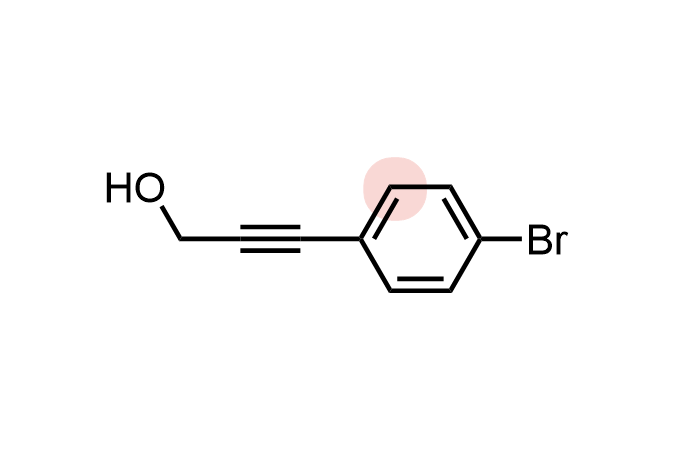}} &
    \adjustbox{valign=c}{\includegraphics[width=0.16\linewidth]{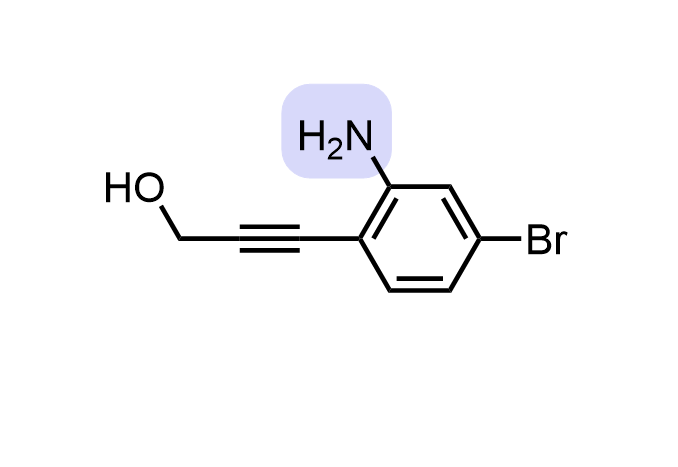}}
    \\
    {\small tPSA: 104} & {\small tPSA: 116} & {\small tPSA: 42} & {\small tPSA: 67} & {\small tPSA: 20} & {\small tPSA: 46}\\[5pt]
    \bottomrule
    \end{tabular}
    \end{adjustbox}
\end{table}

\begin{table}[htb!]
\setlength{\tabcolsep}{10pt}
\fontsize{9}{9}\selectfont
\centering
\caption{\small
Visualization of text-based editing on hydrogen bond acceptors (HBA) and hydrogen bond donors (HBD). For generating molecules with more HBA, we can add heteroatoms to the input molecule such as oxygen, nitrogen, and sulfur, or replace existing groups with heteroatom-containing structural motifs. For generating molecules with more HBD, we can add heteroatoms that bear attached hydrogens, such as functional groups like amines, and heterocycles like pyrroles. The pink and blue regions highlight the modified structure in the input and output molecules, respectively.
}
\vspace{-2ex}
    \begin{adjustbox}{max width=\textwidth}
    \small
    \begin{tabular}{cccccc}
    \toprule
    \multicolumn{6}{c}{Text Prompt: This molecule has \textit{more hydrogen bond acceptors}.}\\
    \cmidrule(lr){1-6}
    {Input Molecule} & {Output Molecule} & {Input Molecule} & {Output Molecule} & {Input Molecule} & {Output Molecule}\\
    \cmidrule(lr){1-1}\cmidrule(lr){2-2}\cmidrule(lr){3-3}\cmidrule(lr){4-4}\cmidrule(lr){5-5}\cmidrule(lr){6-6}
    \adjustbox{valign=c}{\includegraphics[width=0.16\linewidth]{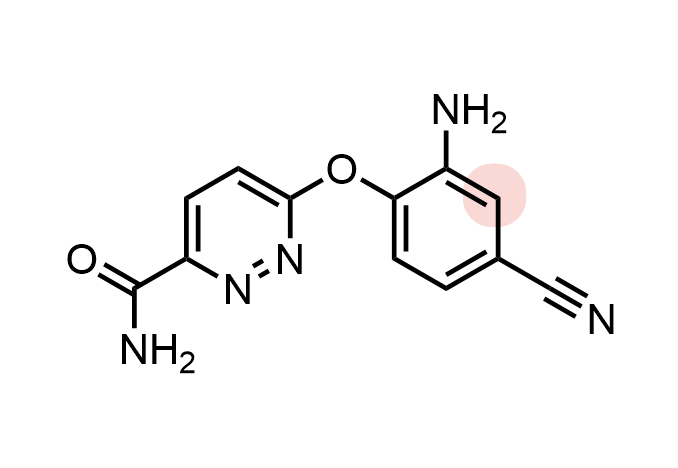}} &
    \adjustbox{valign=c}{\includegraphics[width=0.16\linewidth]{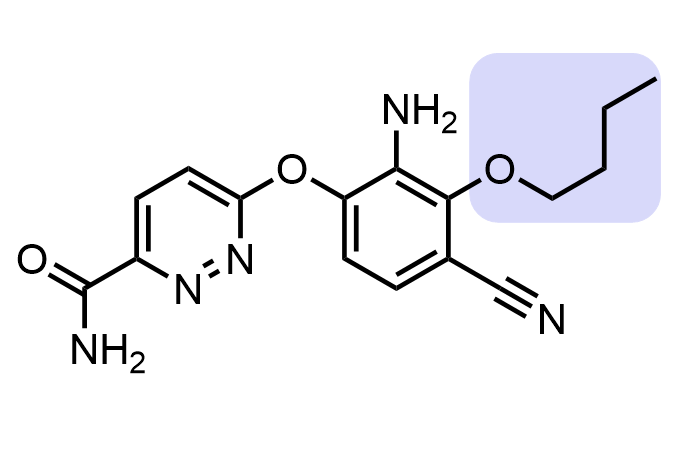}} &
    \adjustbox{valign=c}{\includegraphics[width=0.16\linewidth]{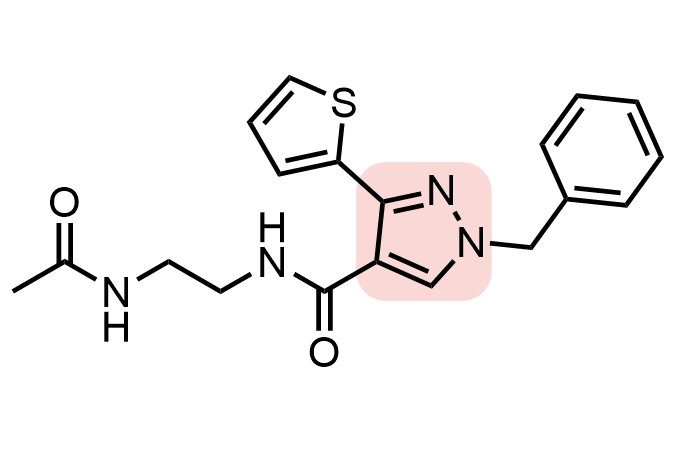}} &
    \adjustbox{valign=c}{\includegraphics[width=0.16\linewidth]{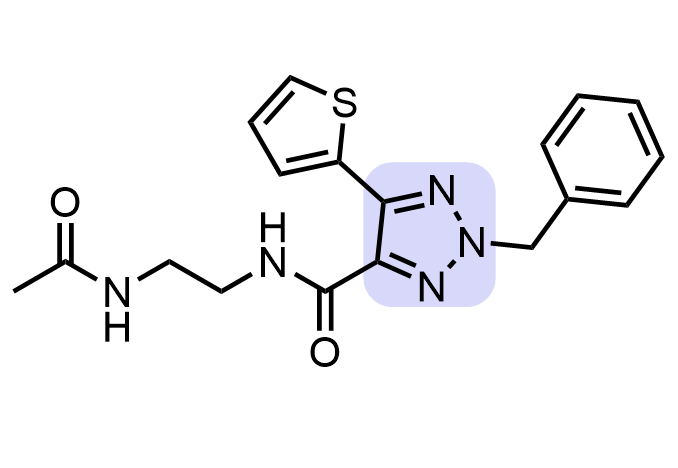}} &
    \adjustbox{valign=c}{\includegraphics[width=0.16\linewidth]{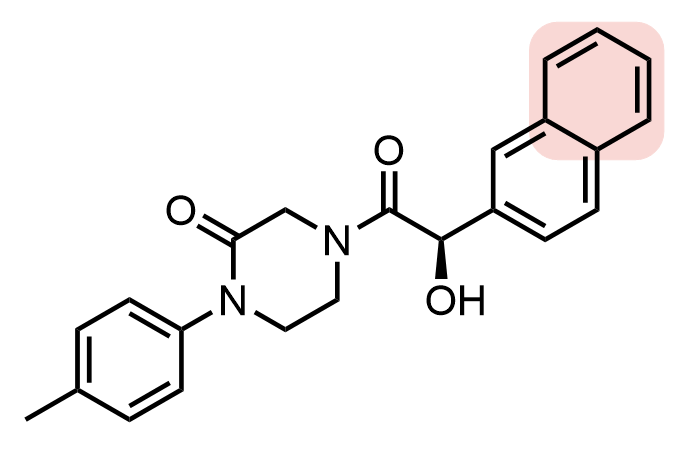}} &
    \adjustbox{valign=c}{\includegraphics[width=0.16\linewidth]{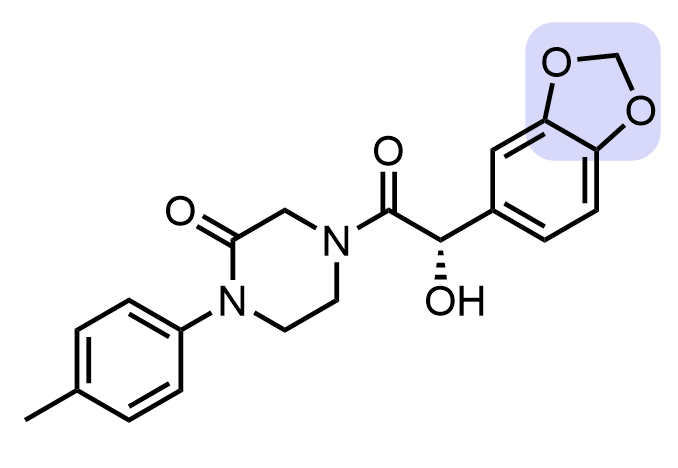}}
    \\
    {\small HBA: 6} & {\small HBA: 7} & {\small HBA: 5} & {\small HBA: 6} & {\small HBA: 3} & {\small HBA: 5}\\[5pt]
    \toprule
    \multicolumn{6}{c}{Text Prompt: This molecule has \textit{more hydrogen bond donors}.}\\
    \cmidrule(lr){1-6}
    {Input Molecule} & {Output Molecule} & {Input Molecule} & {Output Molecule} & {Input Molecule} & {Output Molecule}\\
    \cmidrule(lr){1-1}\cmidrule(lr){2-2}\cmidrule(lr){3-3}\cmidrule(lr){4-4}\cmidrule(lr){5-5}\cmidrule(lr){6-6}
    \adjustbox{valign=c}{\includegraphics[width=0.16\linewidth]{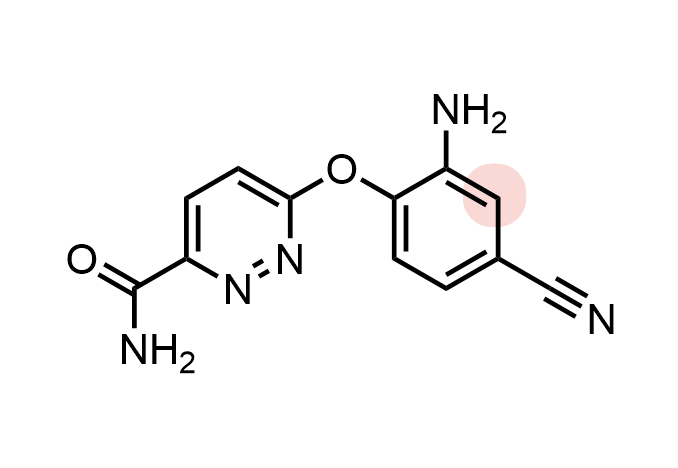}} &
    \adjustbox{valign=c}{\includegraphics[width=0.16\linewidth]{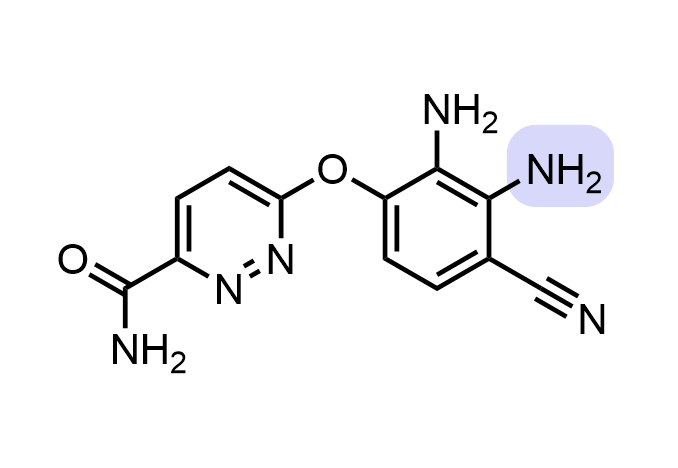}} &
    \adjustbox{valign=c}{\includegraphics[width=0.16\linewidth]{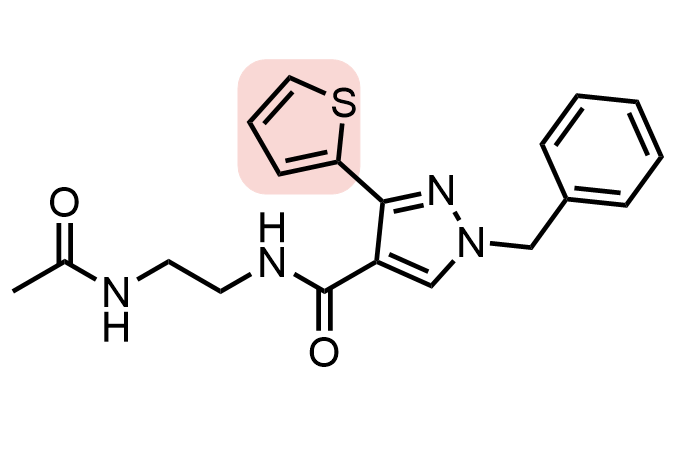}} &
    \adjustbox{valign=c}{\includegraphics[width=0.16\linewidth]{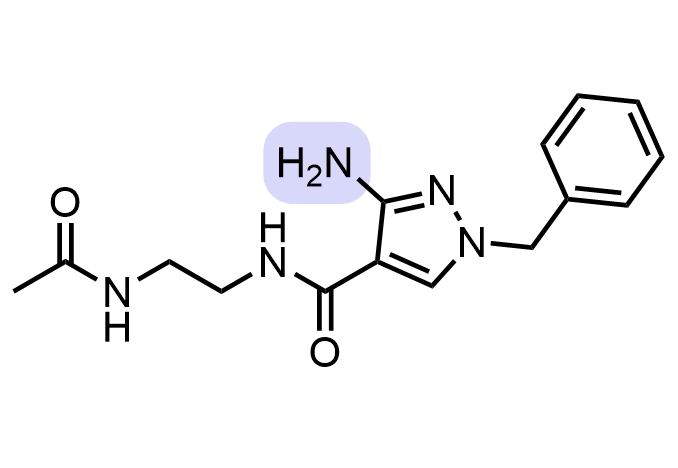}} &
    \adjustbox{valign=c}{\includegraphics[width=0.16\linewidth]{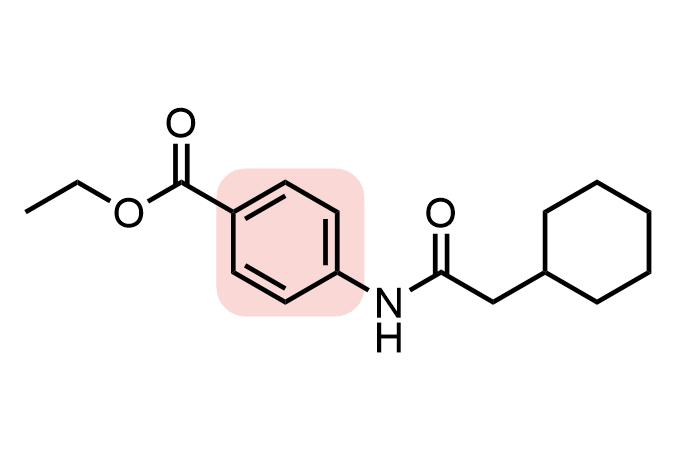}} &
    \adjustbox{valign=c}{\includegraphics[width=0.16\linewidth]{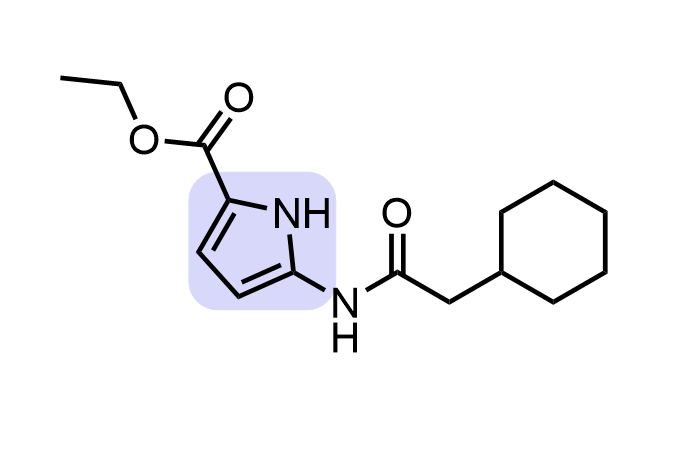}}
    \\
    {\small HBD: 2} & {\small HBD: 3} & {\small HBD: 2} & {\small HBD: 3} & {\small HBD: 1} & {\small HBD: 2}\\[5pt]
    \bottomrule
    \end{tabular}
    \end{adjustbox}
\end{table}

\newpage
\clearpage
%%%%%%%%%%%%%%%%%%%%%%%%%%%%%%%%%%%%%%%%%%%%%%%%%%
\subsection{Multi-objective Molecule Editing} \label{sec:app:multi_objective_property_editing}
We then consider \six{} multi-objective properties for molecule editing. As shown in the Methods section, the definitions of the satisfaction function and threshold $\Delta$ are based on each task specifically. First, for each single-objective, we follow the evaluation metric in~\Cref{sec:app:single_objective_property_editing}, including the solubility, permeability, and the number of HBA and HBD. Then for the multi-objective evaluation, we consider \two{} cases:
\begin{itemize}[noitemsep,topsep=0pt]
    \item The \textbf{simple} case with the loose thresholds, such as threshold 0 and 0 for solubility and permeability simultaneously.
    \item The \textbf{challenging} case with strict thresholds, such as threshold 0.5 and 1 for solubility and HBA/HBD simultaneously and threshold 0.5 and 10 for solubility and permeability simultaneously.
\end{itemize}
Then a successful hit needs to satisfy both conditions simultaneously. Below we show both the quantitative and qualitative results on \six{} multi-objective property molecule editing results.

\begin{table}[htb!]
\caption{
\small
Results on \six{} multi-objective molecule editing. The inputs are 200 molecules randomly sampled from ZINC, and the evaluation is the hit ratio of the property change. The latent optimization is text-based molecule editing with \model{}, with the SMILES string and the molecular graph, respectively.
}
% \label{tab:main_editing_composite_drug_design_properties}
\centering
\vspace{-2ex}
\begin{adjustbox}{max width=\textwidth}
\begin{tabular}{l l c c c c c c c}
\toprule
 & & \multicolumn{4}{c}{baseline} & \multicolumn{2}{c}{latent optimization}\\
\cmidrule(lr){3-6} \cmidrule(lr){7-8}
 & $\Delta$ & Random & PCA & High Variance & GS-Mutate & SMILES & Graph\\
\midrule
\multirow{2}{*}{\makecell[l]{This molecule is \textit{soluble in water}\\and has \textit{more hydrogen bond acceptors}.}} & 0 -- 0 & 9.88 $\pm$ 1.03 & 8.64 $\pm$ 2.06 & 9.09 $\pm$ 1.25 & 14.00 $\pm$ 2.48 & 27.87 $\pm$ 3.86 & 27.43 $\pm$ 3.41\\
 & 0.5 -- 1 & 0.23 $\pm$ 0.33 & 0.45 $\pm$ 0.64 & 0.22 $\pm$ 0.31 & 0.67 $\pm$ 0.62 & 8.80 $\pm$ 0.04 & 11.10 $\pm$ 1.80\\
\midrule
\multirow{2}{*}{\makecell[l]{This molecule is \textit{insoluble in water}\\and has \textit{more hydrogen bond acceptors}.}} & 0 -- 0 & 2.99 $\pm$ 0.38 & 2.00 $\pm$ 0.58 & 2.45 $\pm$ 0.67 & 7.17 $\pm$ 0.85 & 8.55 $\pm$ 2.75 & 8.21 $\pm$ 0.81\\
 & 0.5 -- 1 & 0.45 $\pm$ 0.32 & 0.00 $\pm$ 0.00 & 0.22 $\pm$ 0.31 & 0.17 $\pm$ 0.24 & 2.93 $\pm$ 0.30 & 0.00 $\pm$ 0.00\\
\midrule
\multirow{2}{*}{\makecell[l]{This molecule is \textit{soluble in water}\\and has \textit{more hydrogen bond donors}.}} & 0 -- 0 & 2.28 $\pm$ 1.15 & 2.23 $\pm$ 1.16 & 4.44 $\pm$ 0.58 & 13.83 $\pm$ 2.95 & 33.51 $\pm$ 4.08 & 49.23 $\pm$ 1.71\\
 & 0.5 -- 1 & 0.00 $\pm$ 0.00 & 0.00 $\pm$ 0.00 & 0.00 $\pm$ 0.00 & 0.00 $\pm$ 0.00 & 9.98 $\pm$ 1.03 & 23.94 $\pm$ 1.09\\
\midrule
\multirow{2}{*}{\makecell[l]{This molecule is \textit{insoluble in water}\\and has \textit{more hydrogen bond donors}.}} & 0 -- 0 & 0.69 $\pm$ 0.58 & 1.96 $\pm$ 0.87 & 1.79 $\pm$ 0.66 & 5.67 $\pm$ 0.62 & 17.03 $\pm$ 2.75 & 14.42 $\pm$ 3.43\\
 & 0.5 -- 1 & 0.00 $\pm$ 0.00 & 0.00 $\pm$ 0.00 & 0.00 $\pm$ 0.00 & 0.00 $\pm$ 0.00 & 2.59 $\pm$ 1.14 & 3.84 $\pm$ 0.71\\
\midrule
\multirow{2}{*}{\makecell[l]{This molecule is \textit{soluble in water}\\and has \textit{high permeability}.}} & 0 -- 0 & 5.06 $\pm$ 1.21 & 3.53 $\pm$ 0.38 & 4.88 $\pm$ 2.21 & 8.17 $\pm$ 1.03 & 35.69 $\pm$ 3.19 & 39.74 $\pm$ 2.26\\
 & 0.5 -- 10 & 1.16 $\pm$ 0.68 & 0.67 $\pm$ 0.55 & 0.66 $\pm$ 0.54 & 0.00 $\pm$ 0.00 & 19.15 $\pm$ 0.73 & 22.66 $\pm$ 1.90\\
\midrule
\multirow{2}{*}{\makecell[l]{This molecule is \textit{soluble in water}\\and has \textit{low permeability}.}} & 0 -- 0 & 12.17 $\pm$ 1.05 & 10.43 $\pm$ 2.88 & 13.08 $\pm$ 2.28 & 19.83 $\pm$ 2.46 & 44.35 $\pm$ 0.68 & 30.87 $\pm$ 0.62\\
 & 0.5 -- 10 & 6.20 $\pm$ 0.64 & 6.23 $\pm$ 2.31 & 6.67 $\pm$ 0.53 & 4.83 $\pm$ 0.85 & 28.67 $\pm$ 2.22 & 20.06 $\pm$ 1.26\\ 
\bottomrule
\end{tabular}
\end{adjustbox}
\end{table}

\begin{table}[htb]
\setlength{\tabcolsep}{10pt}
\fontsize{9}{9}\selectfont
\centering
\caption{\small
Visualization of text-based editing on multi-objective (compositionality) properties: solubility and hydrogen bond donors (HBD), measured by LogP and number of HBD of the molecules. Molecules with more HBD are likely also soluble in water, such as replacing hydrophobic groups (benzene, thiophene, bromide, etc.) with polar groups or rings containing hydrogen-attached heteroatoms (alcohol, azaindole, carboxylic acid, etc.) in the input molecules. Nevertheless, we can add HBD to the input molecule while reducing its solubility, such as replacing high-polarity structural motifs (amide, lactone, etc.) with less hydrophilic HBD (indole, thiol, etc.) in the input molecules. The pink and blue regions highlight the modified structure in the input and output molecules, respectively.
}
\vspace{-2ex}
    \begin{adjustbox}{max width=\textwidth}
    \small
    \begin{tabular}{cccccc}
    \toprule
    \multicolumn{6}{c}{Text Prompt: This molecule is \textit{soluble in water} and has \textit{more hydrogen bond donors}.}\\
    \cmidrule(lr){1-6}
    {Input Molecule} & {Output Molecule} & {Input Molecule} & {Output Molecule} & {Input Molecule} & {Output Molecule}\\
    \cmidrule(lr){1-1}\cmidrule(lr){2-2}\cmidrule(lr){3-3}\cmidrule(lr){4-4}\cmidrule(lr){5-5}\cmidrule(lr){6-6}
    \adjustbox{valign=c}{\includegraphics[width=0.16\linewidth]{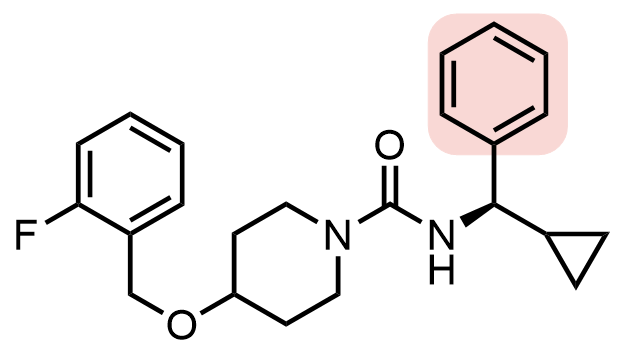}} &
    \adjustbox{valign=c}{\includegraphics[width=0.16\linewidth]{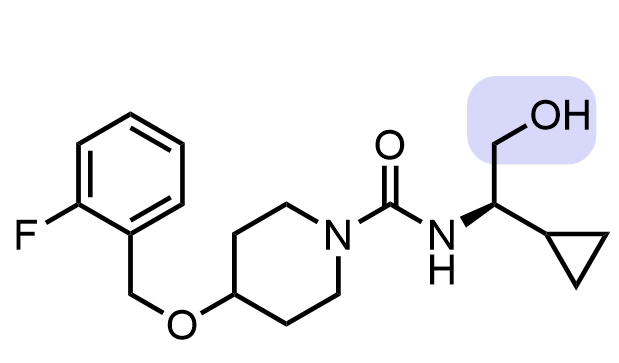}} &
    \adjustbox{valign=c}{\includegraphics[width=0.16\linewidth]{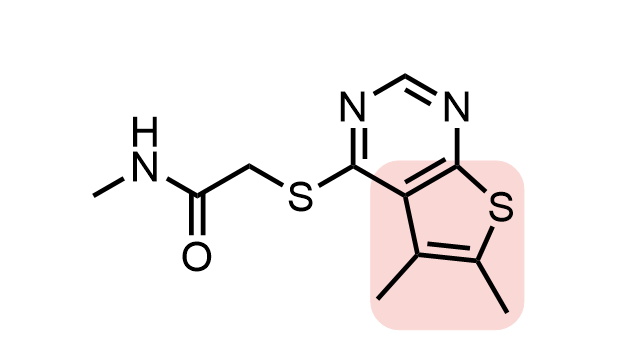}} &
    \adjustbox{valign=c}{\includegraphics[width=0.16\linewidth]{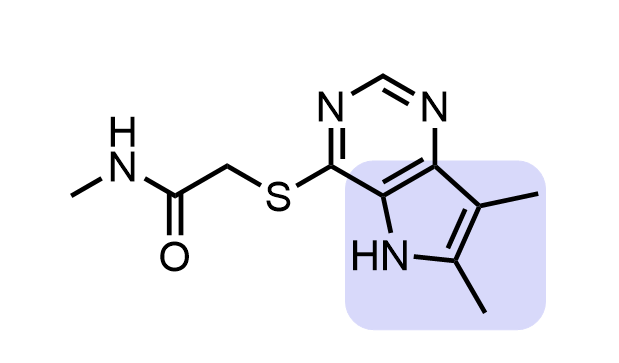}} &
    \adjustbox{valign=c}{\includegraphics[width=0.16\linewidth]{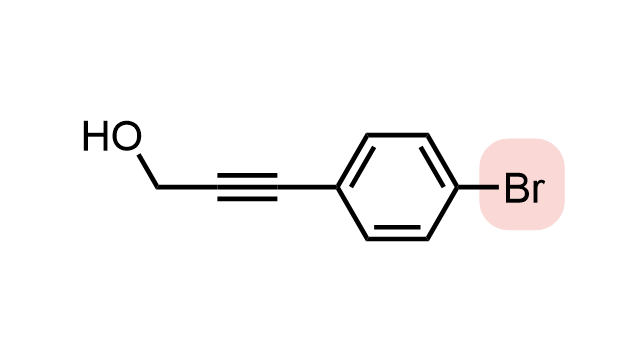}} &
    \adjustbox{valign=c}{\includegraphics[width=0.16\linewidth]{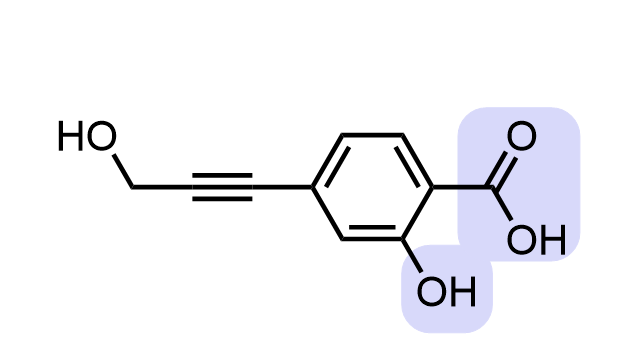}}
    \\
    {\small LogP: 4.67, HBD: 1} & {\small LogP: 2.29, HBD: 2} & {\small LogP: 2.15, HBD: 1} & {\small LogP: 1.41, HBD: 2} & {\small LogP: 1.79, HBD: 1} & {\small LogP: 0.43, HBD: 3}\\[5pt]
    \toprule
    \multicolumn{6}{c}{Text Prompt: This molecule is \textit{insoluble in water} and has \textit{more hydrogen bond donors}.}\\
    \cmidrule(lr){1-6}
    {Input Molecule} & {Output Molecule} & {Input Molecule} & {Output Molecule} & {Input Molecule} & {Output Molecule}\\
    \cmidrule(lr){1-1}\cmidrule(lr){2-2}\cmidrule(lr){3-3}\cmidrule(lr){4-4}\cmidrule(lr){5-5}\cmidrule(lr){6-6}
    \adjustbox{valign=c}{\includegraphics[width=0.16\linewidth]{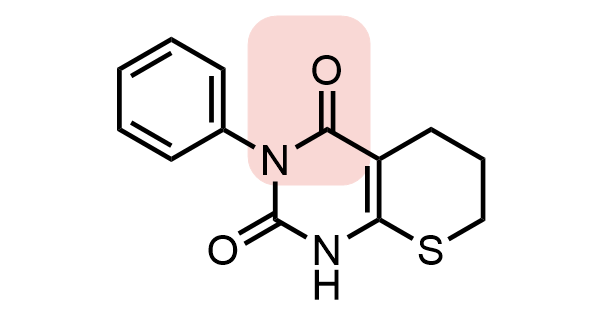}} &
    \adjustbox{valign=c}{\includegraphics[width=0.16\linewidth]{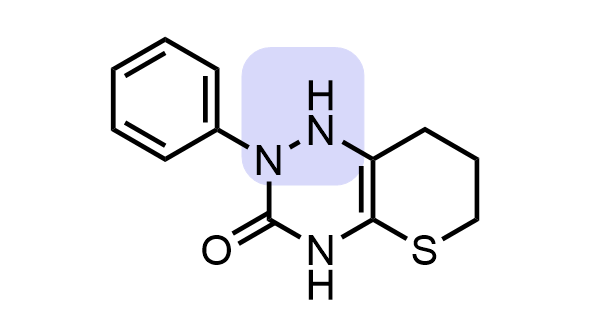}} &
    \adjustbox{valign=c}{\includegraphics[width=0.16\linewidth]{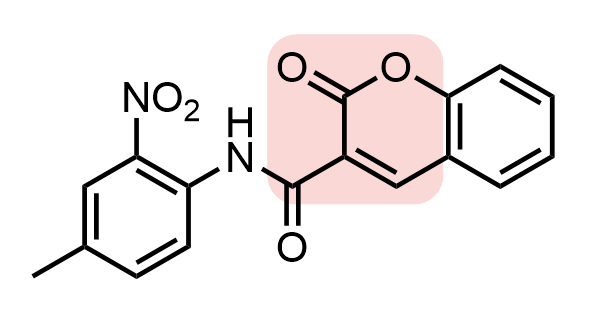}} &
    \adjustbox{valign=c}{\includegraphics[width=0.16\linewidth]{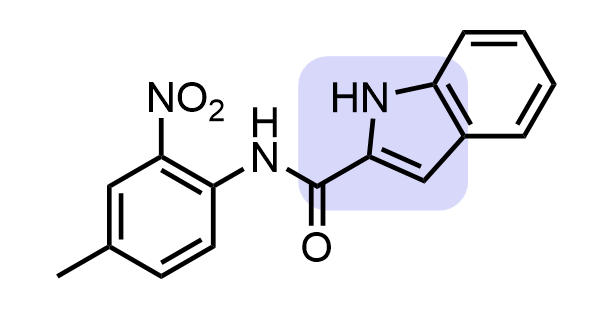}} &
    \adjustbox{valign=c}{\includegraphics[width=0.16\linewidth]{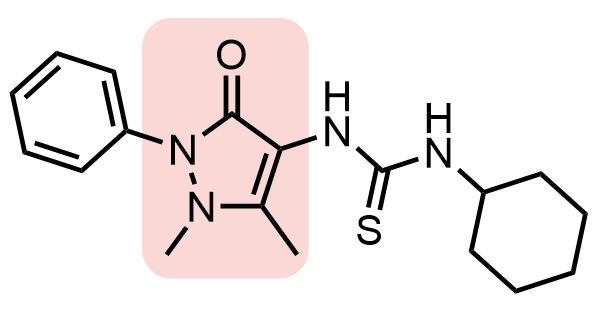}} &
    \adjustbox{valign=c}{\includegraphics[width=0.16\linewidth]{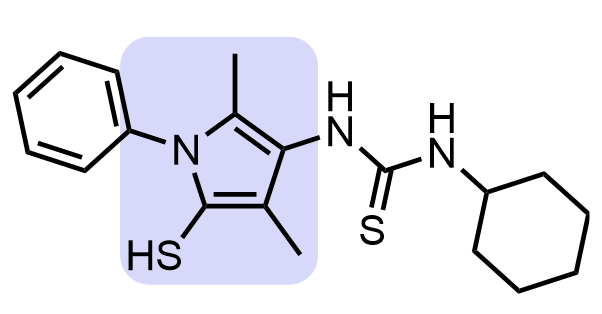}}
    \\
    {\small LogP: 1.56, HBD: 1} & {\small LogP: 2.42, HBD: 2} & {\small LogP: 3.26, HBD: 1} & {\small LogP: 3.64, HBD: 2} & {\small LogP: 3.10, HBD: 2} & {\small LogP: 5.00, HBD: 3}\\[5pt]
    \bottomrule
    \end{tabular}
    \end{adjustbox}
\end{table}

\newpage
\clearpage
\null
\begin{table}[t]
\setlength{\tabcolsep}{10pt}
\fontsize{9}{9}\selectfont
\centering
\caption{\small
Visualization of text-based editing on multi-objective (compositionality) properties: solubility and permeability, measured by LogP and tPSA of the molecules. Molecules with low permeability are likely also soluble in water, such as adding polar functional groups ({\eg}, amide, amine) and removing hydrocarbons ({\eg}, methyl, phenyl) with regard to the input molecules. Nevertheless, we can increase both the solubility and permeability of the molecule, such as removing hydrocarbons and polar moieties simultaneously or reducing the size of the heterocycles ({\eg}, \textit{[1,2]oxazolo[5,4-b]pyridine} to imidazole) in the input molecules. The pink and blue regions highlight the modified structure in the input and output molecules, respectively.
}
\vspace{-2ex}
    \begin{adjustbox}{max width=\textwidth}
    \small
    \begin{tabular}{cccccc}
    \toprule
    \multicolumn{6}{c}{Text Prompt: This molecule is \textit{soluble in water} and has \textit{low permeability}.}\\
    \cmidrule(lr){1-6}
    {Input Molecule} & {Output Molecule} & {Input Molecule} & {Output Molecule} & {Input Molecule} & {Output Molecule}\\
    \cmidrule(lr){1-1}\cmidrule(lr){2-2}\cmidrule(lr){3-3}\cmidrule(lr){4-4}\cmidrule(lr){5-5}\cmidrule(lr){6-6}
    \adjustbox{valign=c}{\includegraphics[width=0.16\linewidth]{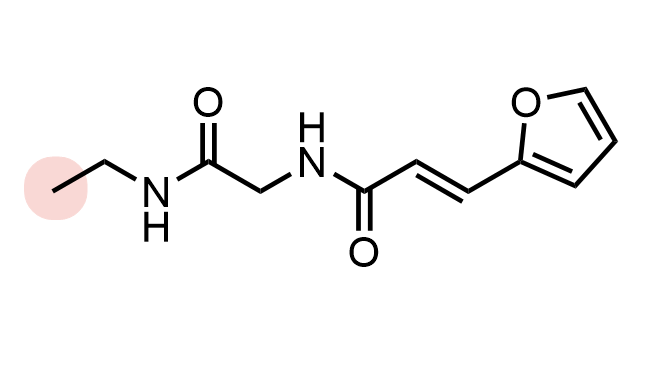}} &
    \adjustbox{valign=c}{\includegraphics[width=0.16\linewidth]{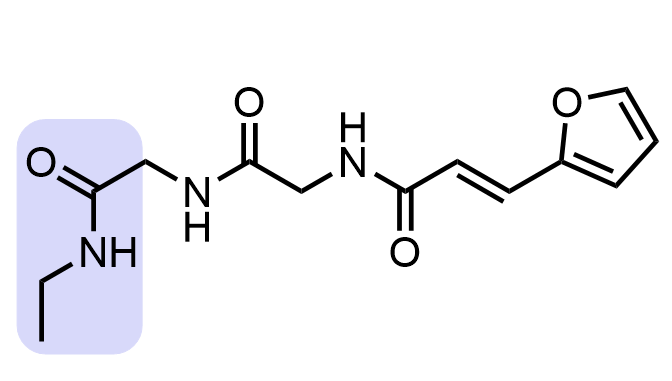}} &
    \adjustbox{valign=c}{\includegraphics[width=0.16\linewidth]{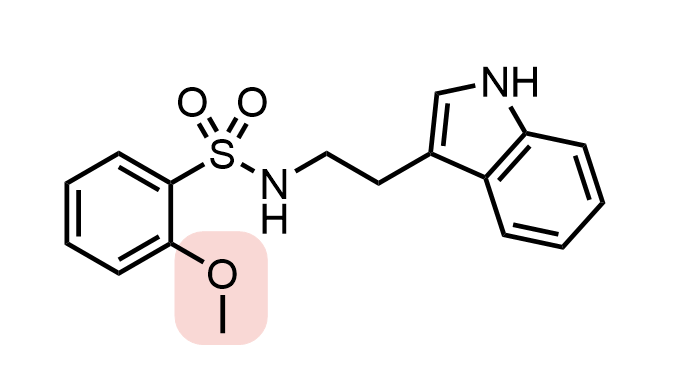}} &
    \adjustbox{valign=c}{\includegraphics[width=0.16\linewidth]{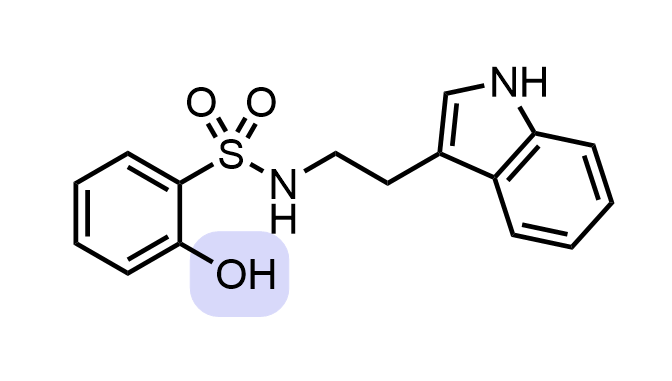}} &
    \adjustbox{valign=c}{\includegraphics[width=0.16\linewidth]{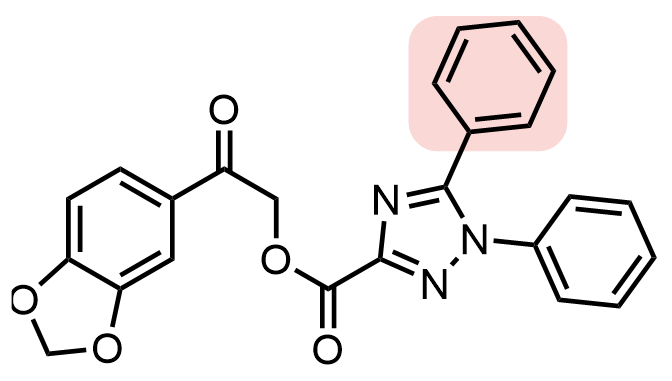}} &
    \adjustbox{valign=c}{\includegraphics[width=0.16\linewidth]{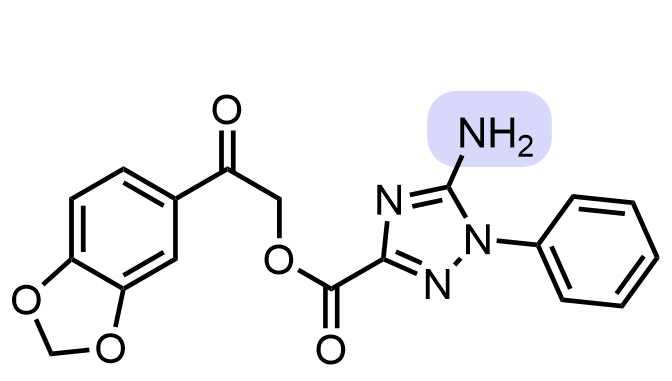}}
    \\
    {\small LogP: 0.55, tPSA: 71} & {\small LogP: -0.34, tPSA: 100} & {\small LogP: 2.70, tPSA: 71} & {\small LogP: 2.39, tPSA: 82} & {\small LogP: 3.70, tPSA: 93} & {\small LogP: 1.62, tPSA: 119}\\[5pt]
    \toprule
    \multicolumn{6}{c}{Text Prompt: This molecule is \textit{soluble in water} and has \textit{high permeability}.}\\
    \cmidrule(lr){1-6}
    {Input Molecule} & {Output Molecule} & {Input Molecule} & {Output Molecule} & {Input Molecule} & {Output Molecule}\\
    \cmidrule(lr){1-1}\cmidrule(lr){2-2}\cmidrule(lr){3-3}\cmidrule(lr){4-4}\cmidrule(lr){5-5}\cmidrule(lr){6-6}
    \adjustbox{valign=c}{\includegraphics[width=0.16\linewidth]{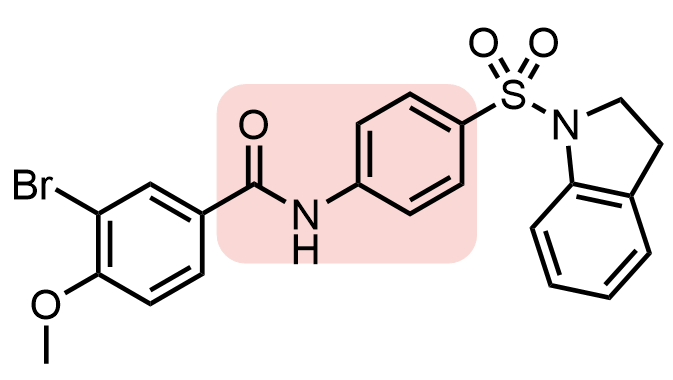}} &
    \adjustbox{valign=c}{\includegraphics[width=0.16\linewidth]{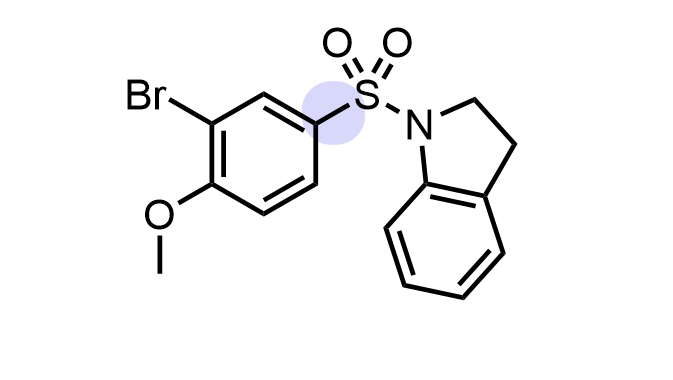}} &
    \adjustbox{valign=c}{\includegraphics[width=0.16\linewidth]{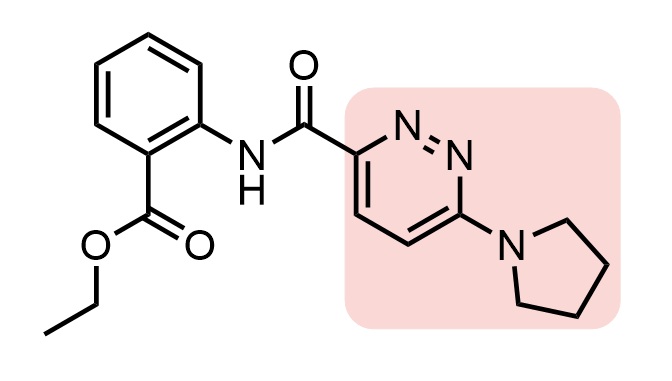}} &
    \adjustbox{valign=c}{\includegraphics[width=0.16\linewidth]{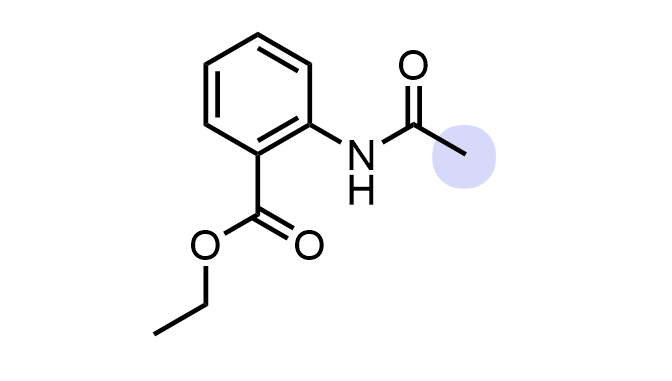}} &
    \adjustbox{valign=c}{\includegraphics[width=0.16\linewidth]{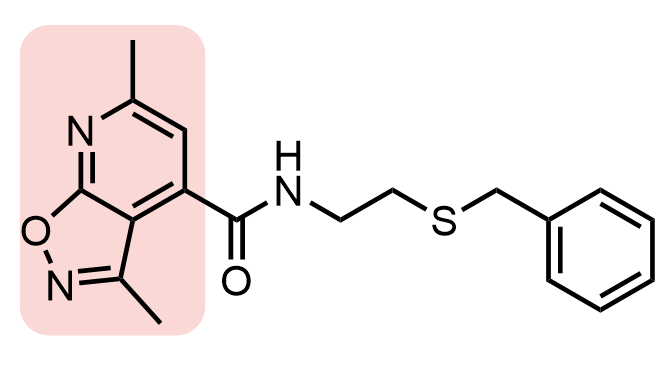}} &
    \adjustbox{valign=c}{\includegraphics[width=0.16\linewidth]{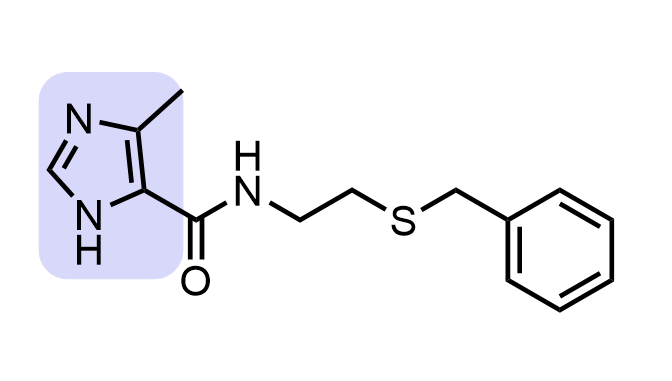}}
    \\
    {\small LogP: 4.46, tPSA: 76} & {\small LogP: 3.21, tPSA: 47} & {\small LogP: 2.51, tPSA: 84} & {\small LogP: 1.82, tPSA: 55} & {\small LogP: 3.50, tPSA: 68} & {\small LogP: 2.38, tPSA: 58}\\[5pt]
    \bottomrule
    \end{tabular}
    \end{adjustbox}
\end{table}

\newpage
\clearpage
%%%%%%%%%%%%%%%%%%%%%%%%%%%%%%%%%%%%%%%%%%%%%%%%%%
\subsection{Binding-affinity-based Molecule Editing} \label{sec:app:ChEMBL_assay_editing}
We further apply text-based editing on the binding affinity assays. In specific, we take \six{} binding affinity tasks from ChEMBL~\cite{mendez2018chembl}. Each assay has a textual description, as listed in~\Cref{tab:chembl_assay_description}.

\begin{table}[htb]
\caption{
\small
ChEMBL assay descriptions.
}
\label{tab:chembl_assay_description}
\centering
\vspace{-2ex}
\begin{adjustbox}{max width=\textwidth}
\begin{tabular}{p{0.15\textwidth}p{0.85\textwidth}}
\toprule
ChEMBL ID & Assay Description\\
\midrule
1613777 & This molecule is tested positive in \textit{an assay that are inhibitors and substrates of an enzyme protein. It uses molecular oxygen inserting one oxygen atom into a substrate and reducing the second into a water molecule}.\\
1613797 & This molecule is tested positive in \textit{an assay for Anthrax Lethal, which acts as a protease that cleaves the N-terminal of most dual specificity mitogen-activated protein kinase kinases}.\\
2114713 & This molecule is tested positive in \textit{an assay for Activators of ClpP, which cleaves peptides in various proteins in a process that requires ATP hydrolysis and has a limited peptidase activity in the absence of ATP-binding subunits}.\\
1613838 & This molecule is tested positive in \textit{an assay for activators involved in the transport of proteins between the endosomes and the trans Golgi network}.\\
1614236 & This molecule is \textit{an inhibitor of a protein that prevents the establishment of the cellular antiviral state by inhibiting ubiquitination that triggers antiviral transduction signal and inhibits post-transcriptional processing of cellular pre-mRNA}.\\
1613903 & This molecule is tested positive \textit{in the high throughput screening assay to identify inhibitors of the SARS coronavirus 3C-like Protease, which cleaves the C-terminus of replicase polyprotein at 11 sites}.\\
\bottomrule
\end{tabular}
\end{adjustbox}
\end{table}

\begin{figure}[htb!]
\centering
\includegraphics[width=0.8\linewidth]{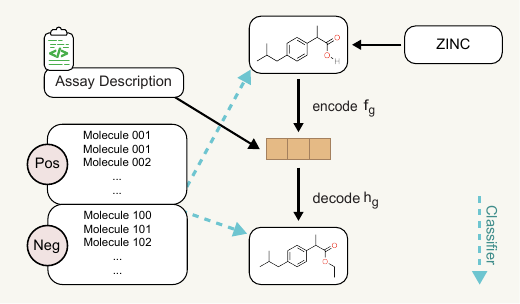}
\caption{\small
Pipeline for binding-affinity-based molecule editing. The input molecules are randomly sampled from ZINC, and the text prompt is the assay description. For evaluation, the small molecules for each assay are used to train a binary classifier, and two types of models (random forest and logistic regression) are considered.
}
\label{fig:ChEMBL_editing_pipeline}
\end{figure}

For evaluation, we follow the Methods section. Recall that each binding affinity assay can correspond to molecules with positive and negative labels. Thus, we can train a classifier on these data points, and the satisfy criteria here is if the output molecules can have higher confidence than the input molecule, where the confidence is predicted using the classifier for each task. The pipeline can be found in~\Cref{fig:ChEMBL_editing_pipeline}.

\newpage
The hit ratio results are shown in~\Cref{tab:chembl_assay_editing}. Notice that to better prove the validity of our results, we train \two{} classifiers for each assay: random forest (RF) and logistic regression (LR), with the fingerprint as featurization. the $\Delta$ is 0.
\begin{table}[htb]
\caption{
Results on \six{} ChEMBL assay editing. Each ChEMBL assay is a binary task and we train a classifier to obtain the confidence score of each molecule (input and output molecules). The inputs are 200 molecules randomly sampled from ZINC, and the evaluation is the hit ratio of the confidence change. The latent optimization is the text-based molecule editing with \model{}, with the SMILES string and the molecular graph, respectively.
}
\label{tab:chembl_assay_editing}
\vspace{-2ex}
\centering
\begin{adjustbox}{max width=\textwidth}
\begin{tabular}{c c c c c c c c}
\toprule
 & & \multicolumn{4}{c}{baseline} & \multicolumn{2}{c}{latent optimization}\\
\cmidrule(lr){3-6} \cmidrule(lr){7-8}
 ChEMBL ID & & Random & PCA & High Variance & GS-Mutate & SMILES & Graph\\
\midrule
\multirow{2}{*}{1613777} & RF & 44.99 $\pm$ 2.08 & 44.49 $\pm$ 1.22 & 44.45 $\pm$ 1.01 & 39.17 $\pm$ 3.66 & 48.70 $\pm$ 2.06 & 44.53 $\pm$ 1.60\\
 & LR & 47.34 $\pm$ 5.53 & 49.13 $\pm$ 0.86 & 49.69 $\pm$ 6.75 & 51.50 $\pm$ 2.86 & 54.09 $\pm$ 1.94 & 50.55 $\pm$ 3.14\\
\midrule
\multirow{2}{*}{1613797} & RF & 44.76 $\pm$ 2.18 & 46.25 $\pm$ 0.97 & 46.92 $\pm$ 3.34 & 46.67 $\pm$ 1.55 & 55.03 $\pm$ 2.23 & 49.03 $\pm$ 0.03\\
 & LR & 48.40 $\pm$ 3.71 & 49.92 $\pm$ 4.31 & 48.67 $\pm$ 1.64 & 49.17 $\pm$ 3.01 & 57.98 $\pm$ 3.34 & 54.95 $\pm$ 3.74\\
\midrule
\multirow{2}{*}{2114713} & RF & 39.87 $\pm$ 2.32 & 42.91 $\pm$ 2.64 & 42.19 $\pm$ 3.68 & 41.33 $\pm$ 1.25 & 49.20 $\pm$ 2.11 & 60.93 $\pm$ 2.53\\
 & LR & 51.39 $\pm$ 1.15 & 52.62 $\pm$ 1.64 & 52.24 $\pm$ 1.07 & 50.50 $\pm$ 1.47 & 56.93 $\pm$ 3.67 & 58.77 $\pm$ 2.41\\
\midrule
\multirow{2}{*}{1613838} & RF & 44.49 $\pm$ 1.48 & 44.71 $\pm$ 1.80 & 45.30 $\pm$ 2.47 & 36.00 $\pm$ 2.68 & 43.94 $\pm$ 3.75 & 49.13 $\pm$ 2.52\\
 & LR & 50.22 $\pm$ 4.23 & 49.73 $\pm$ 2.33 & 44.69 $\pm$ 2.41 & 41.33 $\pm$ 3.17 & 47.50 $\pm$ 2.28 & 56.13 $\pm$ 1.50\\
\midrule
\multirow{2}{*}{1614236} & RF & 41.33 $\pm$ 3.59 & 42.28 $\pm$ 1.91 & 42.85 $\pm$ 2.88 & 45.33 $\pm$ 1.65 & 57.90 $\pm$ 2.39 & 35.71 $\pm$ 4.19\\
 & LR & 46.57 $\pm$ 0.51 & 49.34 $\pm$ 1.80 & 50.62 $\pm$ 3.86 & 56.00 $\pm$ 1.08 & 65.78 $\pm$ 5.67 & 46.36 $\pm$ 2.53\\
\midrule
\multirow{2}{*}{1613903} & RF & 44.28 $\pm$ 0.77 & 43.83 $\pm$ 2.65 & 42.00 $\pm$ 3.19 & 46.17 $\pm$ 0.85 & 56.82 $\pm$ 3.96 & 58.70 $\pm$ 1.43\\
 & LR & 53.94 $\pm$ 3.30 & 48.63 $\pm$ 4.49 & 56.19 $\pm$ 2.51 & 56.33 $\pm$ 0.94 & 58.31 $\pm$ 2.98 & 64.64 $\pm$ 5.23\\
\bottomrule
\end{tabular}
\end{adjustbox}
\vspace{-2ex}
\end{table}

\begin{figure}[htb]
\centering
\includegraphics[width=\linewidth]{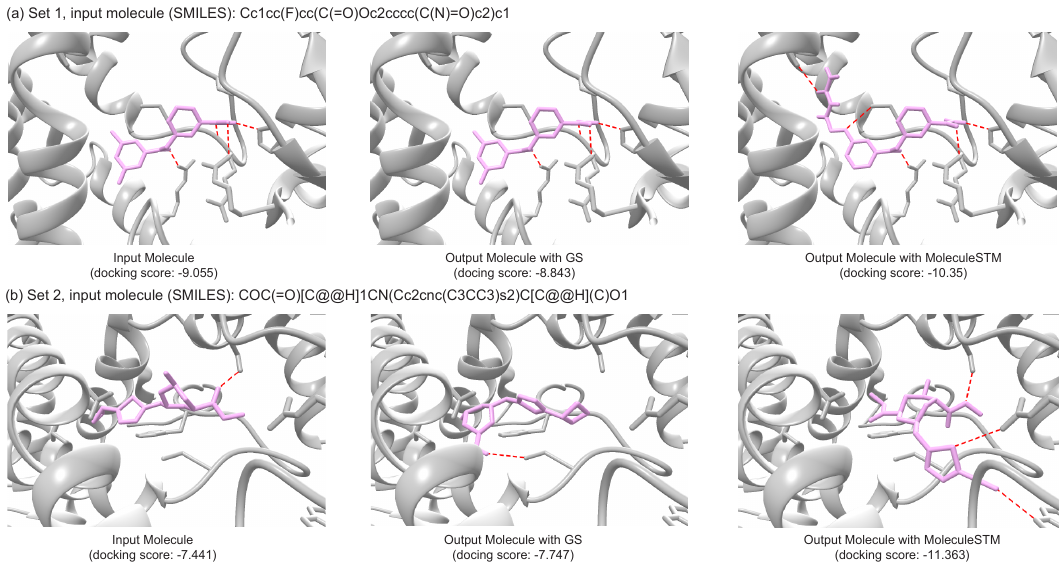}
\vspace{-4ex}
\caption{\small
Two sets of docking visualization for binding-affinity-based molecule editing. The text prompt is from ChEMBL 1613777 (``This molecule is tested positive in \textit{an assay that are inhibitors and substrates of an enzyme protein. It uses molecular oxygen inserting one oxygen atom into a substrate, and reducing the second into a water molecule}.''). For visualization, the input molecule and output molecules with GS and \model{} are displayed. It is observed that \model{} can generate molecules with the lowest docking scores (with the most Hydrogen bonds, and marked in red dashed lines). In set 1 (a), the output molecules are sharing the same molecule scaffold. In set 2 (b), the motif of the output molecule using \model{} also changes.
}
\label{fig:docking_visualization}
\vspace{-3ex}
\end{figure}

\newpage
Then we add docking for visualization in~\Cref{fig:docking_visualization}. We choose the ChEMBL 1613777 with the available PDB structure. In specific, we first extract the output molecules using \model{} with confidence (RF and LR) higher than the ones generated with baselines. Then we run the molecular docking software for the results. The details of docking settings are listed below.
\begin{itemize}[noitemsep,topsep=0pt]
    \item We use Merck molecular force field (MMFF)~\cite{halgren1996merck} provided in RDKit~\cite{landrum2013rdkit} to embed (generate) 3D conformers for each molecule. The dielectric constant is set to be 80 and the maximum iteration of optimization is 1000 for MMFF, and the up-to-5 conformers from each molecule are used for further analysis.
    \item For the binding target, we consider assay P450 (CYP) 2C19~\cite{reynald2012structural} (CHEMBL id: 1613777) and select the corresponding crystal structure available in the Protein Data Bank (PDB) (PDB id: 4GQS). Further, we take chain A for docking running. Later for the binding, the binding pockets are aligned with the original ligand in the crystal structure of PDB complexes: the center is set to (-81.48, 16.55, -41.6), and the box is (20.0, 23.0, 25.0).
    \item Then we take a preprocessing step to complement the hydrogen atoms and add partial charges. We utilize meeko v0.3.3 for small molecules and AutoDock Flexible Receptor (ADFR) suite v1.2 for proteins.
    \item For docking, we use AutoDock Vina v1.2.3~\cite{trott2010autodock}. Each molecule conformer is docked with \textit{exhausitiveness} being 32, and the pose with the best (lowest) docking score is picked and used for visualization. For visualization, we use UCSF Chimera.\looseness=-1
\end{itemize}

\clearpage
%%%%%%%%%%%%%%%%%%%%%%%%%%%%%%%%%%%%%%%%%%%%%%%%%%
\subsection{Drug Relevance Editing} \label{sec:app:common_drug_editing}
As a proof-of-concept, we further take \four{} editing tasks on common drug editing. The text prompts used here are to make the input molecules look like an existing drug, {\eg}, ``This molecule looks like \textit{Penicillin}.'' Following the Methods section, the satisfy function used is the Tanimoto similarity, and the threshold $\Delta$ takes the value of 0 and 0.05.

\begin{table}[h]
\caption{
\small
Results on \four{} common drug molecule editing. The inputs are 200 molecules randomly sampled from ZINC, and the evaluation is the hit ratio on the increase of the Tanimoto similarity with the common drug. The latent optimization is text-based molecule editing with \model{}, with the SMILES string and the molecular graph, respectively.
}
\label{tab:editing_common_drug_editing}
\centering
\vspace{-2ex}
\begin{adjustbox}{max width=\textwidth}
\begin{tabular}{l l c c c c c c c}
\toprule
 & & \multicolumn{4}{c}{baseline} & \multicolumn{2}{c}{latent optimization}\\
\cmidrule(lr){3-6} \cmidrule(lr){7-8}
 & $\Delta$ & Random & PCA & High Variance & GS-Mutate & SMILES & Graph\\
\midrule
\multirow{2}{*}{This molecule \textit{looks like Penicillin}.} & 0 & 43.61 $\pm$ 2.23 & 46.51 $\pm$ 3.02 & 44.42 $\pm$ 3.56 & 28.67 $\pm$ 0.94 & 58.13 $\pm$ 0.97 & 50.91 $\pm$ 2.80\\
 & 0.05 & 0.69 $\pm$ 0.55 & 0.23 $\pm$ 0.32 & 0.89 $\pm$ 0.30 & 0.67 $\pm$ 0.62 & 11.01 $\pm$ 0.58 & 3.64 $\pm$ 0.57\\
\midrule
\multirow{2}{*}{This molecule \textit{looks like Aspirin}.} & 0 & 43.82 $\pm$ 1.41 & 43.12 $\pm$ 5.35 & 44.63 $\pm$ 3.33 & 25.00 $\pm$ 2.16 & 40.13 $\pm$ 1.33 & 54.05 $\pm$ 3.58\\
 & 0.05 & 2.99 $\pm$ 0.38 & 3.08 $\pm$ 0.82 & 2.45 $\pm$ 0.33 & 0.33 $\pm$ 0.47 & 4.28 $\pm$ 1.22 & 10.84 $\pm$ 1.26\\
\midrule
\multirow{2}{*}{This molecule \textit{looks like Caffeine}.} & 0 & 42.71 $\pm$ 3.16 & 40.33 $\pm$ 0.71 & 40.64 $\pm$ 3.89 & 26.17 $\pm$ 1.31 & 46.08 $\pm$ 3.81 & 51.01 $\pm$ 1.22\\
 & 0.05 & 0.69 $\pm$ 0.01 & 0.23 $\pm$ 0.32 & 0.44 $\pm$ 0.31 & 0.33 $\pm$ 0.24 & 1.61 $\pm$ 0.67 & 0.61 $\pm$ 0.01\\
\midrule
\multirow{2}{*}{This molecule \textit{looks like Dopamine}.} & 0 & 42.00 $\pm$ 3.08 & 42.50 $\pm$ 2.12 & 41.33 $\pm$ 2.86 & 30.50 $\pm$ 1.63 & 47.00 $\pm$ 4.11 & 55.50 $\pm$ 2.73\\
 & 0.05 & 0.00 $\pm$ 0.00 & 0.44 $\pm$ 0.31 & 0.22 $\pm$ 0.31 & 0.83 $\pm$ 0.24 & 2.30 $\pm$ 0.44 & 6.24 $\pm$ 0.56\\
\bottomrule
\end{tabular}
\end{adjustbox}
\end{table}

%%%%%%%%%%%%%%%%%%%%%%%%%%%%%%%%%%%%%%%%%%%%%%%%%%
\subsection{Case Studies on Neighborhood Searching for Patent Drug Molecules} \label{sec:app:patent_drug_searching}
To demonstrate the utility of text-based molecule editing, we show \three{} case studies of generating approved drugs from their analogs. Lead optimization is a critical phase of drug discovery in which closely related compounds are made based on the lead molecule, aiming to improve its efficacy and DMPK (drug metabolism and pharmacokinetics) properties and ultimately identifying a drug candidate~\cite{hughes2011principles}. A text prompt calling for greater drug-like properties will thus be informative towards improving on deficiencies in the lead molecule and accelerating drug discovery research.

In specific here, the input molecules are the patented analogs of each approved drug molecule, and the input text prompt is single-objective, like the ones in~\Cref{sec:app:single_objective_property_editing}. The goal here is to check if the approved drugs can be successfully generated as the output molecules, with the structural changes consistent with the property improvement reflected in the text prompt. For example, in~\Cref{tab:case_study_on_drugs} (a), Erlotinib is successfully generated from an analog by replacing an imidazole substituent to a methoxy group~\cite{schnur1998alkynyl}. This change reflects a tPSA drop from 83 to 75, consistent with the text prompt indicating a higher permeability. \Cref{tab:case_study_on_drugs} (b) generates Celecoxib from its amino-substituted derivative~\cite{talley1998substituted}, where the removal of the amino group yields a greater intestinal permeability of the molecule leading to higher bioavailability. Bioavailability is the fraction of a drug molecule that reaches the systemic circulation, a key factor for oral drug absorption~\cite{pharmaceutics11080411}. Finally, \Cref{tab:case_study_on_drugs} (c) illustrates how potential metabolic liabilities in a molecule can be addressed via text-based editing. A text calling for a metabolically stable molecule successfully turns a trimethoxy arene to a dimethoxy arene in Donepezil~\cite{doi:10.1021/jm00024a009}, where the former represents an electron-rich aromatic compound known to undergo oxidative phase I metabolisms~\cite{guroff1967hydroxylation}.

\begin{table}[htb]
\setlength{\tabcolsep}{10pt}
\fontsize{9}{9}\selectfont
\centering
\caption{\small
Visualization on \three{} single-objective molecule editing on drug analogs that generates approved drugs based on the text prompt. The pink and blue regions highlight the modified structure in the input and output molecules, respectively.
}
\label{tab:case_study_on_drugs}
\vspace{-2ex}
    \begin{adjustbox}{max width=\textwidth}
    \small
    \begin{tabular}{cccccc}
    \toprule
    \multicolumn{2}{c}{(a) Prompt: This molecule has \textit{high permeability}.} &
    \multicolumn{2}{c}{(b) Prompt: This molecule has \textit{high bioavailability}.} & 
    \multicolumn{2}{c}{(c) Prompt: This molecule is \textit{metabolically stable}.} \\
    \cmidrule(lr){1-2} \cmidrule(lr){3-4} \cmidrule(lr){5-6} 
    {Input Molecule} & {Output Molecule} & {Input Molecule} & {Output Molecule} & {Input Molecule} & {Output Molecule}\\
    \cmidrule(lr){1-1}\cmidrule(lr){2-2}
    \cmidrule(lr){3-3}\cmidrule(lr){4-4}
    \cmidrule(lr){5-5}\cmidrule(lr){6-6}
    \adjustbox{valign=c}{\includegraphics[width=0.2\linewidth]{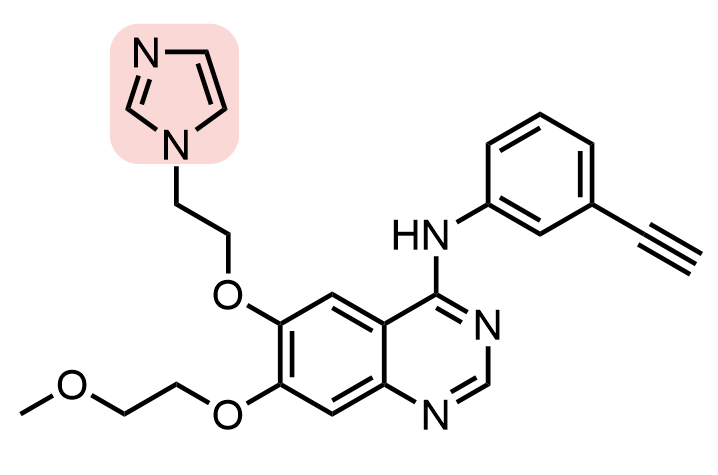}} &
    \adjustbox{valign=c}{\includegraphics[width=0.2\linewidth]{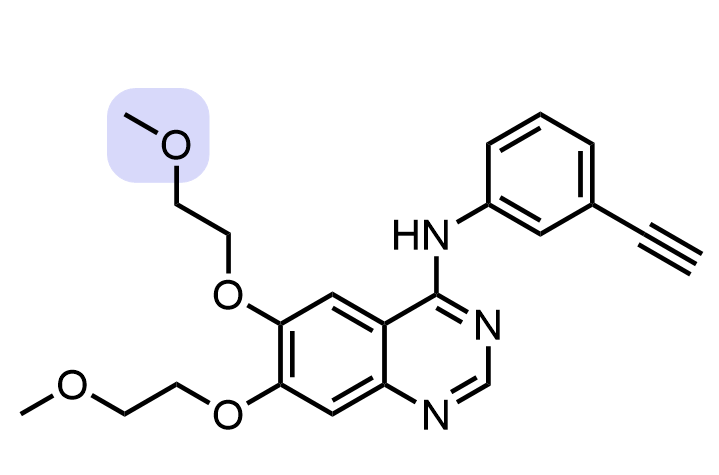}} &
    \adjustbox{valign=c}{\includegraphics[width=0.2\linewidth]{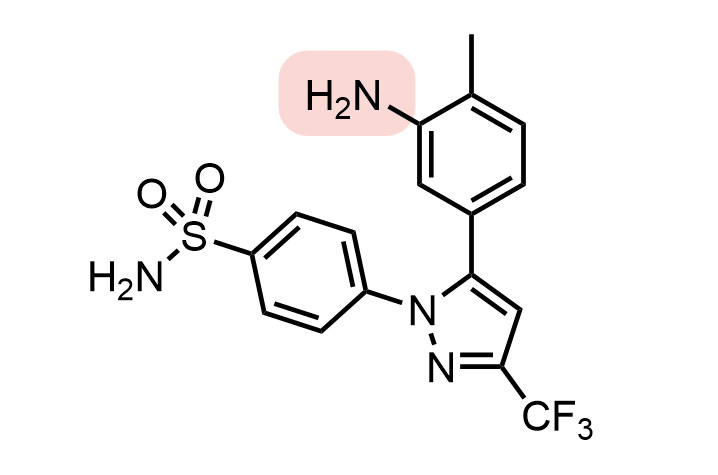}} &
    \adjustbox{valign=c}{\includegraphics[width=0.2\linewidth]{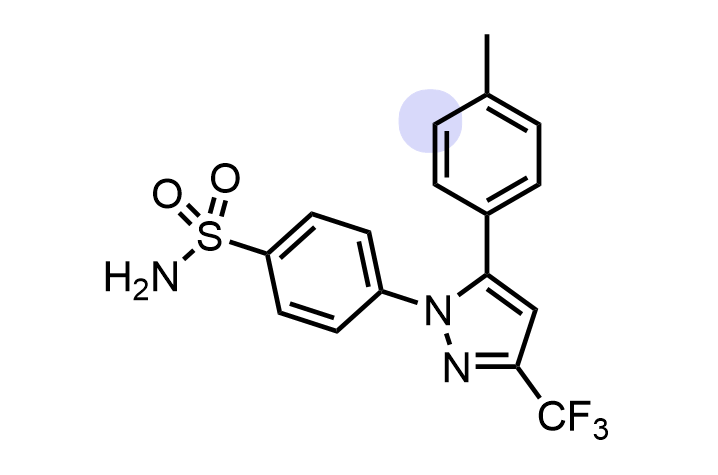}} &
    \adjustbox{valign=c}{\includegraphics[width=0.2\linewidth]{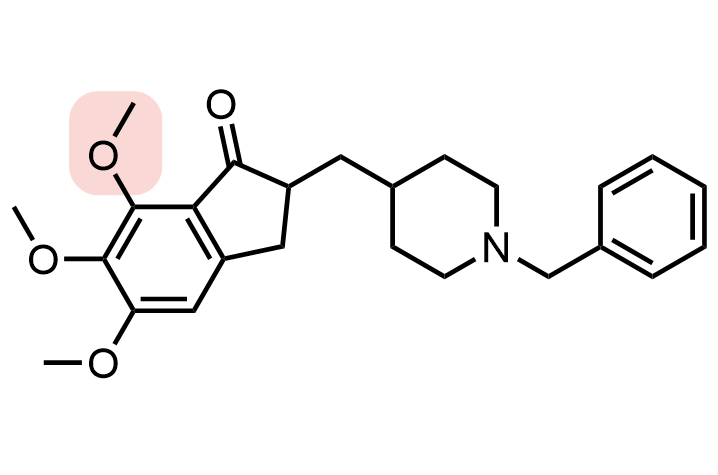}} &
    \adjustbox{valign=c}{\includegraphics[width=0.2\linewidth]{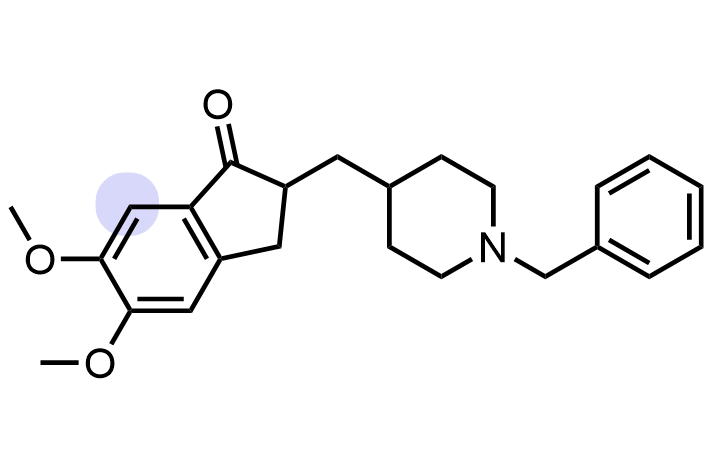}}
    \\
    {\small CAS: 183320-43-6} & {\small Tarceva (Erlotinib)} & 
    {\small CAS: 170570-28-2} & {\small Celebrex (Celecoxib)} &
    {\small CAS: 120013-52-7} & {\small Aricept (Donepezil)}
    \\[5pt]
    \bottomrule
    \end{tabular}
    \end{adjustbox}
\end{table}

\clearpage
\section{Downstream: Molecular Property Prediction}
In this section, we review \two{} main categories of datasets used for molecular property prediction downstream tasks from MoleculeNet and molecule benchmarking works~\cite{wu2018moleculenet,wang2022evaluating}.

\paragraph{Molecular Property: Pharmacology}
The Blood-Brain Barrier Penetration (BBBP)~\cite{martins2012bayesian} dataset measures whether a molecule will penetrate the central nervous system. All \three{} toxicity-related datasets, Tox21~\cite{tox21}, ToxCast~\cite{wu2018moleculenet}, and ClinTox~\cite{gayvert2016data} are related to the toxicity of molecular compounds. The Side Effect Resource (SIDER)~\cite{kuhn2015sider} dataset stores the adverse drug reactions on a marketed drug database.

\paragraph{Molecular Property: Biophysics}
Maximum Unbiased Validation (MUV)~\cite{doi:10.1021/ci8002649} is another sub-database from PCBA, and is obtained by applying a refined nearest neighbor analysis. HIV is from the Drug Therapeutics Program (DTP) AIDS Antiviral Screen~\cite{zaharevitz2015aids}, and it aims at predicting the inhibition of HIV replication. BACE measures the binding results for a set of inhibitors of $\beta$-secretase 1 (BACE-1) and is gathered in MoleculeNet~\cite{wu2018moleculenet}.

\begin{table}[H]
\centering
\small
\caption{Summary for the molecule chemical datasets.}
\vspace{-2ex}
\begin{tabular}{l l r r r r}
    \toprule
    Dataset & Task & \# Tasks & \# Molecules\\
    \midrule
    BBBP & Classification & 1 & 2,039 \\
    Tox21 & Classification & 12 & 7,831 \\
    ToxCast & Classification & 617 & 8,576 \\
    Sider & Classification & 27 & 1,427 \\
    ClinTox & Classification & 2 & 1,478 \\
    MUV & Classification & 17 & 93,087 \\
    HIV & Classification & 1 & 41,127 \\
    Bace & Classification & 1 & 1,513 \\
    \bottomrule
\end{tabular}
\label{tab:mol_dataset_summary}
\end{table}

For data splitting, we adopt the scaffold splitting~\cite{wu2018moleculenet}. Scaffold measures the skeleton structure of molecules, and scaffold splitting means we will put the molecules with more common scaffolds into training, and the rest into validation and test, so as to mimic the out-of-distribution (OOD) setting. The OOD setting is more common in real scenarios and thus is preferred to test the pretrained molecule representation power.

\paragraph{Implementation Details}
For the SMILES string, we use MegaMolBART~\cite{irwin2022chemformer} as the backbone Transformer model. For the molecular graph, we use the same backbone GIN model, and we use rich features (as used for the regression tasks in GraphMVP~\cite{liu2021pre}). We list the main hyperparameters below.
\begin{table}[htb]
\centering
\setlength{\tabcolsep}{5pt}
\fontsize{9}{9}\selectfont
\caption{
\small
Hyperparameter specifications for molecular property prediction.
}
\vspace{-2ex}
\begin{adjustbox}{max width=\textwidth}
\begin{tabular}{l l l}
\toprule
 & Hyperparameter & Value\\
\midrule
\multirow{3}{*}{Pretraining Baseline}
& epochs & \{100\} \\
& learning rate & \{1e-3\} \\
& weight decay & \{0\} \\
\midrule
\multirow{3}{*}{Downstream}
& epochs & \{100\} \\
& learning rate & \{1e-3, 5e-4\} \\
& weight decay & \{0\} \\
\bottomrule
\end{tabular}
\end{adjustbox}
\end{table}

\textbf{Choice of backbone models.}
We want to clarify that the \model{} is agnostic to the backbone encoders for each modality, {\eg}, the molecule representation model.
\begin{itemize}[noitemsep,topsep=0pt]
    \item For the backbone model, we use the GIN model as the fixed 2D GNN backbone encoder. In other words, the performance of MoleculeSTM is limited by the 2D backbone model.
    \item In the molecule pretraining research line, ({\eg}, AttrMask~\cite{hu2019strategies}, MolCLR~\cite{wang2021molclr}, GraphMVP~\cite{liu2021pre}, MoleculeSDE~\cite{liu2023group}), all of these works adopt GIN as the 2D backbone model, serving as a control to test the effectiveness of various pretraining algorithms. This is a similar case for our proposed \model{}.
    \item In the future, we would like to explore more advanced GNN models on molecules.
\end{itemize}
\clearpage

{
\renewcommand*{\bibfont}{\small}
\printbibliography[title={Supplementary References}]
}
\end{refsection}

\end{document}